\pdfoutput=1

\documentclass[11pt]{article}

\usepackage[final]{acl}

\usepackage{times}
\usepackage{latexsym}

\usepackage[T1]{fontenc}

\usepackage[utf8]{inputenc}

\usepackage{microtype}

\usepackage{inconsolata}

\usepackage{graphicx}
\usepackage{amsmath}
\usepackage{amsfonts}
\usepackage{amssymb}
\usepackage{graphicx} 
\usepackage{appendix}
\usepackage{booktabs}
\usepackage{soul,xspace}
\usepackage{xcolor}
\usepackage[flushleft]{threeparttable}
\usepackage{makecell}
\usepackage{hyperref}
\usepackage{multirow}
\usepackage{multicol}
\usepackage{verbatim}
\usepackage{multirow}
\usepackage{multicol}
\usepackage{soul}
\usepackage{longtable,booktabs,threeparttablex}
\usepackage{subcaption}

\title{\agent: Language-based Intelligent Drug Discovery Agent}

\author{
  \textbf{Reza Averly\textsuperscript{*1}},
  \textbf{Frazier N. Baker\textsuperscript{*1}},
  \textbf{Ian A. Watson\textsuperscript{2}} \&
  \textbf{Xia Ning\textsuperscript{1,3,4,5}} \\
  \textsuperscript{1}Department of Computer Science and Engineering, The Ohio State University, USA\\
  \textsuperscript{2}Independent Researcher\\
  \textsuperscript{3}Department of Biomedical Informatics, The Ohio State University, USA\\
  \textsuperscript{4}Translational Data Analytics Institute, The Ohio State University, USA\\
  \textsuperscript{5}College of Pharmacy, The Ohio State University, USA\\
  \textsuperscript{*}Equal contribution \\
  \texttt{\{averly.1,baker.3239\}@buckeyemail.osu.edu,ianiwatson@gmail.com,}\\
  \texttt{ning.104@osu.edu}
  }

\newcommand{\generate}{\mbox{$\mathop{\textsc{Generate}}\limits$}\xspace}
\newcommand{\optimize}{\mbox{$\mathop{\textsc{Optimize}}\limits$}\xspace}
\newcommand{\code}{\mbox{$\mathop{\textsc{Screen}}\limits$}\xspace}
\newcommand{\evaluator}{\mbox{$\mathop{\textsc{Evaluator}}\limits$}\xspace}
\newcommand{\executor}{\mbox{$\mathop{\textsc{Executor}}\limits$}\xspace}
\newcommand{\planner}{\mbox{$\mathop{\textsc{Reasoner}}\limits$}\xspace}
\newcommand{\memory}{\mbox{$\mathop{\textsc{Memory}}\limits$}\xspace}

\newcommand{\agent}{\mbox{$\mathop{\textsc{LIDDiA}}\limits$}\xspace}

\newcommand{\coscientist}{\mbox{$\mathop{\textsc{CoScientist}}\limits$}\xspace}

\newcommand{\chemcrow}{\mbox{$\mathop{\textsc{ChemCrow}}\limits$}\xspace}
\newcommand{\alab}{\mbox{$\mathop{\textsc{A-Lab}}\limits$}\xspace}
\newcommand{\cactus}{\mbox{$\mathop{\textsc{CACTUS}}\limits$}\xspace}
\newcommand{\drugagent}{\mbox{$\mathop{\textsc{DrugAgent}}\limits$}\xspace}
\newcommand{\protagent}{\mbox{$\mathop{\textsc{ProtAgent}}\limits$}\xspace}

\newcommand{\moleculeset}{\mbox{$\mathop{\mathcal{M}}\limits$}\xspace}
\newcommand{\molecule}{\mbox{$\mathop{m}\limits$}\xspace}
\newcommand{\target}{\mbox{$\mathop{t}\limits$}\xspace}
\newcommand{\tanimoto}{\mbox{$\mathop{\text{sim}_{\mathsf{T}}}\limits$}\xspace}

\newcommand{\ANGMPT}{\mbox{$\mathop{\mathtt{Generated}}\limits$}\xspace}

\newcommand{\ANGVMPT}{\mbox{$\mathop{\mathtt{Valid}}\limits$}\xspace}

\newcommand{\HQ}{\mbox{$\mathop{\mathtt{HQ}}\limits$}\xspace}
\newcommand{\noveltyreq}{\mbox{$\mathop{\mathtt{NVT}}\limits$}\xspace}
\newcommand{\qedreq}{\mbox{$\mathop{\mathtt{QED}}\limits$}\xspace}
\newcommand{\lipinskireq}{\mbox{$\mathop{\mathtt{LRF}}\limits$}\xspace}
\newcommand{\sascorereq}{\mbox{$\mathop{\mathtt{SAS}}\limits$}\xspace}
\newcommand{\vinareq}{\mbox{$\mathop{\mathtt{VNA}}\limits$}\xspace}
\newcommand{\diversityreq}{\mbox{$\mathop{\mathtt{DVS}}\limits$}\xspace}

\newcommand{\successrate}{\mbox{$\mathop{\mathtt{TSR}}\limits$}\xspace}

\begin{document}
\maketitle
\begin{abstract}
Drug discovery is a long, expensive, and complex process, relying 
heavily on human medicinal chemists, who can spend years searching the vast space of potential therapies.
Recent advances in artificial intelligence for chemistry have 
sought to expedite individual drug discovery tasks; however, there remains a critical need for an intelligent agent that 
can navigate the drug discovery process.
Towards this end, we introduce \agent, an autonomous agent capable of intelligently navigating the drug discovery process \textit{in silico}.
By leveraging the reasoning capabilities of large language models,
\agent serves as a low-cost and highly-adaptable tool for autonomous drug discovery. 
We comprehensively examine \agent, demonstrating that (1) it can generate molecules meeting key pharmaceutical criteria on over $70\%$ of 30 clinically relevant targets, (2) it intelligently balances exploration and exploitation in the chemical space, and (3) it identifies one promising novel candidate on AR/NR3C4, a critical target for both prostate and breast cancers.
Code and dataset are available at \url{https://github.com/ninglab/LIDDiA}.

\end{abstract}

\section{Introduction}
\label{sec:intro}

Artificial intelligence (AI) research has long sought to develop 
agents capable of intelligent
reasoning to aid humans 
by autonomously navigating complex, resource-intensive processes.
Drug discovery is one such process, relying 
heavily on human medicinal chemists, who can spend years searching the vast space of potential therapies~\cite{blass_chapter_2021}.
Recent advances~\cite{chen2020retro, zhao_meea_2024, trott_autodock_2010, swanson_admet-ai_2024, zhou_optimization_2019, jensen_graph-based_2019} in AI for chemistry have 
sought to expedite drug discovery by performing individual tasks \emph{in silico}.
However, there remains a critical need for an intelligent, autonomous agent that 
can strategically navigate and facilitate the drug discovery process.

Drug discovery is a complex, nonlinear process with many requirements.
Successful drugs must not only bind well to their therapeutic targets, 
but also exhibit good physicochemical, pharmacodynamic, and pharmacokinetic properties. 
These requirements are not necessarily independent;
changing a molecule to satisfy one may result in the violation of another.
Medicinal chemists combine manual analysis with computational tools to identify promising molecules, 
evaluate their properties, and optimize their structures---an iterative process 
that demands substantial time and effort.

Large language models (LLMs) have emerged as reasoning engines capable 
of intelligent reasoning and planning over complex tasks.
Recent works~\cite{
yao_react_2022, 
liu_conversational_2023, 
boiko_autonomous_2023, 
zhou_automated_2023} have explored leveraging LLMs as intelligent agents,
using natural language as an interface for taking actions and observing results.
By pairing LLM's reasoning capabilities with computational tools for
drug discovery,
we envision
building a digital twin of the medicinal chemist, 
capable of navigating the complexities of the drug discovery process.

In this work, we introduce \agent, an intelligent agent for 
navigating the pre-clinical drug discovery process \emph{in silico}.
\agent is composed of four interconnected components:
(1) \planner, 
(2) \executor, 
(3) \evaluator, 
and (4) \memory.
Each component interacts with the others to collaboratively navigate the drug discovery process.
By harnessing the pre-trained knowledge and reasoning capabilities of LLMs,
\agent enables intelligent and rational decision-making over drug discovery steps, 
mimicking experienced medicinal chemists and steering the drug discovery process toward 
high throughput and success rate.
In doing so, \agent orchestrates the intelligent use 
of computational tools (e.g., docking simulation, property prediction, molecule optimization).
One key strength of \agent lies in its integration of generative AI tools for molecular design, 
enabling it to explore vast chemical spaces beyond conventional molecular libraries.  
With a modular architecture,
\agent is designed for flexibility, allowing it to be seamlessly extended or refined as new capabilities emerge.
To the best of our knowledge, \agent is the first of its kind, representing the first effort toward 
low-cost, high-efficiency, autonomous drug discovery.  

We rigorously benchmark \agent (Section~\ref{sec:results}) 
and demonstrate that it can produce promising drug candidates satisfying key pharmaceutical properties 
on more than $70\%$ of 30 major therapeutic targets (Section~\ref{sec:results:main}). 
We provide an in-depth study (Section~\ref{sec:results:analysis}), 
illustrating that \agent strategically generates, refines, and selects highly favorable molecules, 
well aligned with a real-world drug discovery workflow (Section~\ref{sec:results:behaviour}). 
We also identify a salient pattern underpinning successful outcomes for \agent: 
effectively balancing exploration and exploitation in the chemical space (Section~\ref{sec:results:case_study_action}). 
Lastly, we highlight one promising drug
candidate for AR/NR3C4, an important target for prostate and breast cancers
(Section~\ref{sec:results:case_study_compounds}).
%

\section{Related Work}

LLMs equipped with tools
have recently shown great promise as autonomous agents for scientific discoveries, including
drug discovery~\cite{gao_empowering_2024}.
For instance, AutoBA~\cite{zhou_automated_2023} uses LLMs
to automate multi-omics bioinformatics analysis,
generating new insights using well-known bioinformatics 
libraries.
Another example for biomedical research is \protagent~\cite{ghafarollahi_protagents_2024}, an LLM agent system for \emph{de novo} protein design, equipped with physics-based simulations to ground its design process.
Materials science can also benefit from LLM agents as well, as shown by
\alab~\cite{szymanski_autonomous_2023}, a self-driving laboratory that uses an LLM agent to control both analysis tools and laboratory hardware for semiconductor material design.

For drug discovery, 
\coscientist~\cite{boiko_autonomous_2023} uses LLMs to perform web search,
conduct technical documentation, program, and operate physical hardware modules
to plan and control chemical synthetic experiments.
It demonstrates the viability LLM agents equipped with both physical and computational tools to 
act as self-driving laboratories for organic chemistry. 
However, \coscientist does not integrate any domain-specific tools for grounding, but rather relies upon the LLM's intrinsic chemistry knowledge, web search, and results from the physical experiments.

In addition, \chemcrow~\cite{m_bran_augmenting_2024} is an LLM agent equipped with specific tools for small molecule organic chemistry.
These tools support generating molecule structures from natural language descriptions, predicting molecule properties,
conducting \emph{in silico} safety checks, and performing retrosynthesis planning.
Grounded by these tools, \chemcrow demonstrates an ability to perform complex, multi-step chemistry tasks commonly found in the drug discovery process. 
\cactus~\cite{mcnaughton_cactus_2024} is a similar agent to \chemcrow,
emphasizing tools that can predict properties important to drug discovery.
\drugagent~\cite{inoue_drugagent_2024} is an LLM agent for drug repurposing, equipped with tools to search databases of existing drugs to identify candidates likely to interact with a protein target.
Notably, none of these LLM agents are grounded with well-established computational tools for 
novel structure-based drug discovery (SBDD).  

\section{\agent Framework}
\label{sec:methods}

\begin{figure*}[ht]
    \centering
    \includegraphics[width=0.95\linewidth]{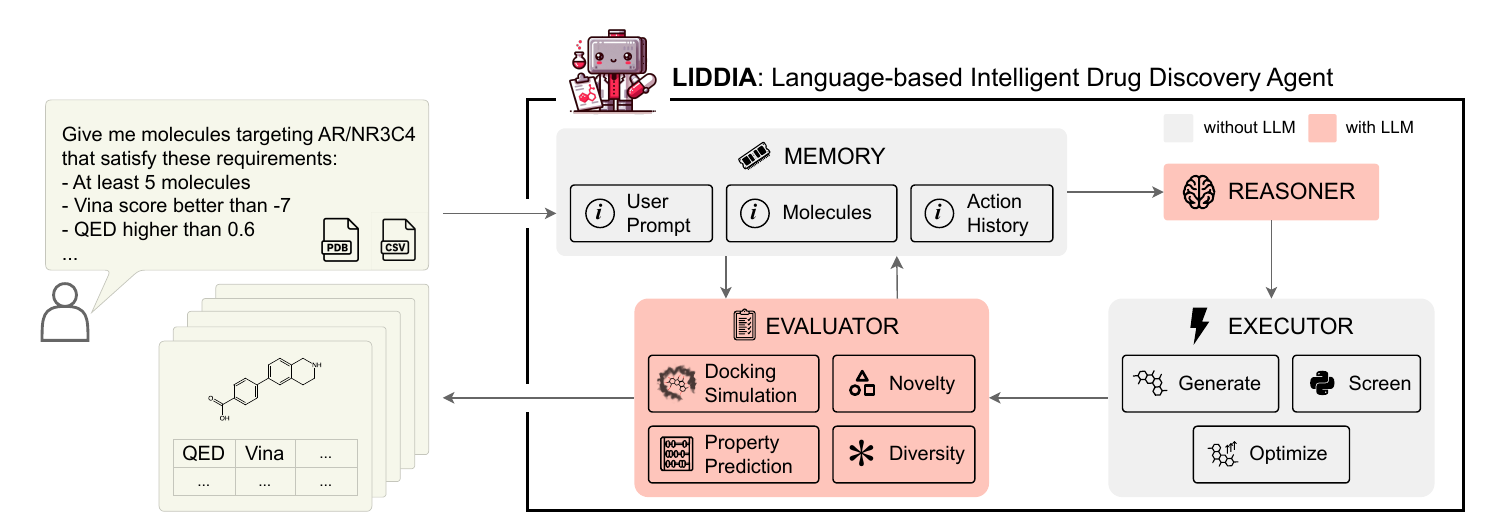}
    \vspace{-8pt}
    \caption{Overview of the \agent Framework}\vspace{-11pt} 
    \label{fig:overview}
    \vskip -8pt
\end{figure*}

\agent is an automated, agentic framework for navigating the drug discovery process by combining computational tools and reasoning capabilities.
As illustrated in Figure~\ref{fig:overview}, \agent is composed of four components:
(1) \planner (Section~\ref{sec:methods:reasoner}), 
which plans \agent's actions and directs \agent to conduct drug discovery;   
(2) \executor (Section~\ref{sec:methods:executor}), 
which executes \planner's actions using state-of-the-art computational tools;
(3) \evaluator (Section~\ref{sec:methods:evaluator}), 
which assesses candidate molecules;   
and 
(4) \memory (Section~\ref{sec:methods:memory}), which keeps  
all the information produced along the drug discovery process. 
Each of these components represents a logical abstraction from the traditional drug discovery, 
enhanced by computational tools and generative AI.
The ultimate goal is, given a target of interest and property specifications 
on its potential drugs (e.g., at least 5 molecules, binding affinities better than -7, drug-likeness better than 0.5),  
\agent produces a diverse set of 
high-quality molecules that satisfy these specifications and can be considered as 
potential drug candidates for the target. 
Overall, \agent represents an innovative initiative towards autonomous drug discovery, 
integrating AI-driven planning, execution, evaluation, and memory management to 
accelerate the identification and optimization of novel therapeutics. 

\subsection{\planner}
\label{sec:methods:reasoner}

%
\agent's decision-making is conducted through its \planner component. 
Using the information in \memory (e.g., molecules under current consideration 
and their property profiles),
\planner conducts reasoning and
strategically plans the next actions that \agent should take, 
leveraging the pre-trained knowledge and reasoning capabilities of LLMs.
\planner explores three action types:
(1) \generate to generate new molecules; 
(2) \optimize to optimize existing molecules; and 
(3) \code to process the current molecules. 
These actions correspond to several key steps in hit identification (via generative AI) and lead 
optimization in pre-clinical drug discovery. 
Therefore, \planner is key to guiding \agent through the iterative process of molecular design, 
ensuring that each decision aligns with the overall requirements of drug discovery 
while balancing all required properties.

\subsection{\executor}
\label{sec:methods:executor}

\agent executes all drug discovery actions planned by \planner through its \executor.  
\executor is equipped with state-of-the-art computational tools tailored to different actions. 
Specifically, \executor integrates:  
(1). generative models for structure-based
drug design 
to implement the \generate action, 
utilizing methods such as Pocket2Mol~\cite{peng_pocket2mol_2022};
(2). generative models for molecular refinement to implement the \optimize action, 
enhancing drug-like properties and optimizing molecular structures~\cite{jensen_graph-based_2019}; and
(3). a processor to implement the \code action, allowing complex and logic screening, organizing and 
managing molecules, and identifying the most promising ones.  

A key innovation of \agent is that \executor leverages generative models for both hit identification 
(\generate) and lead optimization (\optimize). 
Unlike conventional drug discovery, which relies on searching and modifying existing molecular databases, 
\executor enables \emph{de novo} molecular generation, 
expanding the chemical space beyond known molecules. 
This approach increases the likelihood of discovering novel, diverse, and more effective drug candidates.  
Moreover, by automating lead optimization through generative models, \executor reduces human bias
stemming from individual expertise levels and limited chemical knowledge. 
This ensures a more systematic, data-driven approach to improving molecular properties.  
By integrating generative tools, \agent can operate autonomously, 
designing superior drug candidates more efficiently than traditional human-driven search methods. 
This automation enhances cost-effectiveness and accelerates drug discovery, 
making \agent a powerful AI-driven co-pilot in the drug discovery process.

\subsection{\evaluator}
\label{sec:methods:evaluator}

\agent performs \emph{in silico} assessments over molecules using its \evaluator. 
\evaluator assesses an array of molecule properties essential to successful 
drug candidates, including target binding affinity, drug-likeness, synthetic accessibility, 
Lipinski's rule, novelty, and diversity.
For different properties, \evaluator uses the appropriate computational tools to conduct the evaluation.
Evaluation results are systematically stored in \agent's memory 
(\memory; discussed in Section~\ref{sec:methods:memory}), and subsequently utilized 
by \planner to refine decision-making and guide the next steps in the drug discovery process.
Once \evaluator identifies molecules satisfying all user requirements, it signals \agent to terminate 
the search and return the most promising candidates. 
Thus, \evaluator serves as a critical quality control mechanism, 
systematically steering \agent toward identifying optimal drug candidates 
while minimizing the exploration of suboptimal chemical spaces.
In addition, \evaluator is designed in a plug-and-play fashion, 
allowing additional tools to be added or updated to support new property requirements. 

\subsection{\memory}
\label{sec:methods:memory}

\agent keeps all the information produced throughout its entire drug discovery process
in \memory. 
This includes information provided by users via prompts, such as
protein target structures, 
property requirements, and reference molecules (e.g., known drugs). 
More information will be dynamically generated as \agent progresses 
through the drug discovery process, including 
the trajectories of actions that \agent has taken (planned by \planner), 
molecules generated from prior actions, and their properties. 
The information in \memory is aggregated and provided to \planner to facilitate its planning. 
\memory is dynamically changing and continuously updated. 
This evolving repository enables \planner to be well informed by prior knowledge and newly
generated data, and thus better reflect and refine its strategies, 
enhancing the efficiency and effectiveness of automated drug discovery.

\begin{table}[t!]
\centering
\scriptsize
\caption{Statistics over protein targets.}
\vspace{-8pt}
\label{tab:dataset_stats}
\begin{footnotesize}
\begin{threeparttable}
\begin{tabular}{
    @{\hspace{2pt}}l@{\hspace{30pt}}
    @{\hspace{20pt}}r@{\hspace{2pt}}
}
\toprule
Disease                                        & \#Targets \mbox{(\%)} \\ 
\midrule
Cancers                                        & 15 (50\%)                     \\
Neurological \mbox{Conditions}   & 8 (27\%)                        \\
Cardiovascular \mbox{Diseases} & 6 (20\%)                        \\
Infectious \mbox{Diseases}          & 4 (13\%)                        \\
Diabetes                                       & 3 (10\%)                        \\
Autoimmune \mbox{Diseases}     & 3 (10\%)                        \\ 
\bottomrule
\end{tabular}
\begin{tablenotes}[normal,flushleft]
\begin{scriptsize}
\centering
\setlength\labelsep{0pt}
    \item Some targets are associated with multiple categories.
    \par
\end{scriptsize}
\end{tablenotes}
\vspace{-20pt}
\end{threeparttable}
\end{footnotesize}
\end{table}

\section{Experimental Settings}
\label{sec:experiments}

\subsection{Evaluation Metrics}
\label{sec:experiments:evaluation}

\paragraph{Molecule Qualities}
We use these metrics to evaluate the  molecules generated by different methods. 

\subparagraph{Key molecule properties}
%
We first evaluate the following general properties required for successful drugs:
\textbf{(1)} drug-likeness~\cite{bickerton_quantifying_2012} (\qedreq), 
\textbf{(2)} Lipinski's Rule of Five~\cite{lipinski_experimental_2001} (\lipinskireq),
\textbf{(3)} synthetic accessibility~\cite{ertl_estimation_2009} (\sascorereq), 
and 
\textbf{(4)} binding affinities measured by Vina scores~\cite{trott_autodock_2010} (\vinareq).  
Evaluation on more molecule properties (e.g. toxicity properties) is available in Appendix~\ref{sec:appendix:results:toxicity}.

\subparagraph{Novelty}
%
We measure the novelty (\noveltyreq) of a molecule \molecule with respect to a 
reference set of known drugs $\moleculeset_0$ as follows: 
\vspace{-5pt}
\begin{equation*}
\label{eqn:novelty}
\noveltyreq(\molecule; \moleculeset_0) \! 
= \! 1 - 
\max\nolimits_{\scriptsize{\molecule_i \in \moleculeset_0}}
(\tanimoto(\molecule, \molecule_i)),
\vspace{-5pt}
\end{equation*}
where $\moleculeset_0$ is the reference set of known drugs, 
$\molecule$ and $\molecule_i$ are two molecules,
and $\tanimoto(\molecule, \molecule_i)$ is the Tanimoto similarity 
of $\molecule$ and $\molecule_i$'s Morgan fingerprints~\cite{morgan_generation_1965}.
High novelty indicates that new molecules are different from existing drugs, 
offering new therapeutic opportunities. 
A molecule \molecule is considered novel if $\noveltyreq(\molecule) \ge 0.8$. 

\subparagraph{High-quality molecules}

A molecule \molecule is considered as ``high quality'' (\HQ) for a target \target, 
if its properties satisfy 
$\qedreq\ge\overline{\qedreq}_{\scriptsize{\target}}, 
\lipinskireq \ge \overline{\lipinskireq}_{\scriptsize{\target}}, 
\sascorereq \le \overline{\sascorereq}_{\scriptsize{\target}}, 
\vinareq \le \overline{\vinareq}_{\scriptsize{\target}}$, and 
$\noveltyreq(\molecule) \ge 0.8$, 
where the $\overline{\text{overline}}$ and the subscript ${{\target}}$
indicate the average value from all the known drugs for target \target. 
Such multi-property requirements are typical in drug discovery. 
Meanwhile, this presents a significant challenge, as \agent must identify molecules
with key properties similar to or even better than existing drugs but structurally significantly 
different from them.  
In our dataset, we include existing drugs for targets as the gold standard 
for evaluation purposes.

\paragraph{Molecule Set Diversity}

We measure the diversity (\diversityreq) of a set of generated molecules \moleculeset
defined as follows, 
\vspace{-5pt}
\begin{equation*}
\label{eqn:diversity}
\diversityreq(\moleculeset) = 1 - \mathop{\mathbb{E}}\nolimits_{\scriptsize{\{\molecule_i, \molecule_j\}} \subseteq \moleculeset}\left[\tanimoto(\molecule_i, \molecule_j)\right], 
\vspace{-5pt}
\end{equation*}
where $\molecule_i$ and $\molecule_j$ are two distinct molecules in $\moleculeset$.
High diversity is preferred, as chemically diverse molecules 
increase the likelihood of identifying successful drug candidates.
A set of molecules \moleculeset is considered diverse if $\diversityreq(\moleculeset) \ge 0.8$. 
This imposes a highly stringent requirement on the diversity of the generated molecules. 
\paragraph{Target Success Rate}

Target success rate, denoted as \successrate, is defined as the percentage of targets 
for which a method can generate a \emph{diverse} set of \emph{at least 5} \emph{high-quality} molecules.
%

\subsection{Protein Target Dataset}
\label{sec:experiments:data}

To evaluate \agent, we manually curated a diverse set of protein targets 
from OpenTargets~\cite{ochoa_opentargets_2023}
that are strongly associated with major human diseases: 
cancers, neurological conditions,
cardiovascular diseases, infectious diseases, diabetes,
and autoimmune diseases.
For each of these protein targets,
we identified an experimentally resolved structure with a small-molecule ligand
from the RCSB Protein Data Bank (PDB)~\cite{berman_protein_2000} 
and extracted the binding pocket according to its ligand's position.
To enable a comparison to existing drugs,
we searched ChEMBL~\cite{bento_chembl_2014, gaulton_chembl_2011}
for all known drugs targeting the selected proteins.
This leads to 30 protein targets with PDB structures, ligands, and existing drugs. 
These targets will be used as input in our experiments. 
Table~\ref{tab:dataset_stats} presents the distribution 
of the targets in terms of their disease associations. 
Please note, some targets are associated with multiple diseases.
A full list of targets is presented in Appendix~\ref{sec:appendix:dataset}. We discuss the importance of manual curation of this dataset in Appendix~\ref{sec:appendix:dataset:extra}.

%

\subsection{Implementation Details}
\label{sec:experiments:implementation}

\agent leverages Claude 3.5 Sonnet~\cite{anthropic_claude-sonnet3-5_2024} as the base model for its \planner and \evaluator
since it achieves state-of-the-art performance in chemistry related tasks~\cite{chen2024scienceagentbench,huang2024olympicarena}.
We designed and fine-tuned specific prompts to guide \planner and \evaluator, respectively. 
Details on these prompts are provided in Appendix~\ref{sec:appendix:prompts}.
\evaluator evaluates all the metrics as defined in Section~\ref{sec:experiments:evaluation}. 
We set the maximum number of actions taken to 10 to ensure a concise yet effective drug discovery trajectory.

\executor executes the \generate action using Pocket2Mol~\cite{peng_pocket2mol_2022}.
As a structure-based drug design tool, Pocket2Mol can generate molecules using only the target protein structure.
This provides \agent with the ability to extend to novel targets without known ligands. 
For efficiency, Pocket2Mol is set to generate a minimum of 100 molecules using a beam size of 300.
The \optimize action is implemented via GraphGA~\cite{jensen_graph-based_2019}, 
a popular graph-based genetic algorithm for molecule optimization.
In \agent, \optimize can refine molecules on three essential properties: 
drug-likeness (\qedreq), synthetic accessibility (\sascorereq), 
and target binding affinity (\vinareq).
However, the actions can be easily expanded to cover additional properties.

\begin{table*}[!t]
\caption{Performance comparison between the baseline methods and \agent. Full results is available in Table~\ref{tab:complete}.} 
  \label{tab:main}
  \centering
   \begin{footnotesize}	
  \begin{threeparttable}
      \begin{tabular}{
        @{\hspace{0pt}}l@{\hspace{0pt}}
        @{\hspace{2pt}}l@{\hspace{1pt}}
        @{\hspace{0pt}}r@{\hspace{1pt}} 
        @{\hspace{1pt}}r@{\hspace{0pt}}
        @{\hspace{5pt}}c@{\hspace{5pt}}	
        @{\hspace{0pt}}r@{\hspace{1pt}} 
        @{\hspace{1pt}}r@{\hspace{0pt}}
        @{\hspace{5pt}}c@{\hspace{5pt}}	
        @{\hspace{0pt}}r@{\hspace{1pt}} 
        @{\hspace{1pt}}r@{\hspace{0pt}}
        @{\hspace{5pt}}c@{\hspace{5pt}}	
        @{\hspace{0pt}}r@{\hspace{1pt}} 
        @{\hspace{1pt}}r@{\hspace{0pt}}
        @{\hspace{5pt}}c@{\hspace{5pt}}	
        @{\hspace{0pt}}r@{\hspace{1pt}} 
        @{\hspace{1pt}}r@{\hspace{0pt}}
        @{\hspace{5pt}}c@{\hspace{5pt}}	
        @{\hspace{0pt}}r@{\hspace{1pt}} 
        @{\hspace{1pt}}r@{\hspace{0pt}}
        @{\hspace{5pt}}c@{\hspace{5pt}}	
        @{\hspace{0pt}}r@{\hspace{1pt}} 
        @{\hspace{1pt}}r@{\hspace{0pt}}
      }
      \toprule
      & 
      & \multicolumn{2}{c}{Pocket2Mol} & 
      & \multicolumn{2}{c}{DiffSMOL}    & 
      & \multicolumn{2}{c}{Claude}        & 
      & \multicolumn{2}{c}{GPT-4o}        & 
      & \multicolumn{2}{c}{o1-mini}        & 
      & \multicolumn{2}{c}{o1}                & 
      & \multicolumn{2}{c}{\agent} 
      \\
      \midrule
      &
      & \%\molecule/\target & \#\molecule/\target & 
      & \%\molecule/\target & \#\molecule/\target & 
      & \%\molecule/\target & \#\molecule/\target & 
      & \%\molecule/\target & \#\molecule/\target & 
      & \%\molecule/\target & \#\molecule/\target & 
      & \%\molecule/\target & \#\molecule/\target & 
      & \%\molecule/\target & \#\molecule/\target \\
      \cmidrule(){3-4} \cmidrule(){6-7} \cmidrule(){9-10} \cmidrule(){12-13} \cmidrule(){15-16} \cmidrule(){18-19} \cmidrule(){21-22}
      \multirow{2}{*}{\rotatebox{90}{\centering{initial}}}
      & \ANGMPT 
      & - & 100.0  & 
      & - & 100.0  & 
      & - & \phantom{0}\phantom{0}5.0 & 
      & - & \phantom{0}\phantom{0}5.0 & 
      & - & \phantom{0}\phantom{0}5.0 & 
      & - & \phantom{0}\phantom{0}5.0 & 
      & - & \phantom{0}24.5 \\
      %
      & \ANGVMPT 
      & 100.0 & 100.0 &
      & 99.9 & 99.9     & 
      & 98.7 & 4.9       &
      & 97.3 & 4.9       & 
      & 91.3 & 4.6       & 
      & 95.3 & 4.8       & 
      & 100.0 & 24.5 \\
      %
      \cmidrule(){2-22}
      %
      \multirow{6}{*}{\rotatebox{90}{\parbox{60pt}{\centering{generated molecules}}}}
      & \quad$\qedreq\!\ge\!\overline{\qedreq}_{t}$ 
      & 53.4 & 53.4 &
      & 60.0 & 60.0 & 
      & \underline{96.7} & 4.8 & 
      & 88.2 & 4.4 & 
      & 90.1 & 4.5 & 
      & 88.3 & 4.4 & 
      & \textbf{97.2} & 21.8 \\
      %
      &\quad$\lipinskireq\!\ge\!\overline{\lipinskireq}_t$
      & \textbf{99.7} & 99.7 & 
      & 72.1 & 72.1 &
      & \underline{98.7} & 4.9 & 
      & 95.9 & 4.8 & 
      & 90.7 & 4.5 & 
      & 95.3 & 4.8 & 
      & 96.7 & 21.8 \\
      %
      &\quad$\sascorereq\!\le\!\overline{\sascorereq}_t$
      & 77.4 & 77.4 &
      & 7.5 & 7.5 & 
      & \textbf{92.7} & 4.6 & 
      & 90.7 & 4.5 &
      & 81.4 & 4.1 & 
      & \underline{92.6} & 4.6 & 
      & 88.3 & 17.4 \\
      %
      & \quad$\vinareq\!\le\!\overline{\vinareq}_t$
      & 15.3 & 15.3 &
      & 24.7 & 24.7 &
      & \underline{63.3} & 3.2 &
      & 59.2 & 3.0 &
      & 47.9 & 2.3 &
      & 34.6 & 1.8 &
      & \textbf{95.8} & 21.2 \\
      %
      & \quad$\noveltyreq\!\ge\!0.8$
      & 87.6 & 87.6 &
      & \textbf{98.2} & 98.2 & 
      & 46.9 & 2.4 &
      & 68.3 & 3.4 & 
      & 64.1 & 3.2 & 
      & 55.9 & 2.8 & 
      & \underline{97.8} & 22.4 \\
      \cmidrule(l{10pt}){2-22}
      & \HQ 
      & 6.4 & 6.4 &
      & 0.7 & 0.7 &
      & 30.3 & 1.5 &
      & \underline{35.0} & 1.7 &
      & 28.2 & 1.4 &
      & 20.7 & 1.0 &
      & \textbf{84.0} & 14.5 \\
      \cmidrule(){1-22}

      \multirow{5}{*}{\rotatebox{90}{\parbox{70pt}{among all targets}}}
      &
      & \%\target & \#\target & 
      & \%\target & \#\target & 
      & \%\target & \#\target & 
      & \%\target & \#\target & 
      & \%\target & \#\target & 
      & \%\target & \#\target & 
      & \%\target & \#\target \\
      \cmidrule(){3-4} \cmidrule(){6-7} \cmidrule(){9-10} \cmidrule(){12-13} \cmidrule(){15-16} \cmidrule(){18-19} \cmidrule(){21-22}
      & \quad$\diversityreq\!\ge\!0.8$
      & \textbf{100.0} & 30 &
      & \textbf{100.0} & 30 &
      & 30.0 & 9 &
      & 90.0 & 27 &
      & 67.7 & 20 &
      & 70.0 & 21 &
      & \underline{97.7} & 29 \\
      %
      & \quad$N$$\ge$5\! \&\! \diversityreq
      & \textbf{100.0} & 30 &
      & \textbf{100.0} & 30 &
      & 27.7 & 8 &
      & 77.7 & 23 &
      & 43.3 & 13 &
      & 57.7 & 17 &
      & \underline{90.0} & 27 \\
      %
      & \quad$N$$\ge$5\! \&\! \HQ 
      & \underline{27.7} & 8 &
      & 3.3 & 1 &
      & 23.3 & 7 &
      & 10.0 & 3 &
      & 0.0 & 0 &
      & 3.3 & 1 &
      & \textbf{73.3} & 22  \\
      %
      & \quad\diversityreq \& \HQ
      & 23.3 & 7 &
      & 10.0 & 3 &
      & 10.0 & 3 &
      & \underline{33.3} & 10 &
      & \underline{33.3} & 10 &
      & 20.0 & 6 &
      & \textbf{90} & 27 \\
      %
      \cmidrule(l{10pt}){2-22}
      & \successrate 
      & \underline{23.3} & 7 &
      & 0.0 & 0 &
      & 6.7 & 2 &
      & 6.7 & 2 &
      & 0.0 & 0 &
      & 0.0 & 0 &
      & \textbf{73.3} & 22 \\
      \midrule
 	%
      %
      \multicolumn{22}{c}{Quality of Generated Molecules}
      \\
      \cmidrule(){2-22}	
      & \noveltyreq$\uparrow$ 
      & \multicolumn{2}{c}{\underline{0.87}} &
      & \multicolumn{2}{c}{\textbf{0.89}} &
      & \multicolumn{2}{c}{0.77} &
      & \multicolumn{2}{c}{0.82} & 
      & \multicolumn{2}{c}{0.79} &
      & \multicolumn{2}{c}{0.80} &
      & \multicolumn{2}{c}{0.86} \\
     &  \qedreq$\uparrow$ 
     & \multicolumn{2}{c}{0.51} &
     & \multicolumn{2}{c}{0.55} &
     & \multicolumn{2}{c}{\textbf{0.78}} &
     & \multicolumn{2}{c}{0.74} &
     & \multicolumn{2}{c}{0.75} &
     & \multicolumn{2}{c}{\underline{0.77}} &
     & \multicolumn{2}{c}{0.69} \\
     & \lipinskireq$\uparrow$ 
     & \multicolumn{2}{c}{\textbf{4.00}} &
     & \multicolumn{2}{c}{3.43} &
     & \multicolumn{2}{c}{\textbf{4.00}} &
     & \multicolumn{2}{c}{\underline{3.99}} &
     & \multicolumn{2}{c}{3.85} &
     & \multicolumn{2}{c}{\textbf{4.00}} &
     & \multicolumn{2}{c}{3.93} \\
     & \sascorereq$\downarrow$ 
     & \multicolumn{2}{c}{2.46} &
     & \multicolumn{2}{c}{6.15} &
     & \multicolumn{2}{c}{2.30} &
     & \multicolumn{2}{c}{2.16} &
     & \multicolumn{2}{c}{\textbf{2.02}} &
     & \multicolumn{2}{c}{\underline{2.03}} &
     & \multicolumn{2}{c}{2.62} \\
     & \vinareq$\downarrow$ 
     & \multicolumn{2}{c}{-4.74} &
     & \multicolumn{2}{c}{-4.23} &
     & \multicolumn{2}{c}{\underline{-6.69}} &
     & \multicolumn{2}{c}{-6.56} &
     & \multicolumn{2}{c}{-6.31} &
     & \multicolumn{2}{c}{-5.97} &
     & \multicolumn{2}{c}{\textbf{-7.17}} \\
      & \diversityreq$\uparrow$ 
      & \multicolumn{2}{c}{\underline{0.88}} &
      & \multicolumn{2}{c}{\textbf{0.89}} &
      & \multicolumn{2}{c}{0.76} &
      & \multicolumn{2}{c}{0.84} &
      & \multicolumn{2}{c}{0.79} &
      & \multicolumn{2}{c}{0.80} &
      & \multicolumn{2}{c}{0.82} \\

      \bottomrule
      \end{tabular}
      \begin{tablenotes}[normal,flushleft]
	\begin{scriptsize}
	\setlength\labelsep{0pt}
    	\item 
	\%\molecule/\target: average percentage of molecules per target; 
	\#\molecule/\target: average number of molecules per target; 
	\ANGMPT: initially generated molecules; 
	\ANGVMPT: generated molecules that are also valid; 
	$\overline{\text{overline}}_{\scriptsize{\target}}$: the average value of corresponding property in the known drugs for 
	the target \target. 
	\%\target: average percentage of targets among all targets; 
	\#\target: average number of targets; 
	$N$$\!\ge\!$5\&\diversityreq: at least 5 molecules are generated and the set is diverse;  
	$N$$\!\ge\!$5\&\HQ: at least 5 molecules are generated and they are of high quality; 
	$\uparrow$/$\downarrow$ indicates higher/lower values are better. 
    \textbf{Bold} and \underline{underline} indicates the best and second-best results, respectively.
    \par
	\end{scriptsize}
  \end{tablenotes}
  \end{threeparttable}
  \end{footnotesize}      
  \vskip -15pt
\end{table*}

\subsection{Baselines}
\label{sec:experiments:baselines}

We compare \agent with two types of baselines:
task-specific molecule generation methods, and general-purpose LLMs. 
For molecule generation methods, 
we use Pocket2Mol~\cite{peng_pocket2mol_2022} and DiffSMol~\cite{chen_generating_2025}.
Pocket2Mol~\cite{peng_pocket2mol_2022} is a well-established generative method 
for structure-based drug design, which uses binding pocket structures as input. 
DiffSMol~\cite{chen_generating_2025}, on the other hand, is a state-of-the-art 
generative method for ligand-based drug design, requiring a binding ligand. 
These two methods represent distinct approaches in computational drug design, 
using different information to generate potential drug candidates. 
Notably, Pocket2Mol is used by \agent in \generate actions, 
capitalizing on the popularity of SBDD and its ability to generate molecules without reference ligands.
%


For general-purpose LLMs, we use 
GPT-4o~\cite{openai_gpt-4o_2024}, o1~\cite{openai_o1_2024}, 
o1-mini~\cite{openai_o1-mini_2024}, and Claude 3.5 Sonnet~\cite{anthropic_claude-sonnet3-5_2024}.
GPT-4o and Claude 3.5 Sonnet are representative state-of-the-art language models; 
o1 and o1-mini are specifically tailored towards scientific reasoning during their training.
We evaluate all four of these models as baselines to 
provide a comprehensive understanding of the performance of state-of-the-art LLMs.

\section{Experimental Results}
\label{sec:results}

\subsection{Main Results}
\label{sec:results:main}

Table~\ref{tab:main} presents the performance of different methods,
including their success rates
and the qualities of their generated molecules. We include the full results in Table~\ref{tab:complete}.

\emph{\textbf{\agent successfully generates novel, diverse, and high-quality molecules as 
potential drug candidates 
for $73.3\%$ of targets (\successrate), significantly outperforming existing methods.}}  
Pocket2Mol, the second-best method, achieves only a $23.3\%$ success rate, 
while most proprietary LLMs fail entirely.  
Crucially, \agent excels in simultaneously optimizing all five key pharmaceutical properties -- 
\qedreq, \lipinskireq, \sascorereq, \vinareq, and \noveltyreq 
-- on average, 85\% of the generated molecules for each target are of high quality (\HQ). 
In contrast, GPT-4o 
achieves only $35\%$ in \HQ, lagging nearly 50 percentage points behind \agent, 
while all other methods perform even worse.  
In terms of the qualities of the generated molecules, 
\agent produces molecules of comparable or superior quality to the limited outputs 
of other methods. 
These results highlight \agent as a highly effective and reliable framework for accelerating drug discovery, 
consistently outperforming existing methods in both success rate and molecule qualities.
We also compare \agent with more recent state-of-the-art methods and observe similar findings as presented in Appendix~\ref{sec:appendix:results:baselines}.

\begin{figure}[ht]
    \centering
    \includegraphics[width=0.75\linewidth]{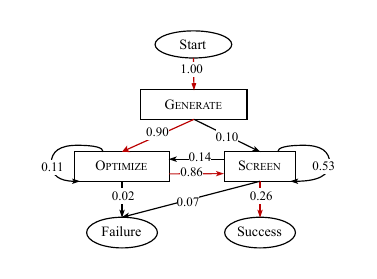}
    \vskip -5pt
    \caption{Action transitions in \agent. The numbers represent the transition probabilities.}
    \label{fig:action_graph}
    \vskip -8pt
\end{figure}

\begin{figure}[ht]
    \centering
    \includegraphics[width=0.75\linewidth]{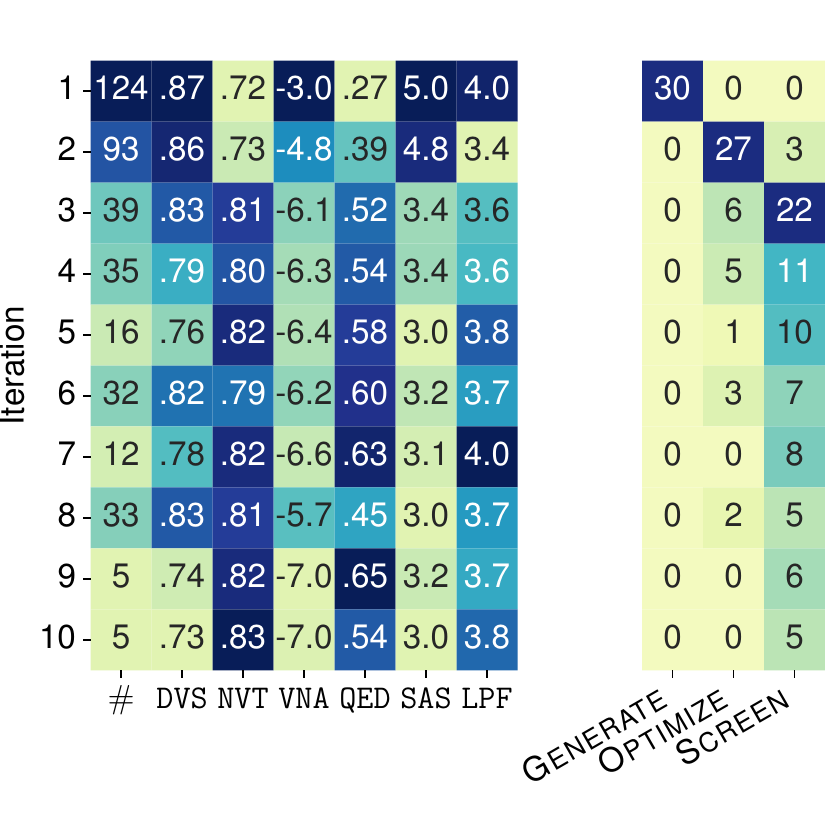}
    \vskip -8pt
    \caption{Molecule quality (left panel) and actions (right panel) over iterations by \agent.} 
    \label{fig:iteration}
    \vskip -20pt
\end{figure}

\emph{\textbf{\agent consistently generates high-quality molecules across 
all key pharmaceutical properties.}}  
Notably,  it produces the most molecules (97.2\%) for each target with \qedreq higher than average, 
and the most molecules (95.8\%) for each target with \vinareq higher than average, compared to other methods.
Most of them ($97.8\%$) are also novel, second only to DiffSMol.
With respect to known drugs, a vast majority (88.3\%) of \agent's molecules exhibit high synthetic accessibility (\sascorereq) and
(96.7\%) adhere to Lipinski’s Rule of Five (\lipinskireq),  
comparable to general-purpose LLMs such as Claude, GPT-4o, and o1.  
On the most stringent metric, \vinareq, \agent significantly outperforms other methods, 
with $95.8\%$ of its generated molecules binding similarly to or better than existing drugs.  
In contrast, existing methods struggle to exceed $65\%$ across these key properties. 
Overall, \agent proves to be a robust and reliable approach, 
surpassing existing methods in generating high-quality drug candidates.
We also show that \agent's generated molecules are better than or comparable to known drugs in terms of their toxicity properties. We provide further discussion in Appendix~\ref{sec:appendix:results:toxicity}.

\begin{figure*}[ht]
    \centering
    \begin{subfigure}[b]{0.33\textwidth}
        \centering
        \includegraphics[width=\textwidth]{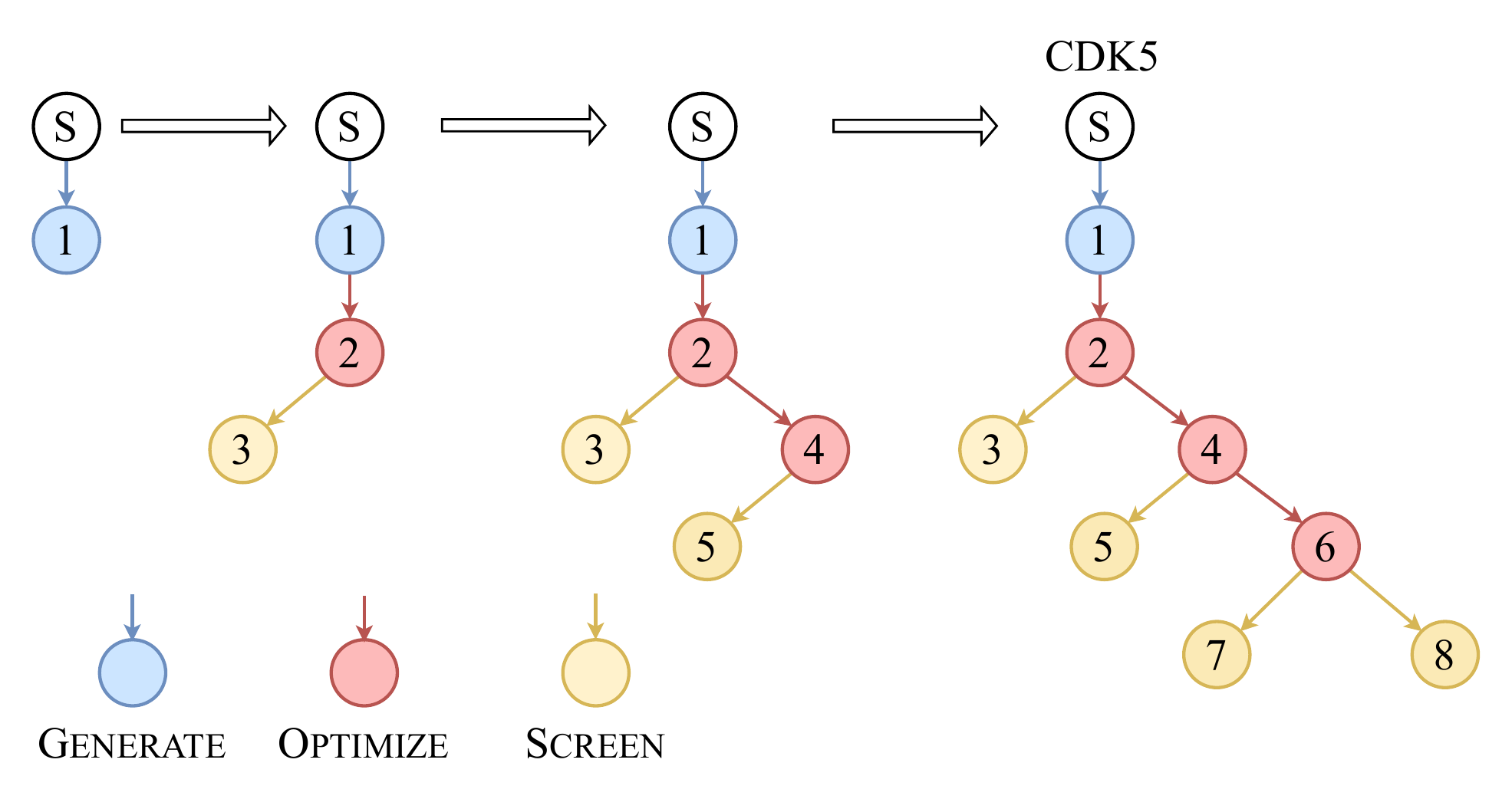}
        \vskip -4pt
        \caption{Successful actions REASfor CDK5}
        \label{fig:sub1}
    \end{subfigure}
    \hfill
    \begin{subfigure}[b]{0.28\textwidth}
        \centering
        \includegraphics[width=\textwidth]{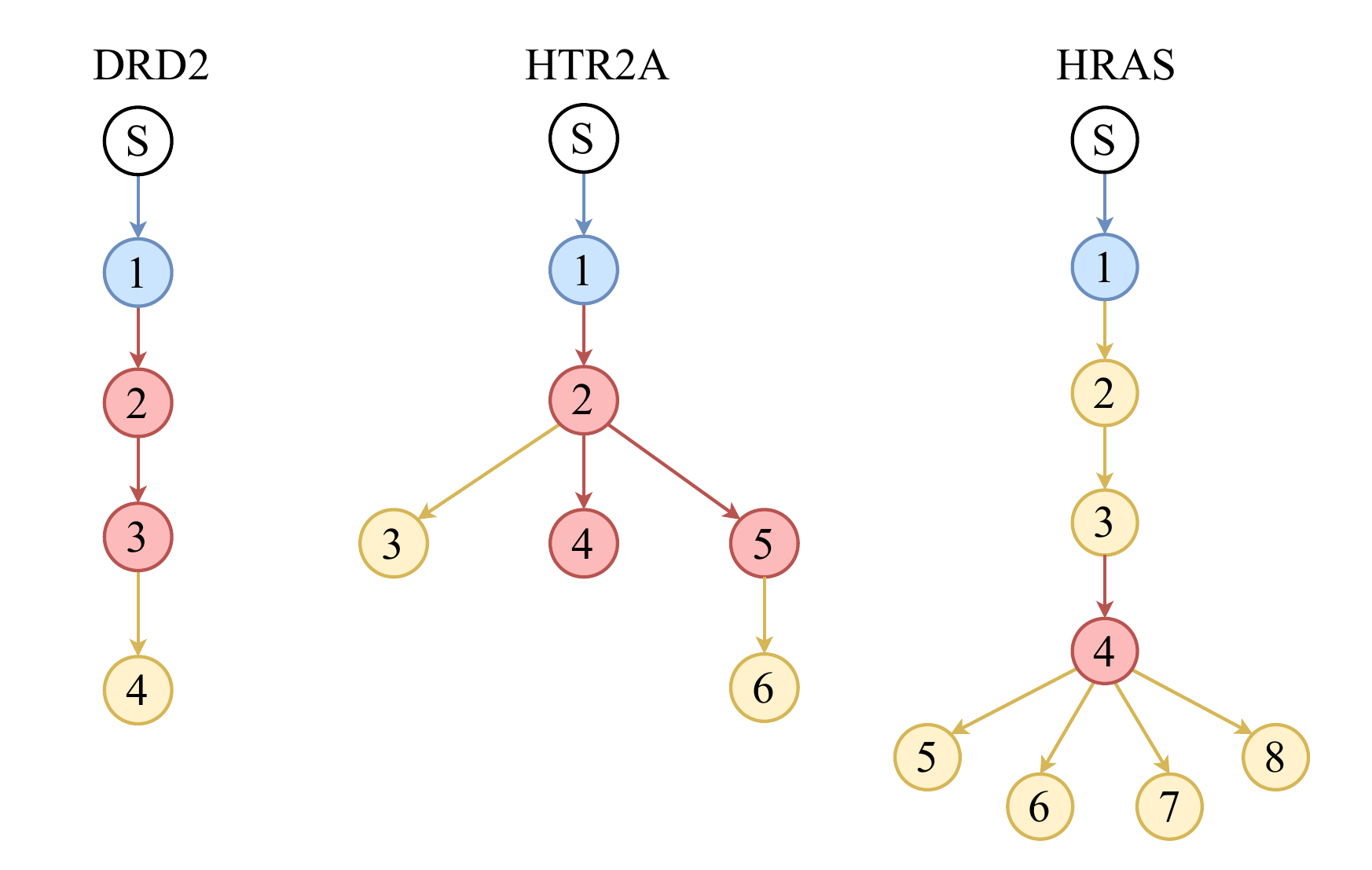}
        \vskip -4pt
        \caption{Successful action trajectories}
        \label{fig:sub2}
    \end{subfigure}
    \hfill
    \begin{subfigure}[b]{0.33\textwidth}
        \centering
        \includegraphics[width=\textwidth]{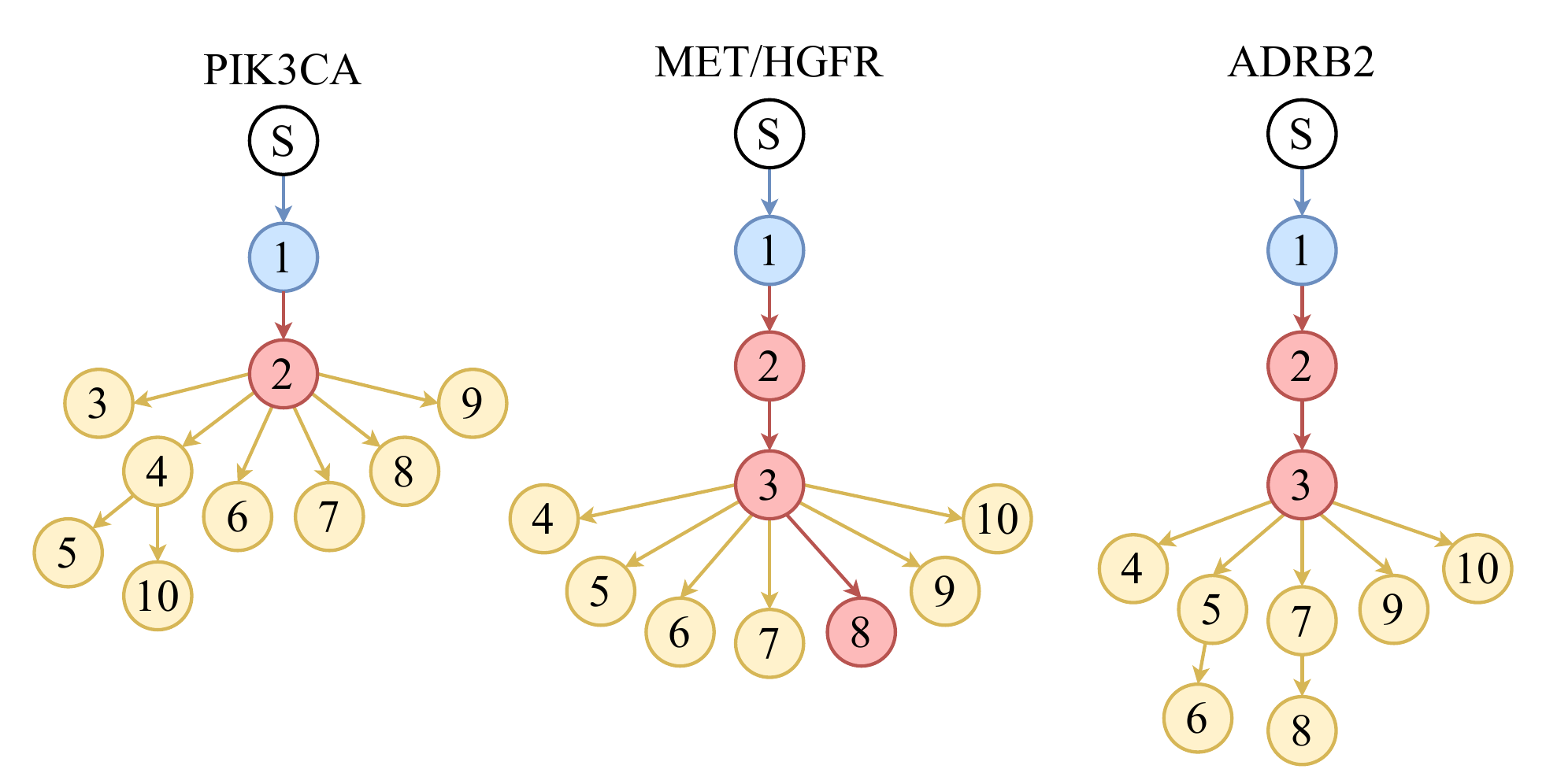}
        \vskip -4pt
        \caption{Failure action trajectories}
        \label{fig:sub3}
    \end{subfigure}
    \vskip -4pt
    \caption{\agent actions trajectories across different targets.}
    \label{fig:case_study_action}
    \vskip -15pt
\end{figure*}

\emph{\textbf{Existing methods face substantial challenges to achieve multiple good properties concurrently.}}
For instance, general-purpose LLMs -- Claude, GPT4o, o1-mini, and o1-- exhibit a trade-off between 
novelty and binding. 
While they achieve impressive \vinareq, 
their generations tend to resemble known drugs closely 
(e.g., \textless$70\%$ novelty). 
In contrast, Pocket2mol and DiffSMol demonstrate the opposite: excel at \noveltyreq but struggle at \vinareq.
It is possible that LLMs often anchor their generations based on prior knowledge 
(e.g., known ligands), thus narrowing their explorations. 
Meanwhile, Pocket2Mol and DiffSMol can generate new binding molecules but not better than existing drugs. 
Note that \agent does not suffer from such compromise (e.g., both \sascorereq and \vinareq \textgreater$95\%$).
%

\subsection{Agent Analysis}
\label{sec:results:analysis}

\subsubsection{\agent action patterns}
\label{sec:results:behaviour}

\autoref{fig:action_graph} presents the transition probabilities of actions that \agent takes 
throughout the drug discovery process across all the targets, from start to finish.

\emph{\textbf{\agent aligns with a typical drug discovery workflow, 
incorporating intelligent refinement at every stage.}}
The most likely strategy of \agent begins with the generation of target-binding molecules 
(\generate), followed by either optimization to enhance their properties (\optimize), or 
screening and selection over the generated molecules (\code).
Typically, optimization is necessary, which is followed by molecule screening
over the optimized molecules. Iterative optimization is possible when no viable molecules exist. 
Similarly, iterative molecule screening is employed 
when plenty of viable molecules exist but are structurally similar. 
For instance, \agent may cluster these molecules and 
subsequently identify the most promising molecules within each cluster.
The most common workflow covers \generate, \optimize, and then \code toward 
successful outcomes.

\textbf{\emph{\code serves as a quality-control mechanism to enable successful outcomes.}}
Successful molecules are only possible after \code completes screening and selection and 
identifies such molecules to output. 
As \generate and \optimize tend to yield more molecules than
necessary, allowing abundant opportunities for \agent to succeed, \code prevents \agent 
from producing low-quality drug candidates.   

\textbf{\emph{Most of the generated molecules from \generate directly go through subsequent 
optimization by \optimize.}}
This occurs in approximately 90\% of the cases. 
Among all the generated molecules by \generate, it is typical that none of them satisfies 
all the property requirements, particularly those properties that are not integrated into the \generate 
tool designs. 
Any screening by \code over such molecules will be futile and wasteful. 
Instead, \agent intelligently executes \optimize, improving the likelihood of successful molecules 
out of \code screening. 
Meanwhile, \agent can still identify high-quality generated molecules and conducts 
screening directly over them. 
This clearly demonstrates the reasoning capability of \agent as an effective tool for drug 
discovery.  

\emph{\textbf{\agent leverages performance-driven insights to determine the most optimal action.}}
Confronted with low-quality outputs from \generate, \agent 
selectively pursues optimization to maximize results.
Whenever additional molecules need to be considered (e.g., \code does not identify good candidates), \agent prioritizes optimizing promising molecules stored in \memory rather than generating entirely new ones.  
This represents a cost-effective, risk-averse strategy, balancing exploration and exploitation by refining known candidates with high potential rather than investing computational resources in \emph{de novo} generation 
with suboptimal outcomes.

\emph{\textbf{\agent favors refining a few highly promising candidates as it continuously progresses.}}
\autoref{fig:iteration} describes both the quality of the molecules produced and the typical actions \agent takes at each step. 
Compared to the initial pool, the output molecules roughly achieve double the \qedreq and \vinareq, 
thus emphasizing the importance of iterative optimization and effective screening strategies.
However, diversity among these top candidates is often limited since molecules 
that satisfy multiple property requirements tend to converge on similar structures.
This further highlights the complexity of drug discovery.~\autoref{fig:iteration} (right) also highlights how \agent tends to focus on refining the molecules via \optimize or \code in later steps, mimicking a typical drug discovery workflow. 

\begin{figure}[t]
    \centering
    \includegraphics[width=0.55\linewidth]{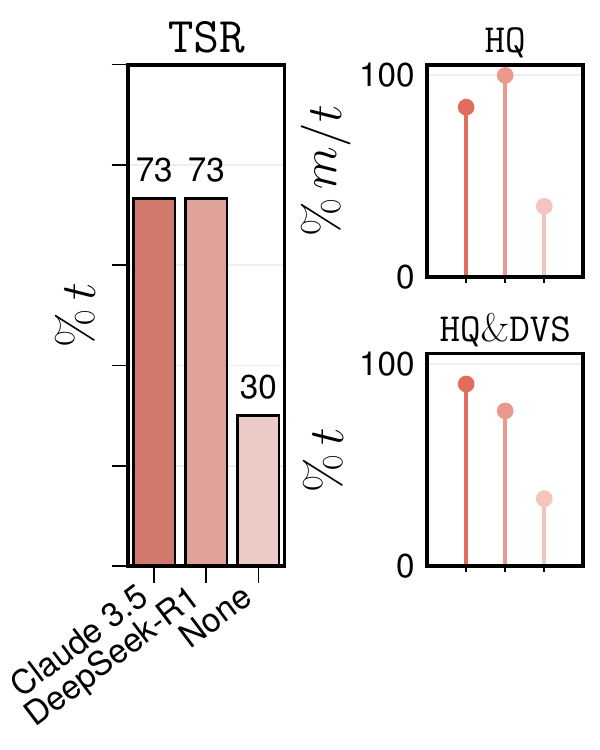}
    \vskip -8pt
    \caption{Ablation study on \agent.} 
    \label{fig:ablation}
    \vskip -15pt
\end{figure}

\subsubsection{Analysis on \agent action trajectories}
\label{sec:results:case_study_action}

Figure~\ref{fig:case_study_action} presents an example of \agent action trajectories to identify 
promising drug candidates for CDK5~\cite{lau_cdk5_2002}, 
a target for neurological conditions, such as Alzheimer's Dementia, along with examples 
of trajectories that lead to success (e.g., HTR2A, HRAS, DRD2) and fail (e.g., PIK3CA, MET, ADRB2)  outcomes, respectively. 

\textit{\textbf{\agent intelligently balances exploration and exploitation, critical to identify promising candidates}}.
In the case of CDK5, where it is highly challenging to identify a good drug candidate as demonstrated 
in the literature~\cite{xie_lessons_2022},
\agent is able to adaptively explore the chemical space via iteratively screening viable molecules and improving any property the molecules fail to meet. 
As shown in~\autoref{fig:case_study_action} (a), \agent starts with \generate (step 1), proceeds with \optimize (step 2), then applies \code (step 3) but fails to find favorable candidates. 
In response, \agent refines the failing property (step 4) and performs another screening (step 5). This process continues (steps 6, 7, 8) until the agent finally converges to a set of promising candidates.

This not an isolated case; \agent consistently displays comparable intelligent decision-making behavior on other targets as observed in~\autoref{fig:case_study_action} (b) and~\autoref{fig:case_study_action} (c).
Notably, in cases with successful outcomes, \agent methodically refines several properties before screening for candidates (DRD2), or strategically determines which molecules to prioritize and what action to take (HTR2A and HRAS). For instance, in the HRAS case, \agent uses several screenings (steps 2 and 3) to identify viable candidates, optimize them (step 4), and conduct further screenings (step 5 to 8) until it identifies favorable candidates. 
This highlights one key strength of \agent -- its capability to adapt to feedback (e.g., molecules quality) from its \evaluator, to explore (e.g., via refinement and generation), and to exploit (e.g., screening existing molecules) the chemical space.
On cases where \agent yields suboptimal results, such as PIK3CA, MET, and ADRB2, 
it still exhaustively performs various actions up to the action limits (i.e., 10 iterations). 
Additional analysis of PIK3CA, MET, and ADRB2 can be found in Appendix~\ref{sec:appendix:results:failure}.

\begin{figure}[t]
\centering
    \begin{subfigure}{0.32\linewidth}
        \includegraphics[width=\linewidth]{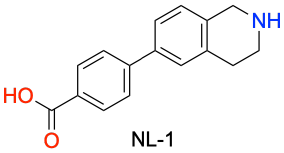}  
        \caption{}
        \label{fig:ar:agent} 
    \end{subfigure}
    \hfill
    \begin{subfigure}{0.32\linewidth}
        \includegraphics[width=\linewidth]{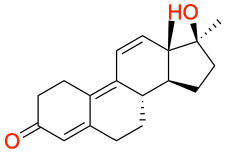}  
        \caption{}
        \label{fig:ar:ligand} 
    \end{subfigure}
    \hfill
    \begin{subfigure}{0.32\linewidth}
        \includegraphics[width=\linewidth]{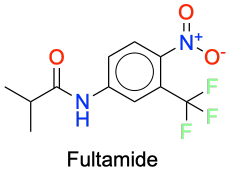}    
        \caption{}
        \label{fig:ar:drugs} 
    \end{subfigure}
    \vspace{-8pt}
    \caption{Case study on AR/NR3C4. (a) Promising molecule (NL-1) generated by \agent. (b) Known ligand for AR/NR3C4. (c) An example of known approved drugs for AR/NR3C4.}
    \vspace{-8pt}
    \label{fig:ar}
\end{figure}

\begin{figure}[t]
\centering
    \begin{subfigure}{0.45\linewidth}
        \includegraphics[width=\linewidth]{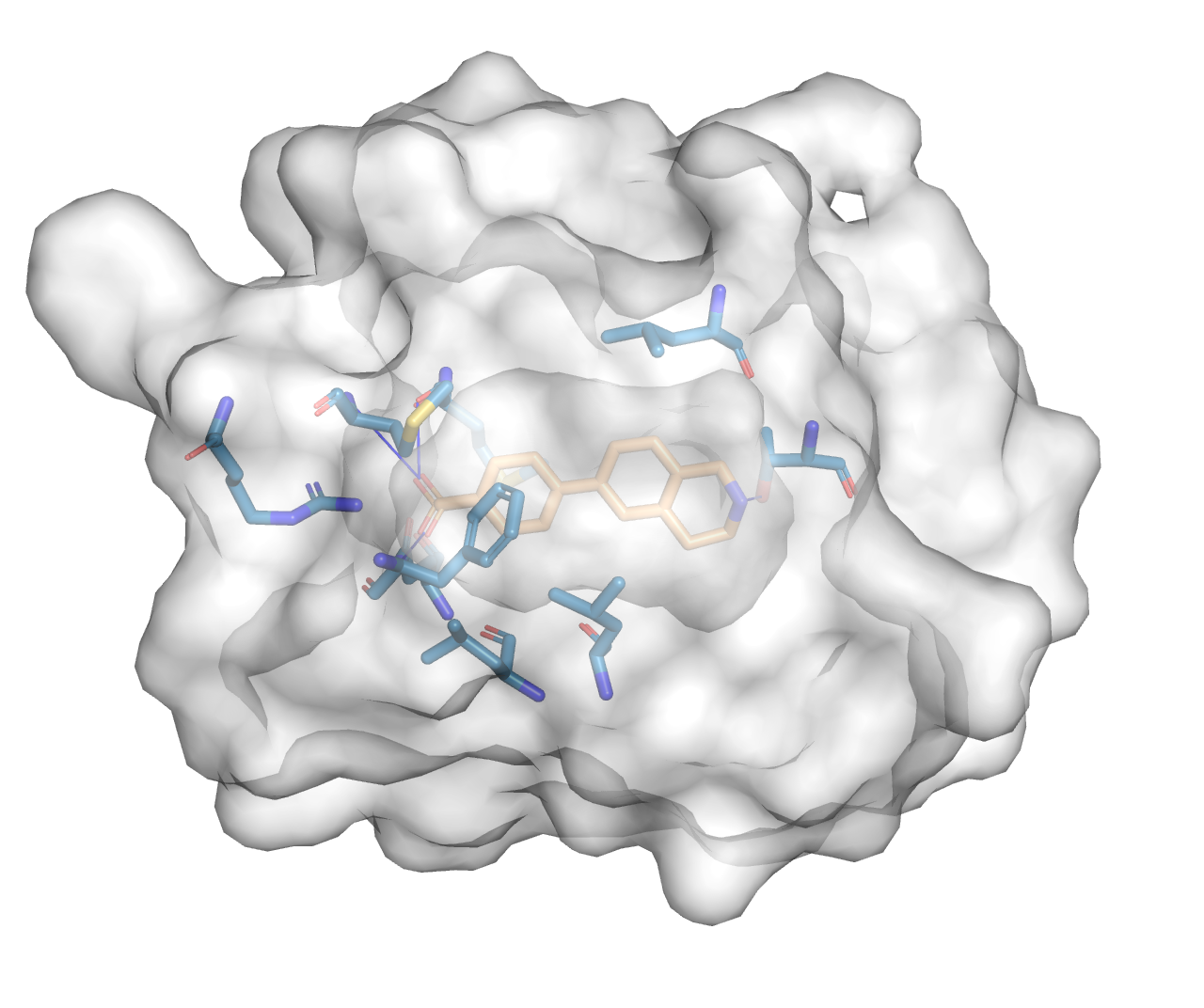}    
        \caption{}
        \label{fig:pos_case_docking:ar} 
    \end{subfigure}
    \begin{subfigure}{0.45\linewidth}
        \includegraphics[width=\linewidth]{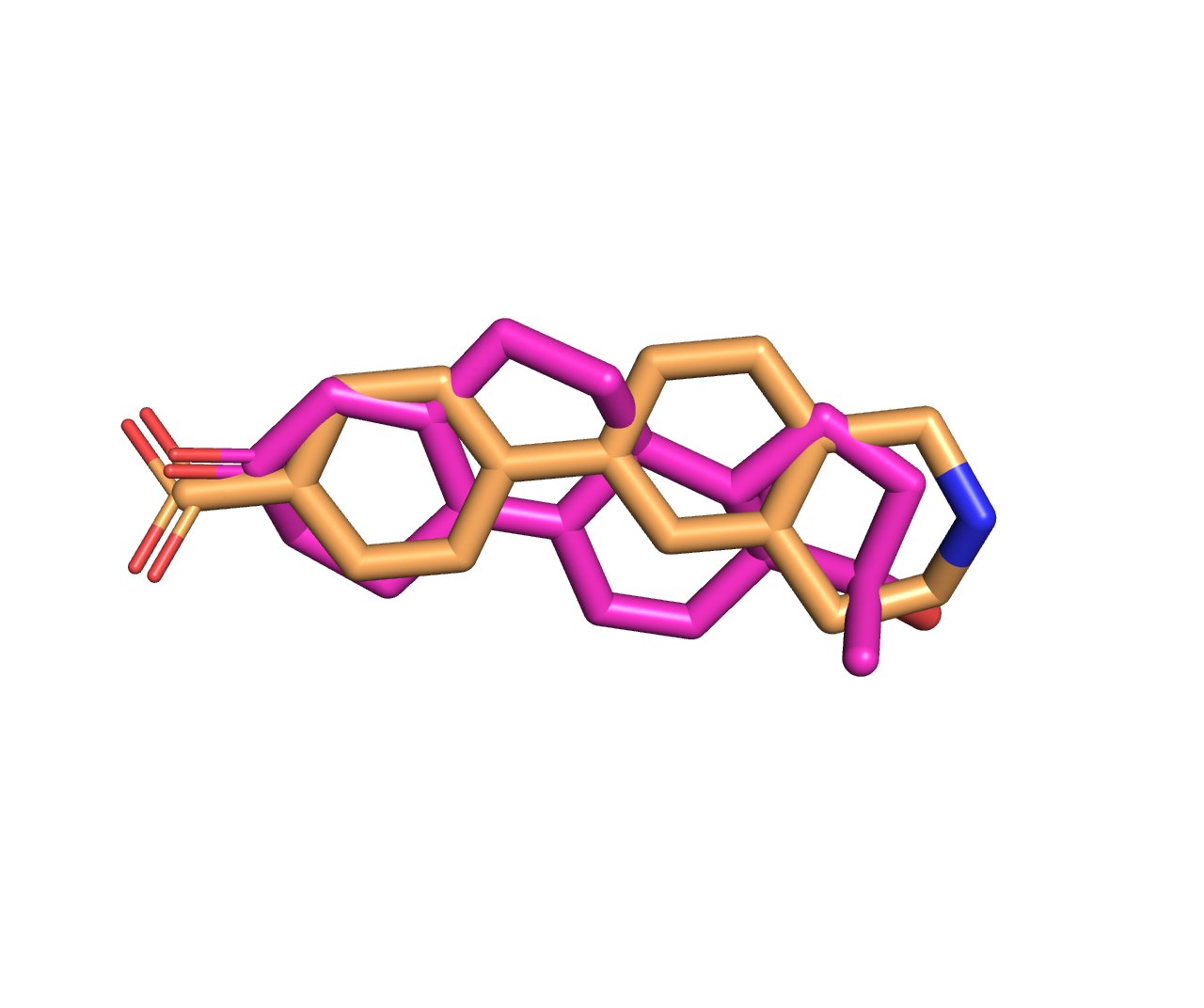}    
        \caption{}
        \label{fig:pos_case_docking:comp} 
    \end{subfigure}
    \vspace{-8pt}
    \caption{(a) Docking of NL-1 within the AR/NR3C4 pocket, with hydrogen bonds shown as solid blue lines. (b) NL-1 superpositioned with the known ligand for the AR/NR3C4 pocket. Orange denotes NL-1; Pink denotes the known ligand.}
    \vspace{-15pt}
\end{figure}

\subsection{Ablation Study}
\label{sec:ablation_study}

We perform additional experiments to test the effectiveness of \agent. 
First, we replace Claude 3.5 Sonnet with DeepSeek-R1 in all components requiring large language models to test \agent's robustness to different backend LLMs.
Additionally, we compare \agent to a simple deterministic loop iterating between \agent's components to analyze the importance of reasoning in \agent. 
Note that this deterministic loop is similar to \agent but without any LLM. 
Concretely, we run \generate once,
followed by a loop of \optimize and \code for $k$ number of times. 
We prioritize properties that fall below requirements when optimizing molecules, and only \code for high-quality molecules.
We set $k$ to 10, same as in Section~\ref{sec:experiments:implementation}.
We show some results in Figure~\ref{fig:ablation} and the full results in the Appendix (Table~\ref{tab:complete})

\textbf{\textit{Reasoning is critical for successful drug discovery}}. 
\agent with reasoning (both Claude 3.5 and DeepSeek-R1) achieves a much higher target success rate than without (more than 40\% absolute difference), indicating its significance. 

\textbf{\textit{\agent is robust to different backend LLMs.}}
Comparing \agent with Claude 3.5 and DeepSeek-R1, they both perform similarly (both with 73\% \successrate), emphasizing that our framework is robust to different backend LLMs.

Interestingly, molecules generated by \agent with DeepSeek-R1 are almost always 
\HQ compared to Claude (\textgreater99\% vs 84\%).
However, only 77\% satisfy the diversity requirements, in contrast to Claude 3.5 (90\%).

\subsection{Case Study on AR/NR3C4}
\label{sec:results:case_study_compounds}

We task \agent with discovering new potential drug therapies targeting 
androgen receptor (AR/NR3C4),
a hormone-driven transcription factor protein that plays a key role in both prostate and breast cancers~\cite{tan_androgen_2015, giovannelli_androgen_2018}.
\agent identifies one molecule (named NL-1) with better \qedreq, \vinareq, and \sascorereq than the ligand and at least one approved drug (e.g., Fulmatide) for the respective targets.
They are illustrated in Figure~\ref{fig:ar}.
NL-1 has several desirable traits, such as zwitterionic (with positive and negative charged atoms on the respective ends)---a trait typical in most biological molecules and drugs~\cite{mobitz2024nonclassical}.
The molecule also passes several computational filters, including PAINS~\cite{baell2010new}, BRENK~\cite{brenk2008lessons}, NIH~\cite{jadhav2010quantitative, doveston2015unified}, Lilly~\cite{bruns2012rules}, and Lipinski~\cite{lipinski_experimental_2001}, further highlighting its attractiveness as a drug.
In terms of binding, the molecule has -8.81 kcal/mol for \vinareq, emphasizing that it can bind well to the pocket.
Notably, 
Figure~\ref{fig:pos_case_docking:ar} shows that
the molecule is buried deep within the pocket, surrounded almost entirely by hydrophobic residues providing many van der Waals contacts.
The molecule’s carboxylic acid group also engages in hydrogen bonding at one end of the pocket, further stabilizing the complex and contributing to binding affinity.
Encouragingly, further evaluation reveals that NL-1 has several established synthetic routes and demonstrates precedent for antagonizing stimulator of interferon genes (STING)~\cite{gulen2023cgas}, thereby useful in the treatment of inflammatory diseases~\cite{li_antagonists_2023}.
The desirable traits, combined with its synthetic accessibility and therapeutic precedents, position the molecule as a promising candidate for the androgen receptor. 

Figure~\ref{fig:pos_case_docking:comp} shows a comparison of NL-1 to the known ligand.
Both molecules feature a hydrophobic core with polar anchors on either end, but NL-1 is slightly less compact, with fewer fused rings than the known ligand.
This reduces conformational rigidity, providing slightly increased flexibility to adapt to the binding pocket.
Furthermore, NL-1 anchors itself with a carboxylic acid in place of the known ligand's ketone, enabling more hydrogen bonds.

We also present another case study on EGFR and extensively discuss the results in Appendix~\ref{sec:appendix:results:case}.

\section{Conclusions}
\label{sec:conclusion}

We present \agent, an agent for autonomous drug discovery. We comprehensively examine its capabilities, demonstrating its performance across many major therapeutical targets and revealing several key insights on its success. Furthermore, we investigate the generated molecules on the highly critical target EGFR and show their potential as drug candidates.

\section{Acknowledgements}
\label{sec:acknowledgement}
This project was made possible, in part, by support from the National Science Foundation
grant nos. IIS-2133650 and IIS-2435819, and the National Library of Medicine grant no.
1R01LM014385. Any opinions, findings and conclusions or recommendations expressed in
this paper are those of the authors and do not necessarily reflect the views of the funding
agency. In addition, we would like to thank our colleagues, Ruoxi Gao, for generating some
of the data and figures and Xinyi Ling for generating some of the figures in this work.

\section{Limitations}
\label{sec:limitation}

\agent is not without its limitations, and we plan to address them in future work. 
(1) We aim to show the utility of LLM in navigating drug discovery \textit{in silico}, and thus, we solely focus on \textit{in silico} evaluations. 
However, \textit{in silico} is only a part of the entire drug discovery pipeline.
To test \agent's efficacy in the clinical world, we plan to add wet-lab validation in our follow-up research.
(2) We focus our evaluation on a few key pharmaceutical properties without undermining others in our current work. We acknowledge that drug discovery requires much more than just a few key properties.
The goal of this paper is not to replicate the entire drug discovery process but to demonstrate the strong potential of agents in facilitating drug discovery through generating and optimizing new drug candidates over a few essential properties.
To this end, we intentionally design \agent to be easily extendable to other metrics.
We will continue developing the agent to cover more properties in our future research.
(3) Our experiments with \agent were built on a limited number of API calls due to budget constraints. Further testing on more API calls will be an interesting avenue for future research.
(4) Finally, we benchmark \agent on a small set of targets given the lack of well-established, large-scale, well-annotated benchmarks for our tasks.
We prioritized a few highly clinically relevant therapeutic targets (such as cancer) with detailed information about their structures, ligands, and binding affinities.
This serves as a trade-off between scope (lack of benchmarks) and substance (focusing on clinically relevant targets).
This follows the example of related LLM agent works~\cite{boiko_autonomous_2023, m_bran_augmenting_2024}, which have used similarly small benchmarks for initial demonstrations of agent capabilities on complex tasks.
Future work will include an expansion of this dataset and additional benchmarking.


\section*{Ethics Statement}
\label{sec:ethic}

\agent is designed to generate small 
molecules meeting the parameters specified in a natural language prompt. However, 
we recognize that not all small molecules are safe,
and such a tool could generate harmful molecules.
As such, we have taken several steps to minimize the potential negative impact.
\agent only functions \emph{in silico} and does not currently generate synthesis plans for any of its molecules.
This prevents any dangerous molecules from being automatically produced without conscious human oversight or intervention.
Additionally, our \evaluator implementation and example prompts focus \agent on priority metrics in designing drugs to benefit humans, such as \qedreq and \lipinskireq.

Despite these efforts, we cannot guarantee that \agent will not generate harmful or incorrect content.
We encourage users to practice discretion when using \agent,
and to follow all applicable safety guidelines and research best practices.

\bibliography{custom}

\clearpage
\onecolumn
\appendix

\renewcommand{\thefigure}{A\arabic{figure}} 
\setcounter{figure}{0} 
\renewcommand{\thetable}{A\arabic{table}} 
\setcounter{table}{0} 

\section{Dataset}
\label{sec:appendix:dataset}

The following table details each protein target in the dataset.  The disease categories column indicates whether the target is associated with autoimmune disease (A), cancer (CA), cardiovascular disease (CV), Diabetes (D), Infectious Disease (I), or Neurological Conditions (N).

\begin{ThreePartTable}
\renewcommand\TPTminimum{\textwidth}
\setlength\LTleft{0pt}
\setlength\LTright{0pt}
\setlength\tabcolsep{0pt}

\begin{longtable}{ l @{\extracolsep{\fill}} *{5}{c} }
\toprule
Gene Name & PDB ID & Pocket Structure & PDB Lig. & Ligand Structure & Disease Cat. \\ 
\midrule
\endhead

\midrule[\heavyrulewidth]
\multicolumn{6}{r}{\textit{continued}}\\
\endfoot  

\midrule[\heavyrulewidth]
\endlastfoot

ADRB1                                 & 7BU7            &   \includegraphics[width=4cm]{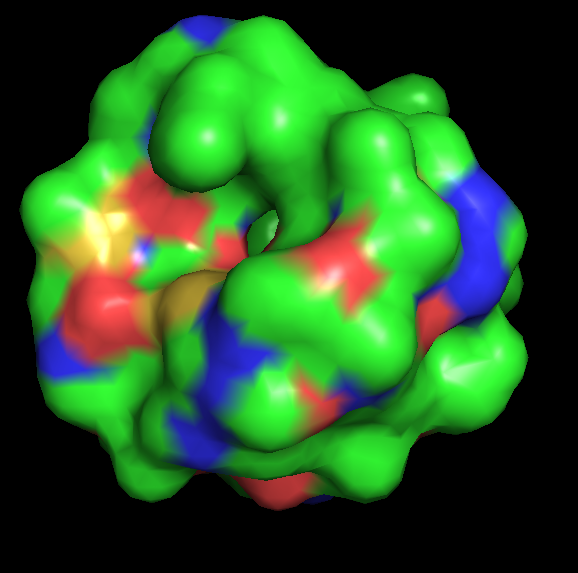}                        & P0G                    & \includegraphics[width=4cm]{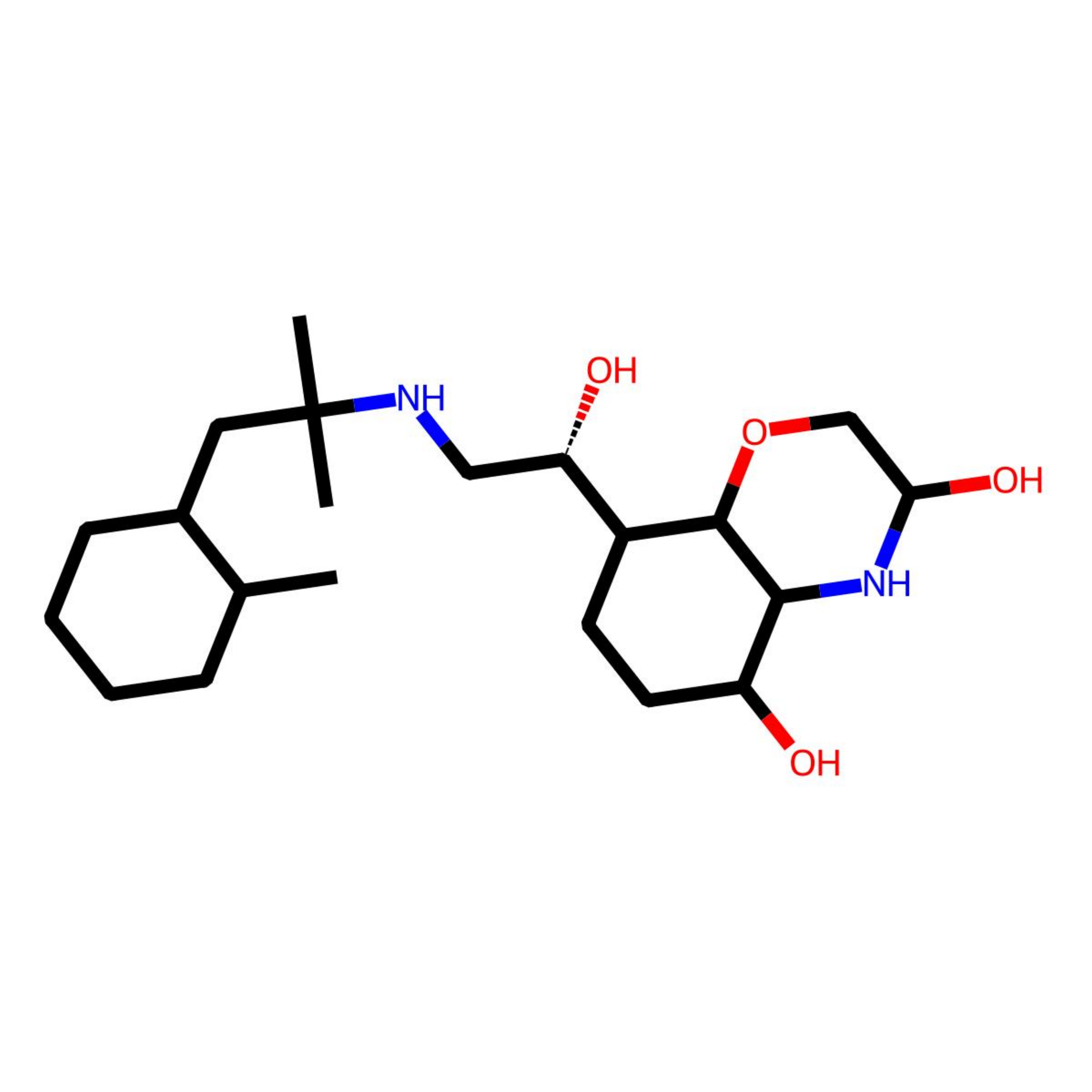}                                                  & CV                              \\
ADRB2                                 & 4LDL            &   \includegraphics[width=4cm]{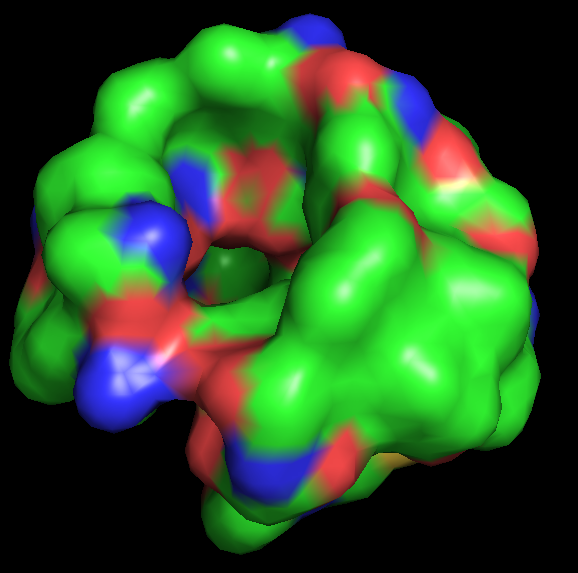}                        & XQC                    &\includegraphics[width=4cm]{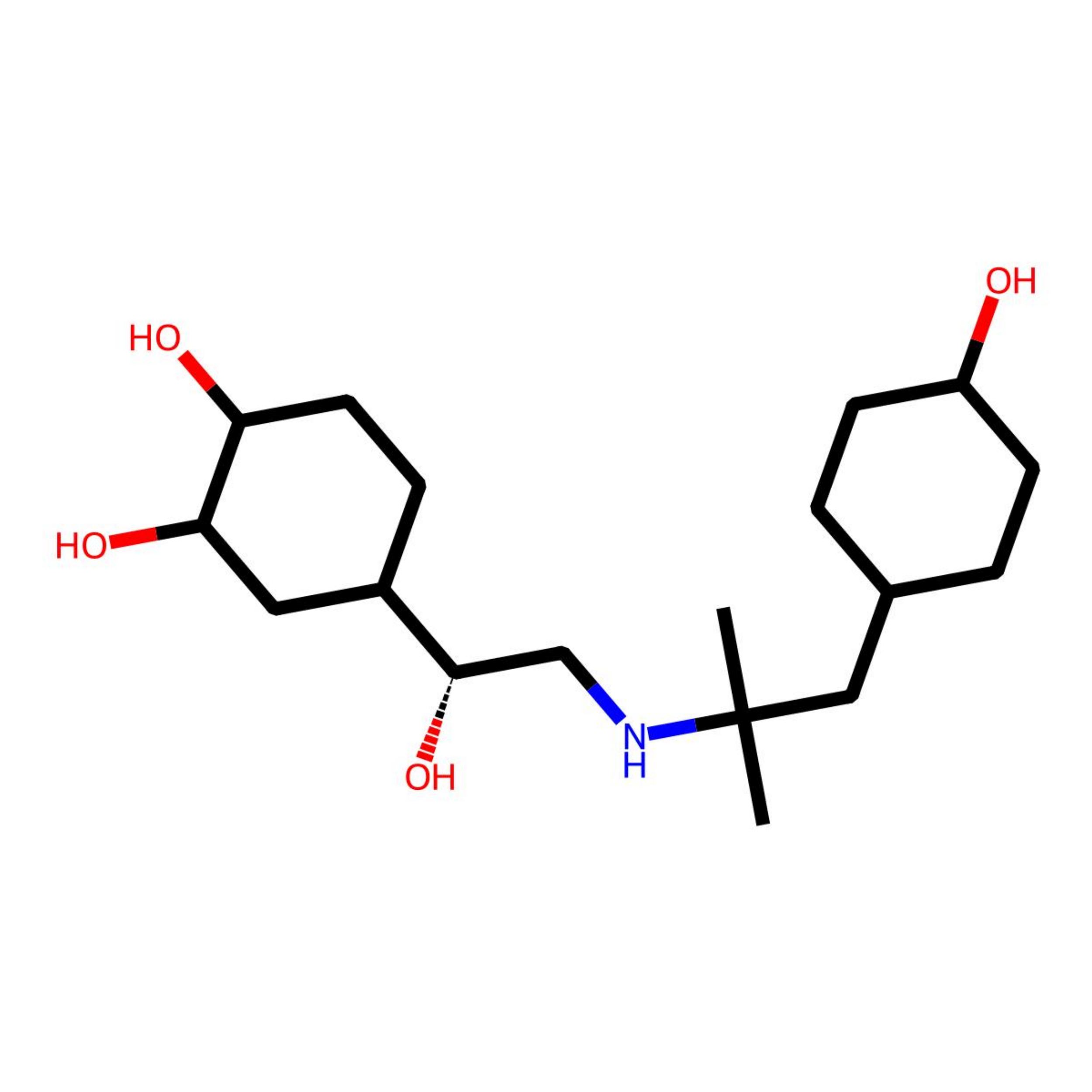}                                                  & CV                              \\
AR (NR3C4)                            & 1E3G            &    \includegraphics[width=4cm]{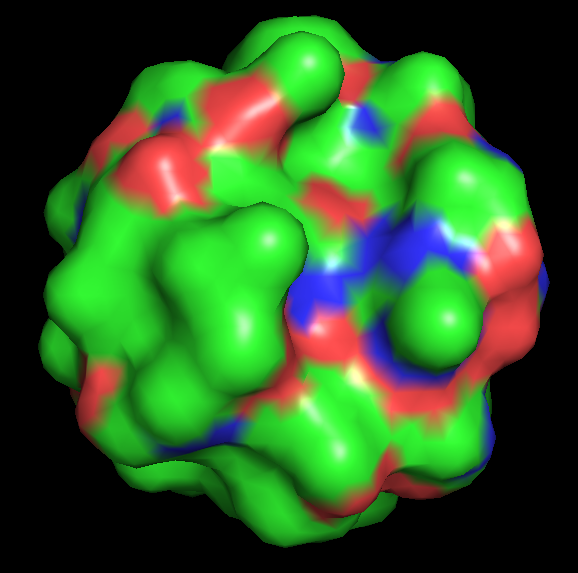}                       & R18                    &\includegraphics[width=4cm]{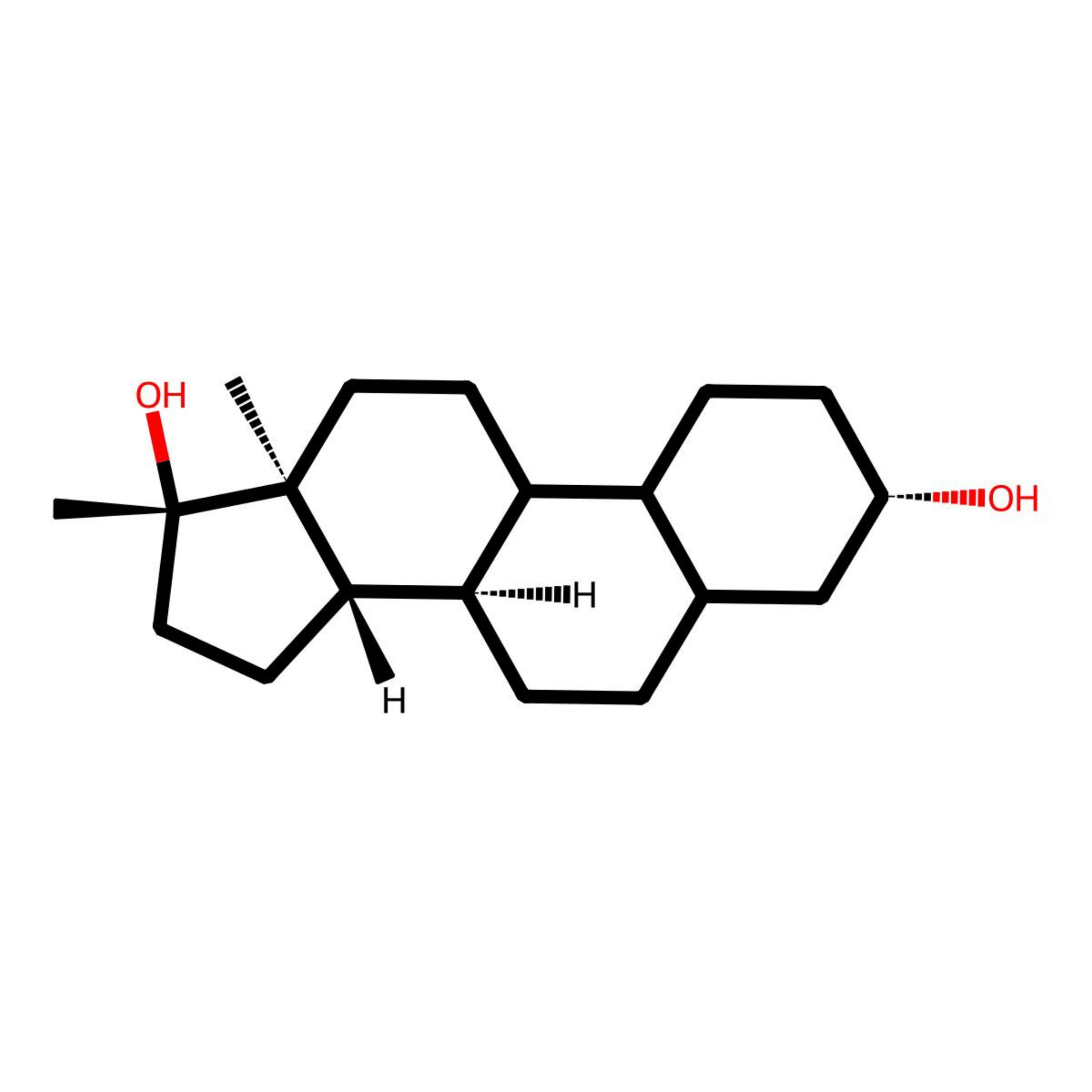}                                                 & CA                                              \\
BCHE                                  & 4TPK            &    \includegraphics[width=4cm]{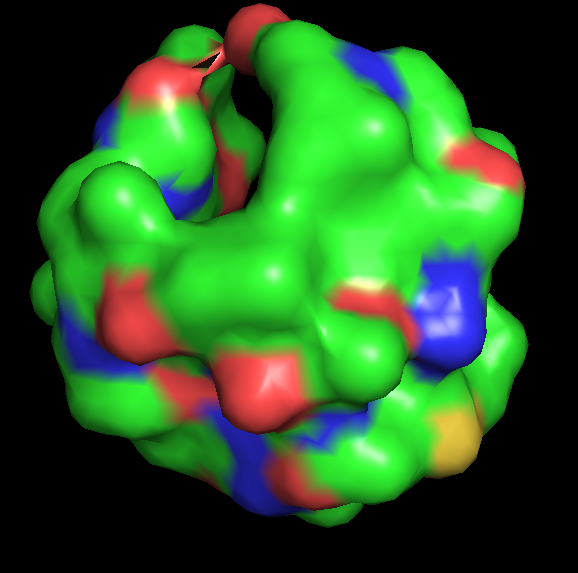}                       & 3F9                    &\includegraphics[width=4cm]{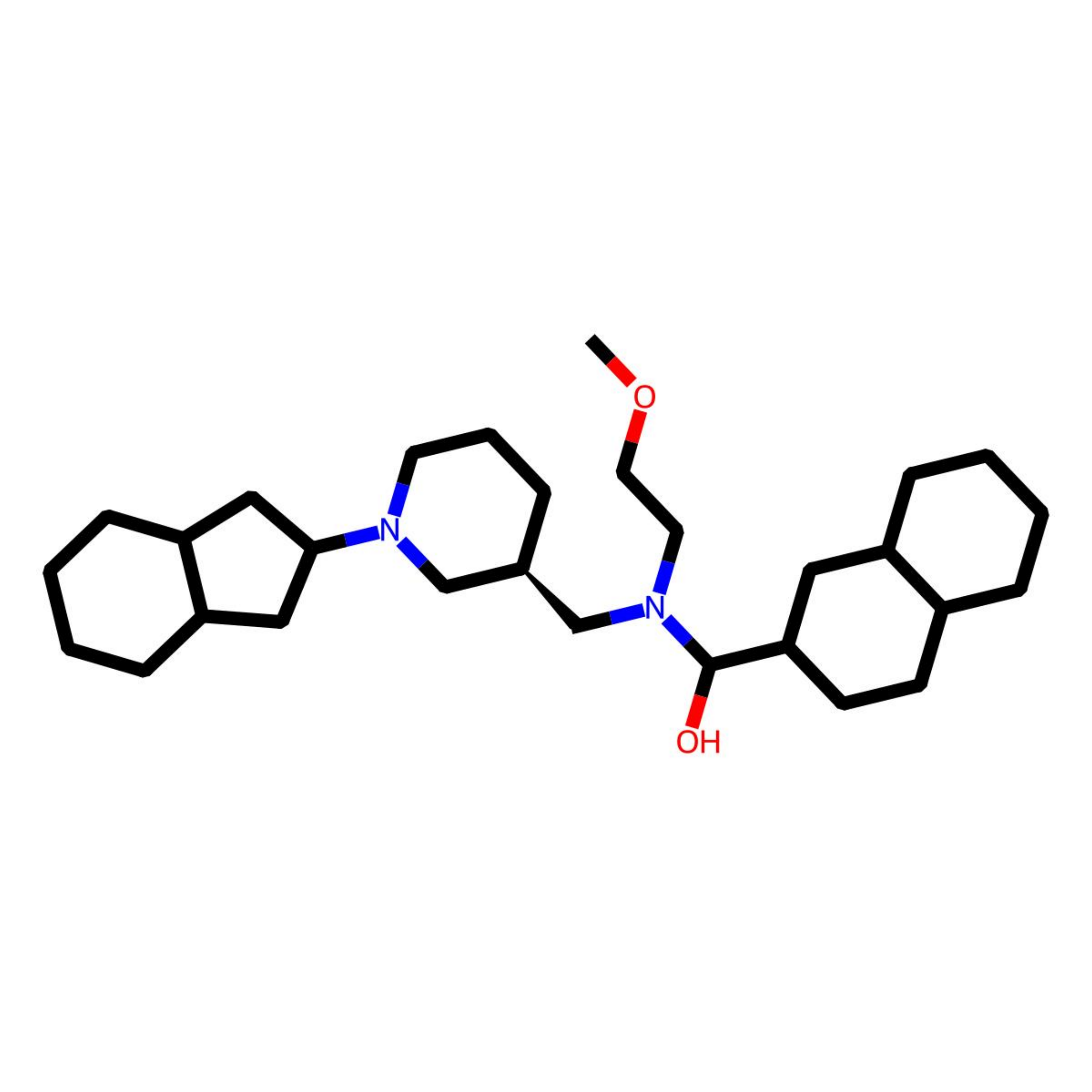}                                                 & N                             \\
CDK5                                  & 1UNG            &     \includegraphics[width=4cm]{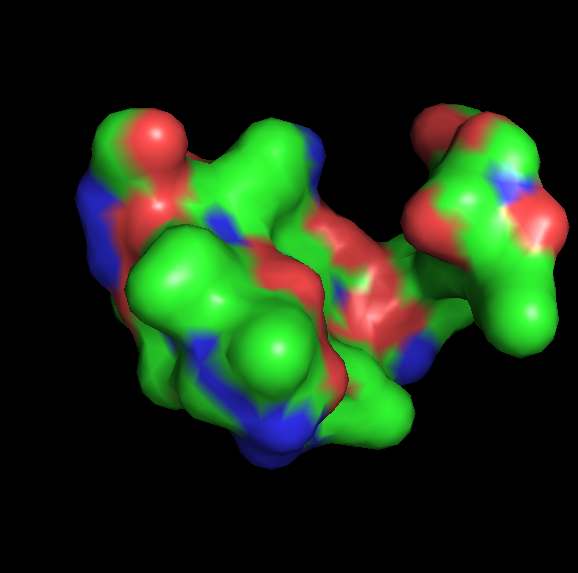}                      & ALH                    &\includegraphics[width=4cm]{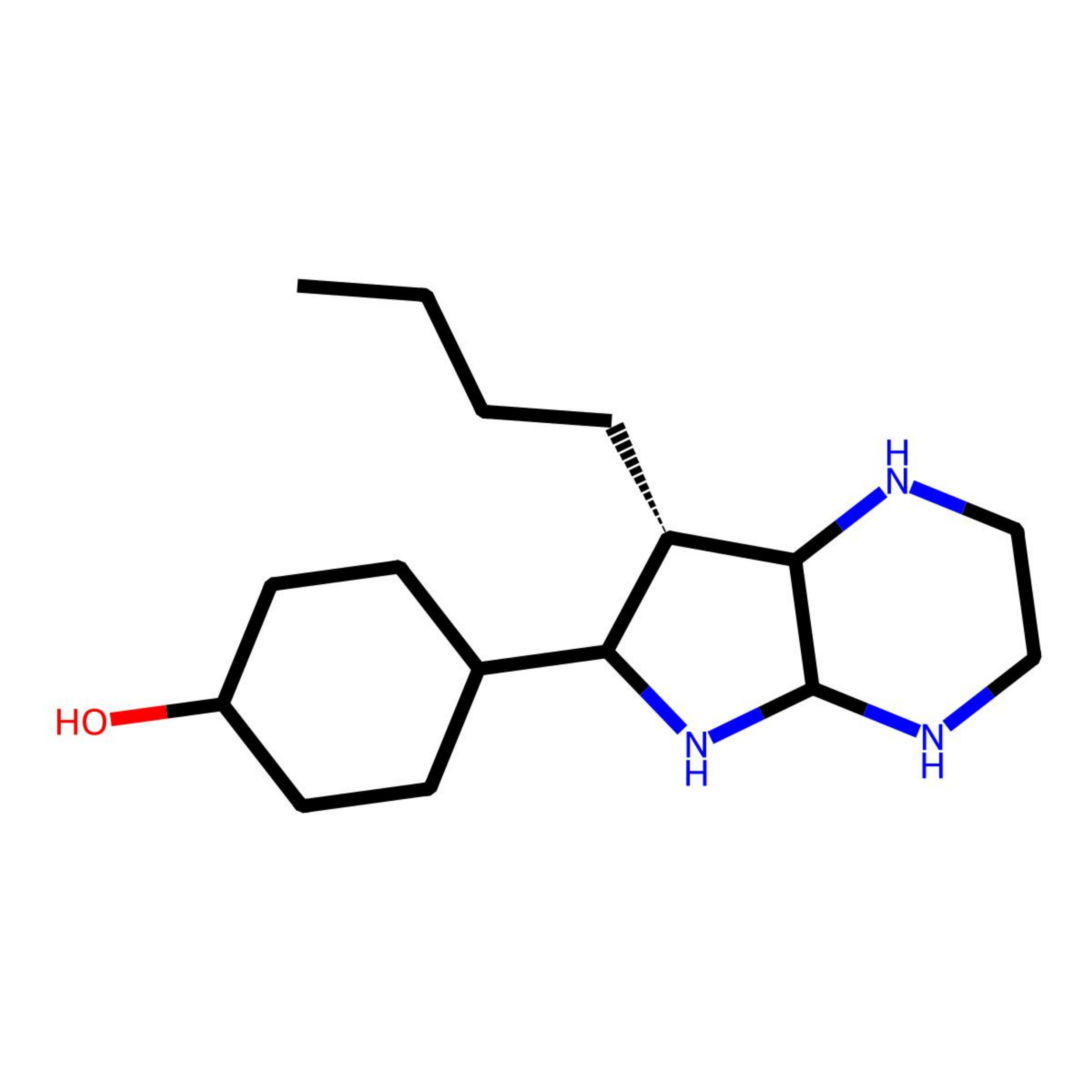}                                                & N                             \\
CHK2                                  & 2W7X            &      \includegraphics[width=4cm]{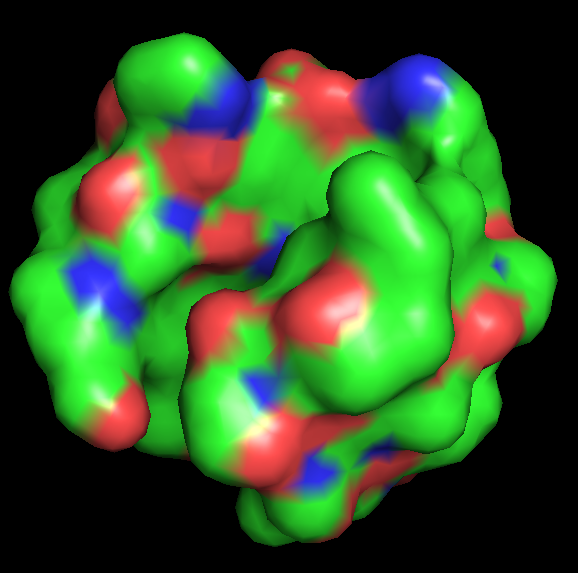}                     & D1A                    &\includegraphics[width=4cm]{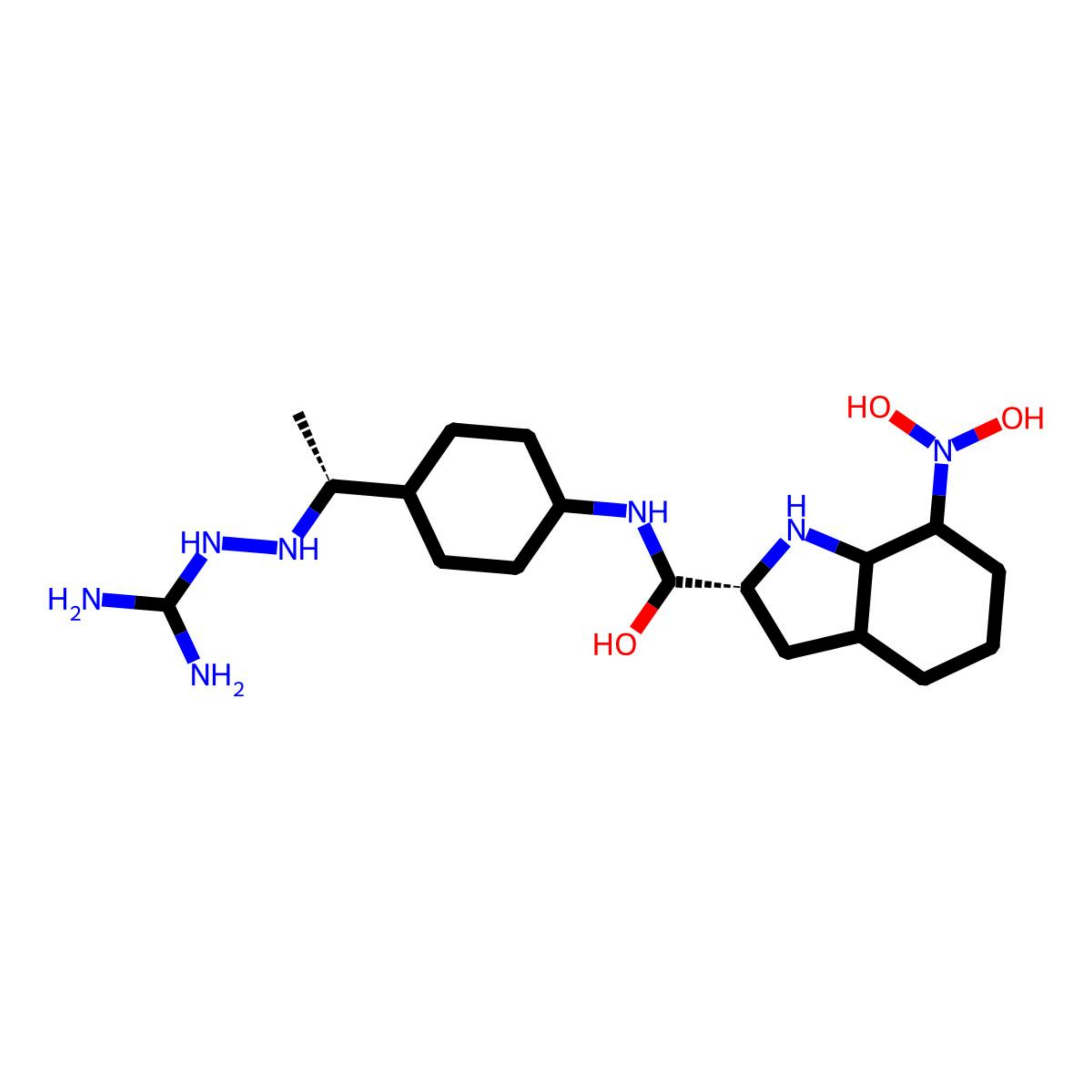}                                               & CA                                              \\
CYP3A4                                & 6MA6            &      \includegraphics[width=4cm]{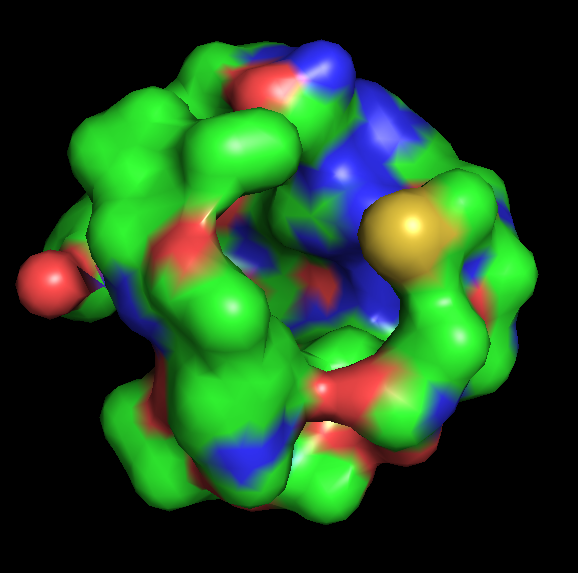}                     & MYT                    &\includegraphics[width=4cm]{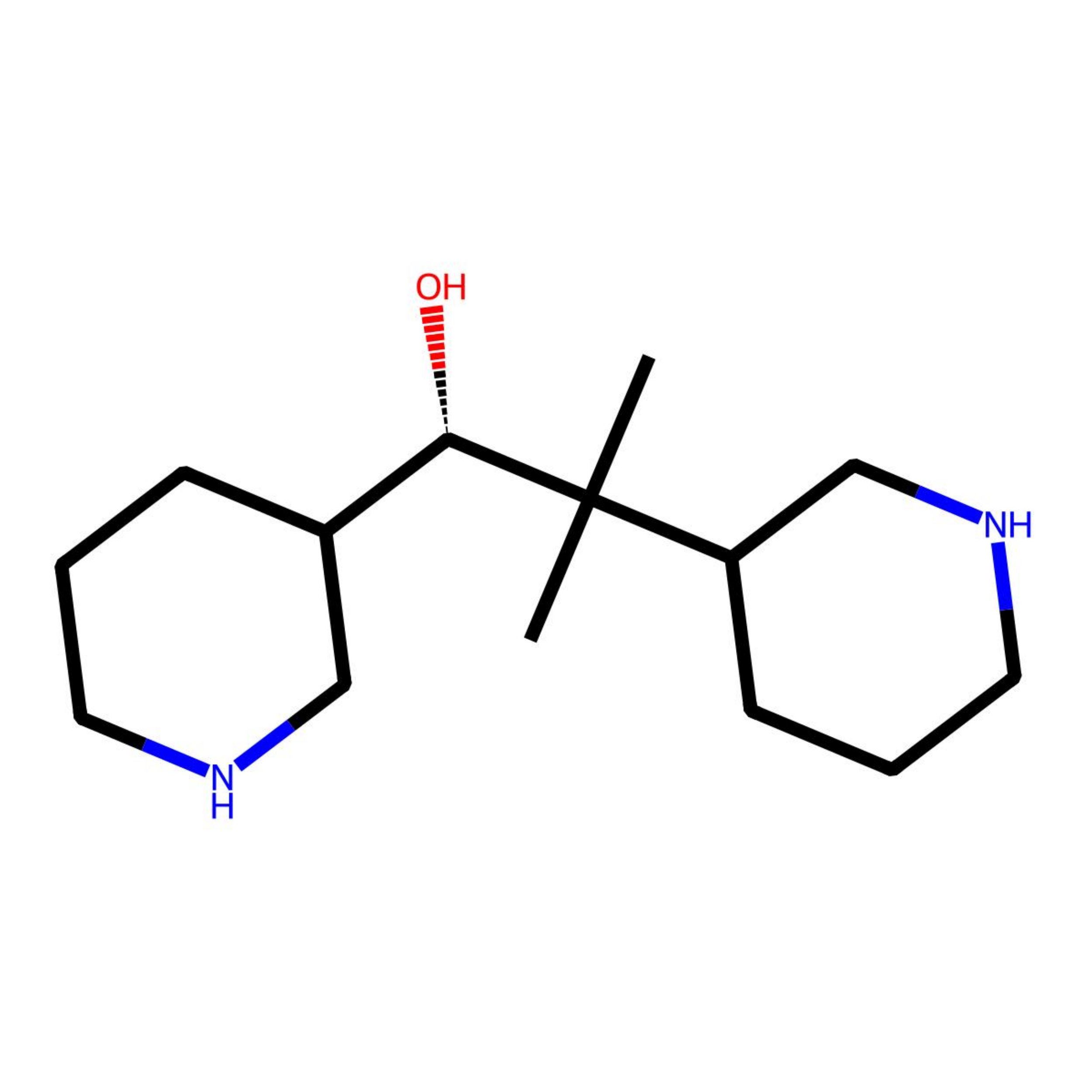}                                               & CA, I                          \\
CYP3A5                                & 7SV2            &      \includegraphics[width=4cm]{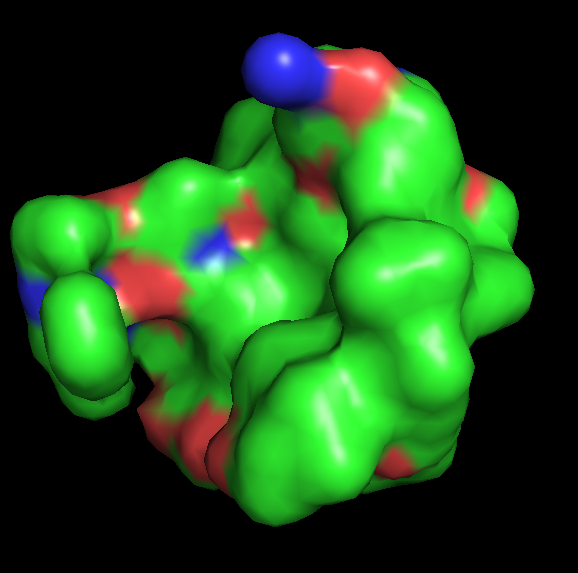}                     & MWY                    &\includegraphics[width=4cm]{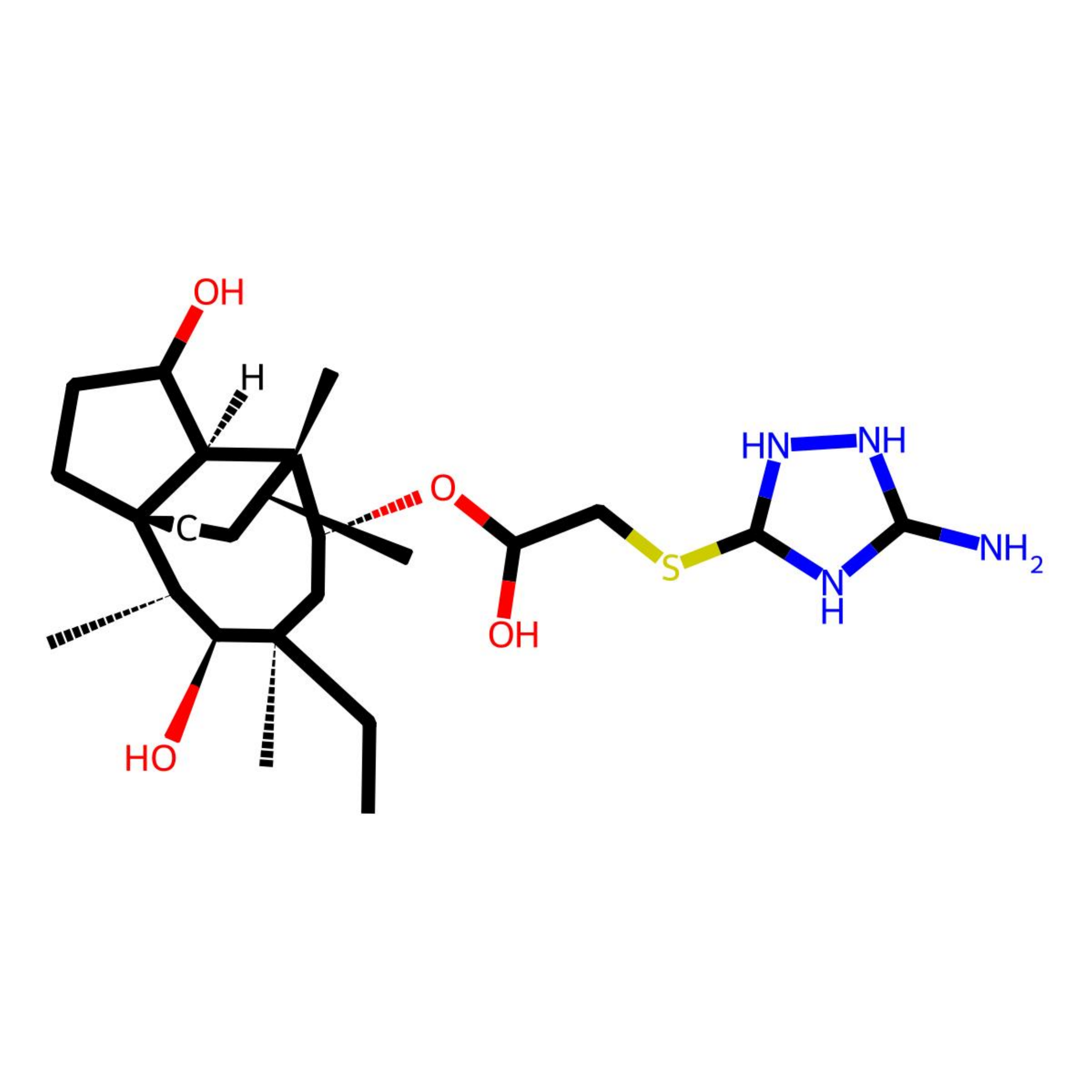}                                               & CA, I                          \\
DRD2                                  & 6CM4            &      \includegraphics[width=4cm]{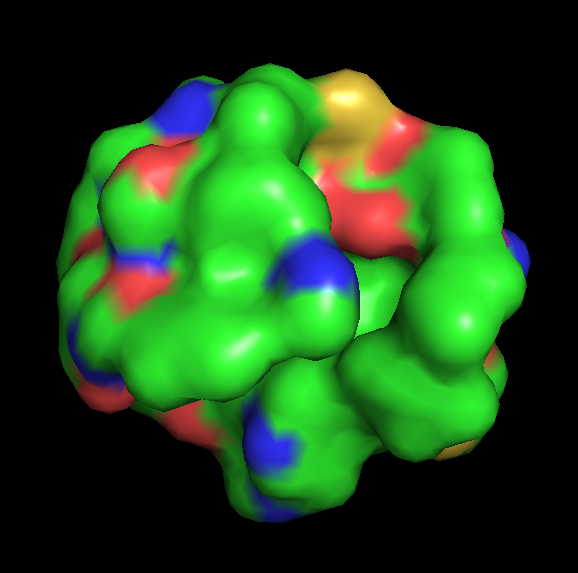}                     & 8NU                    &\includegraphics[width=4cm]{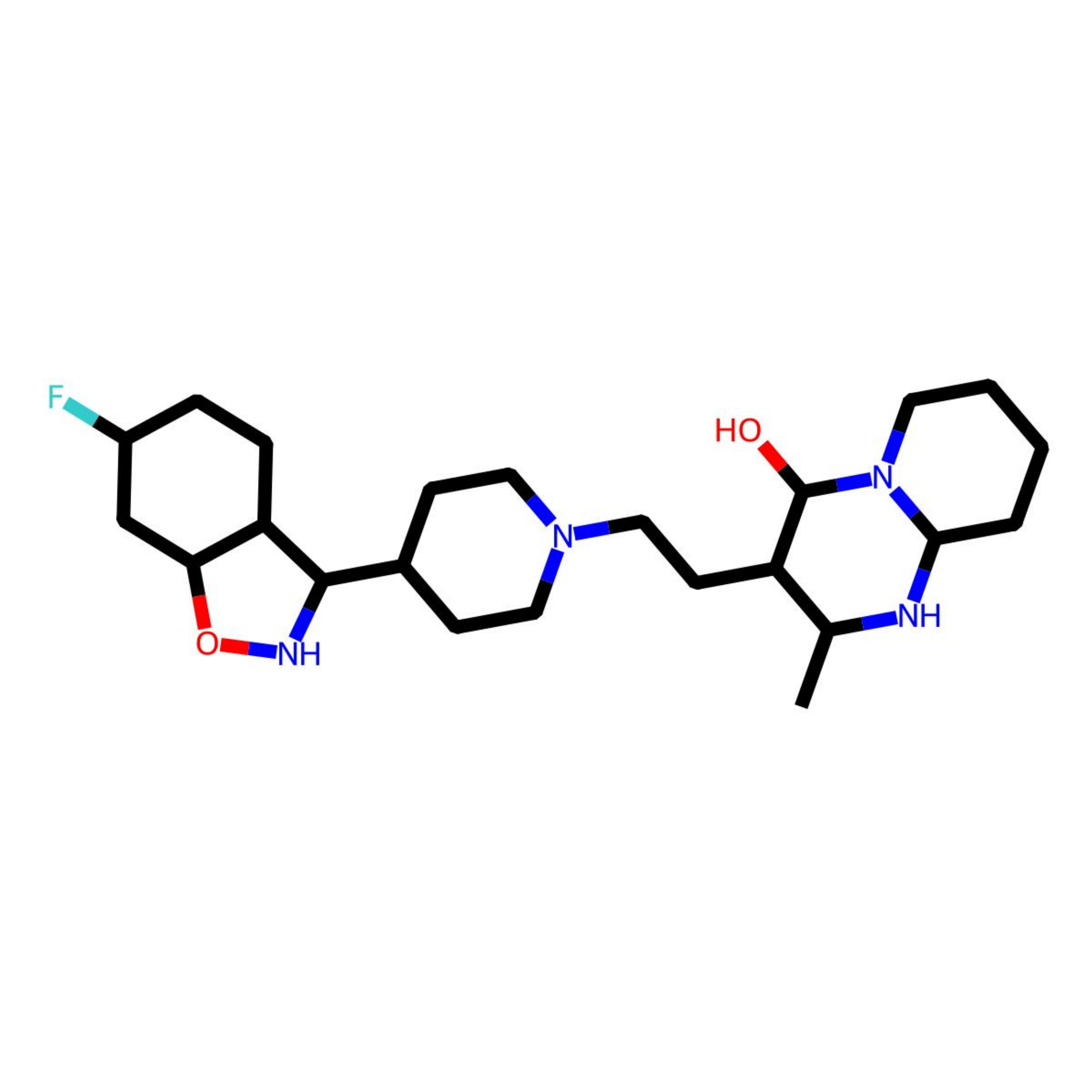}                                               & N                             \\
DRD3                                  & 3PBL            &       \includegraphics[width=4cm]{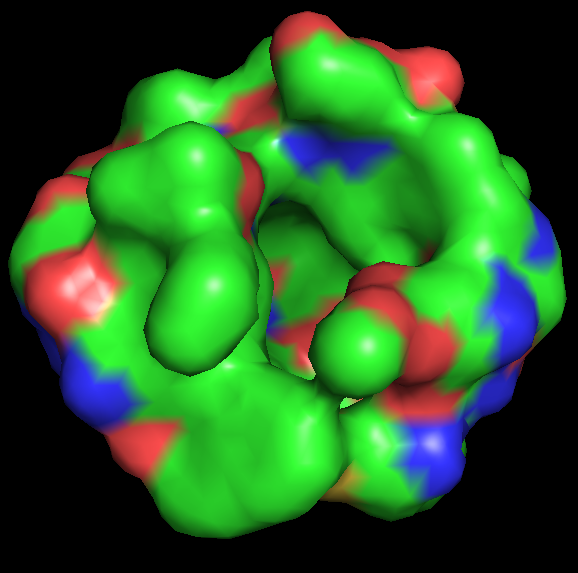}                    & ETQ                    &\includegraphics[width=4cm]{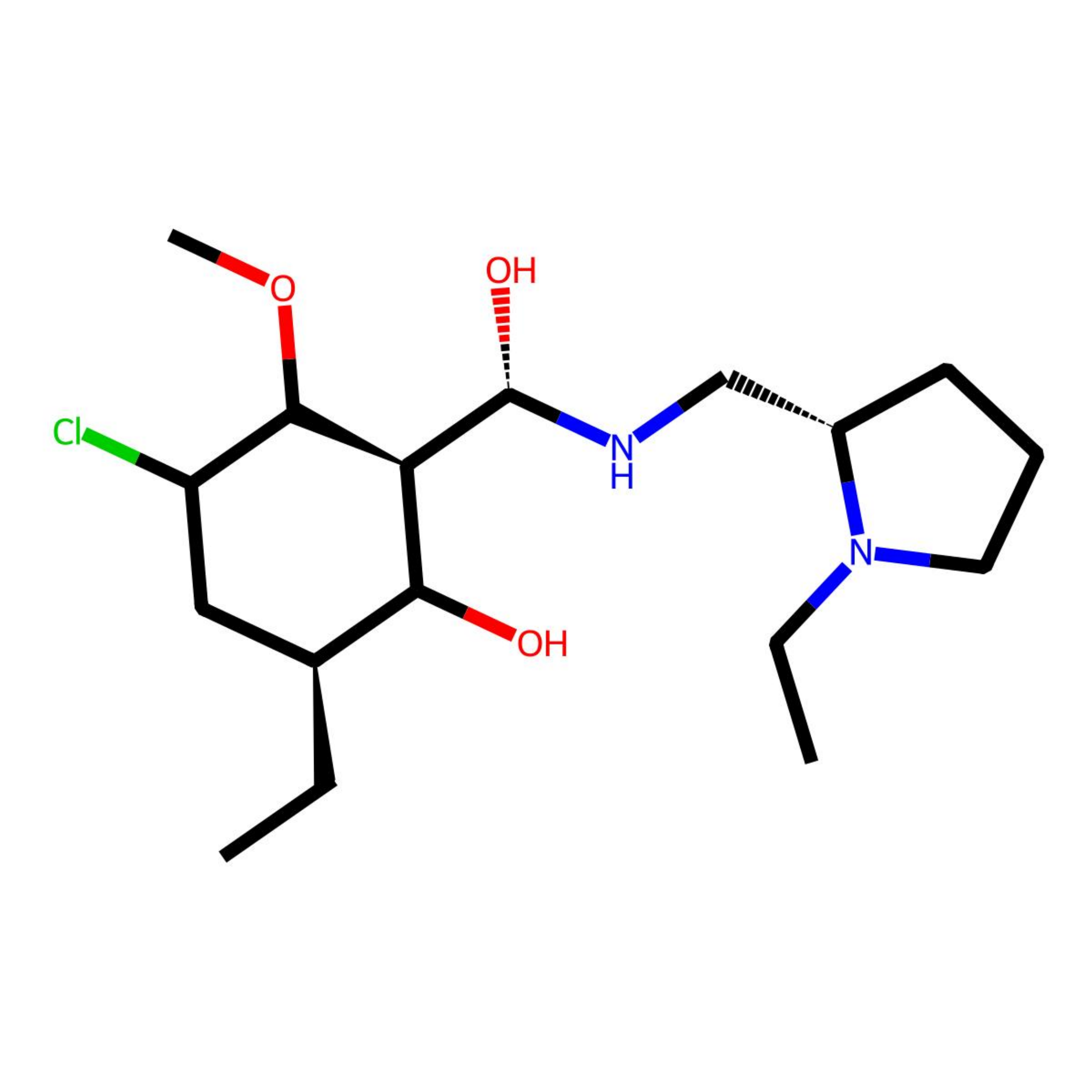}                                              & N                             \\
EGFR                                  & 1M17            &       \includegraphics[width=4cm]{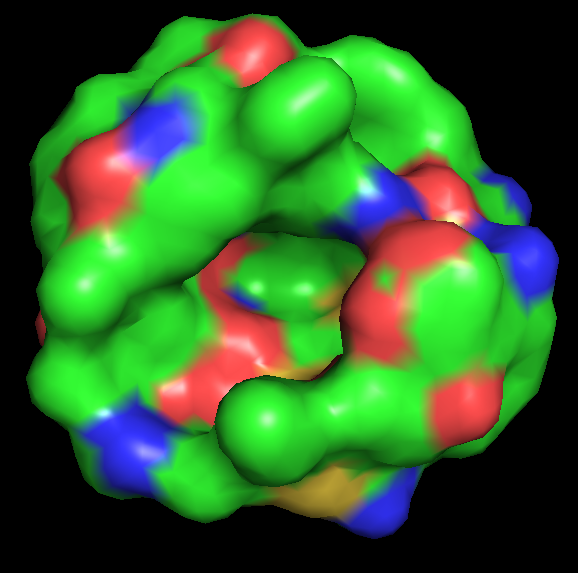}                    & AQ4                    &\includegraphics[width=4cm]{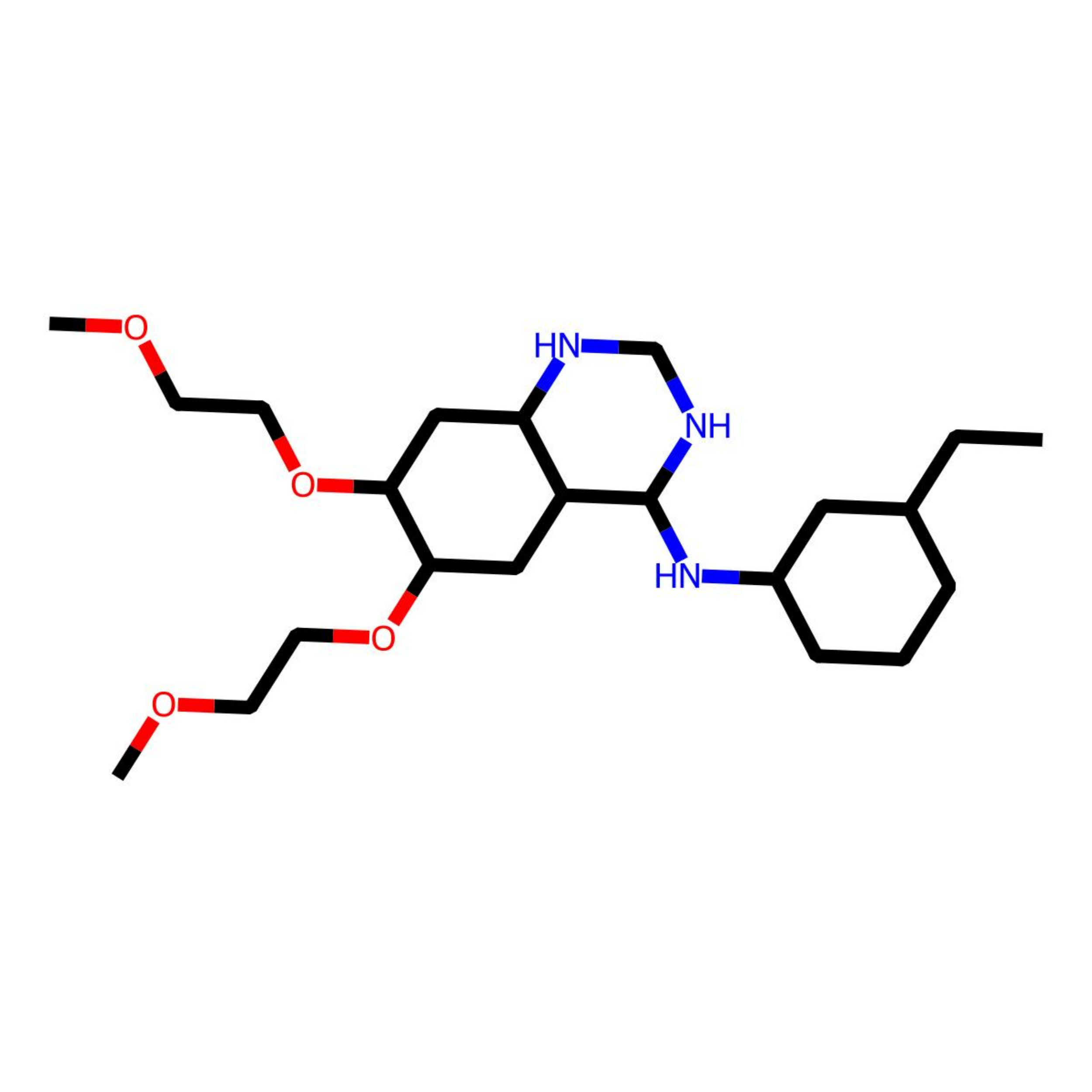}                                              & CA                                              \\
EZH2                                  & 7AT8            &        \includegraphics[width=4cm]{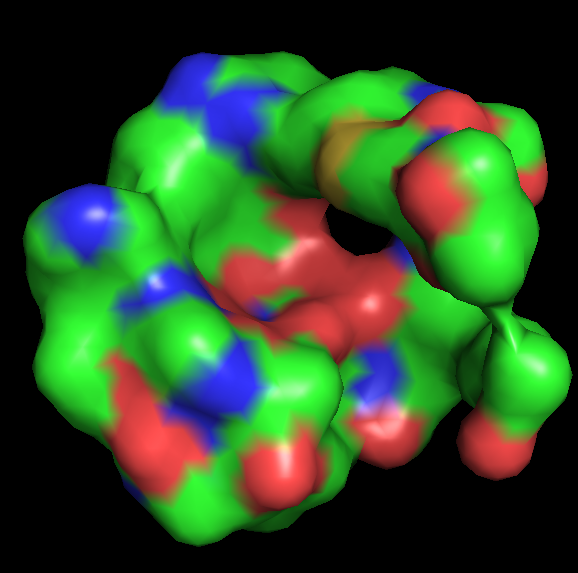}                   & SAH                    &\includegraphics[width=4cm]{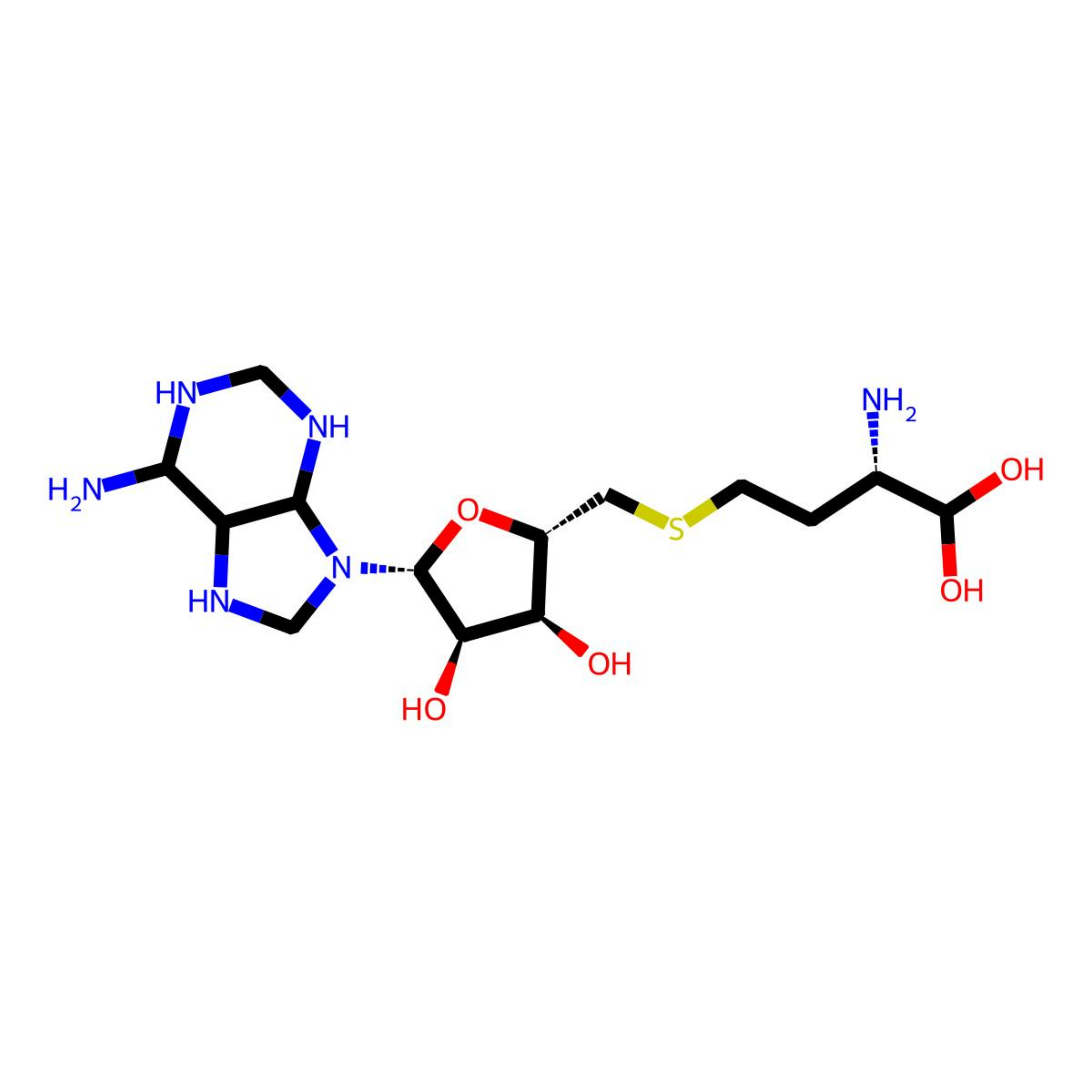}                                             & CA                                              \\
FLT3                                  & 4RT7            &          \includegraphics[width=4cm]{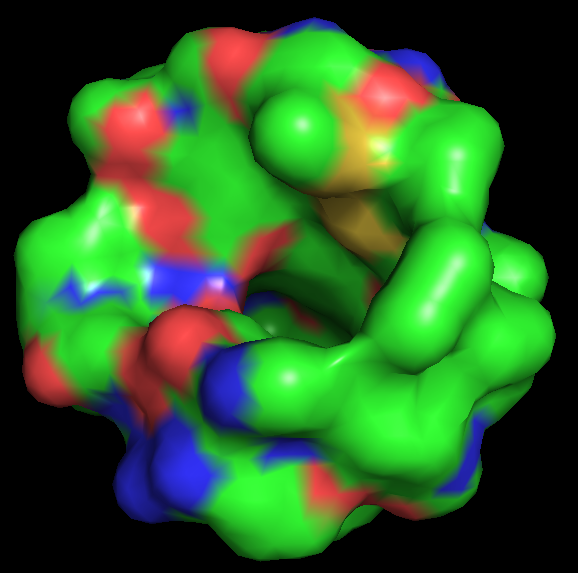}                 & P30                    &\includegraphics[width=4cm]{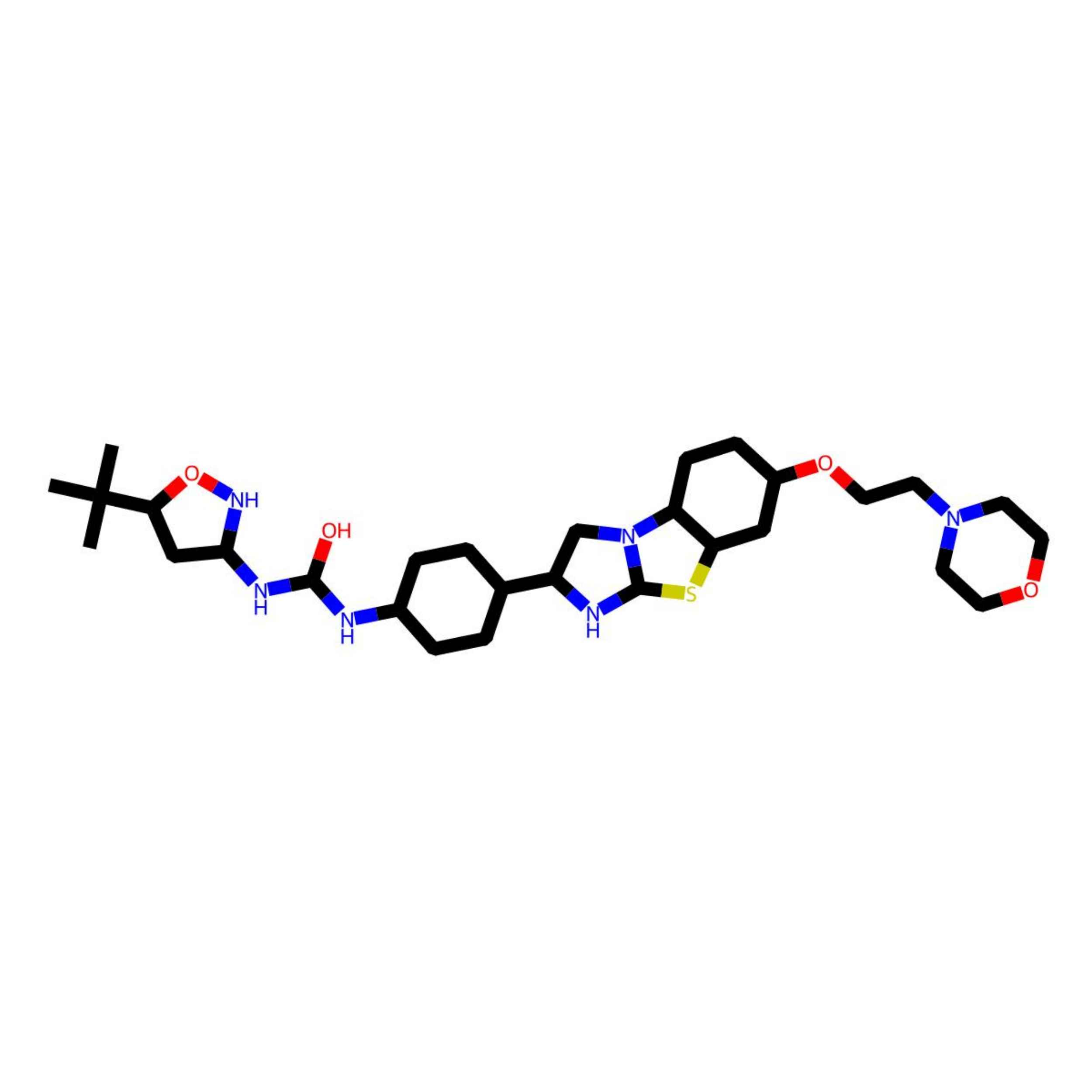}                                           & CA                                              \\
GCK (HK4)                             & 3H1V            &      \includegraphics[width=4cm]{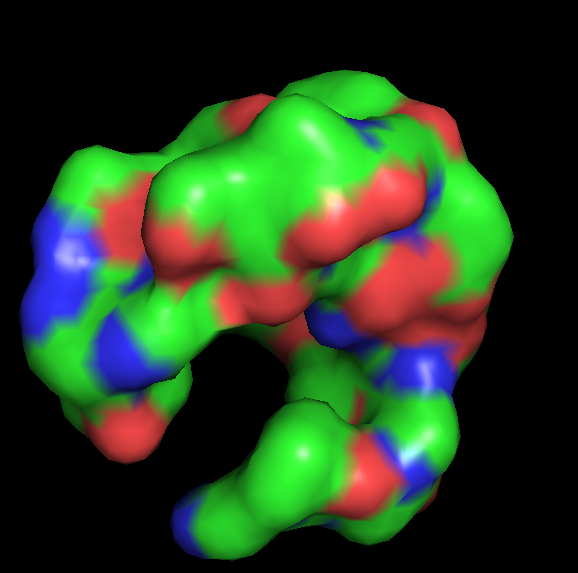}                     & GCK                    &\includegraphics[width=4cm]{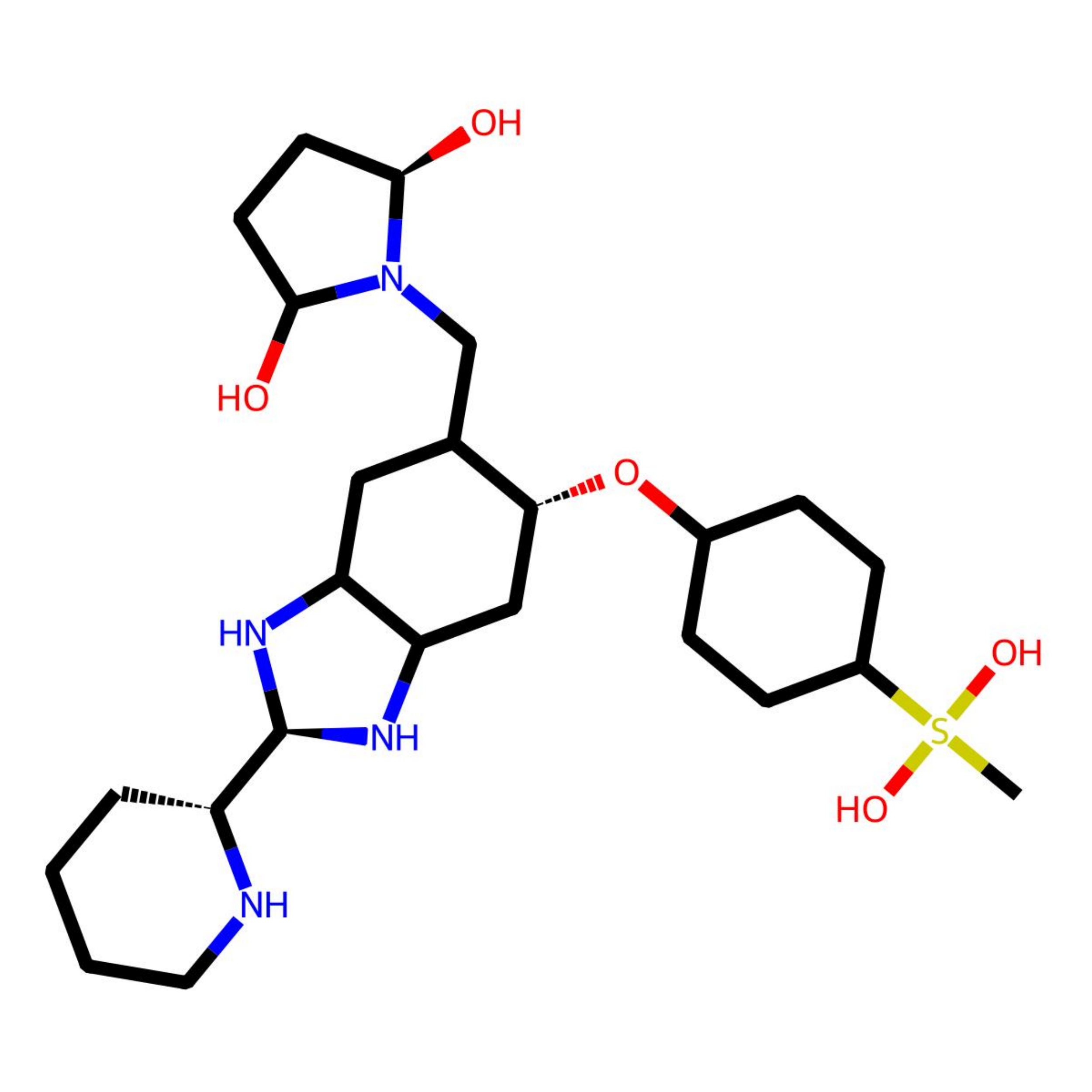}                                               & D                                            \\
GLP1R                                 & 6ORV            &      \includegraphics[width=4cm]{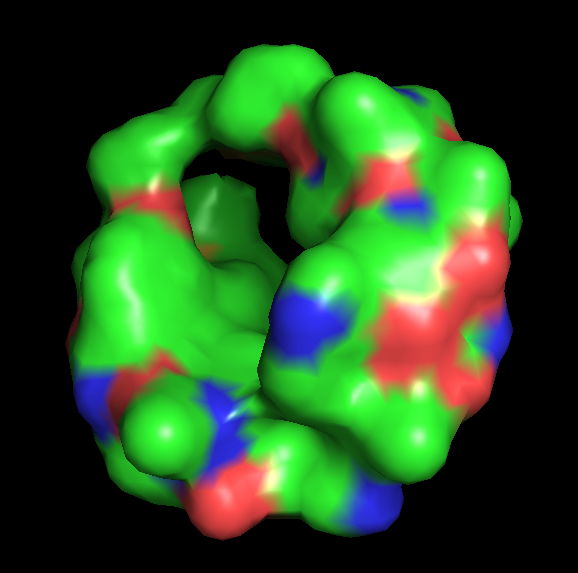}                     & N2V                    &\includegraphics[width=4cm]{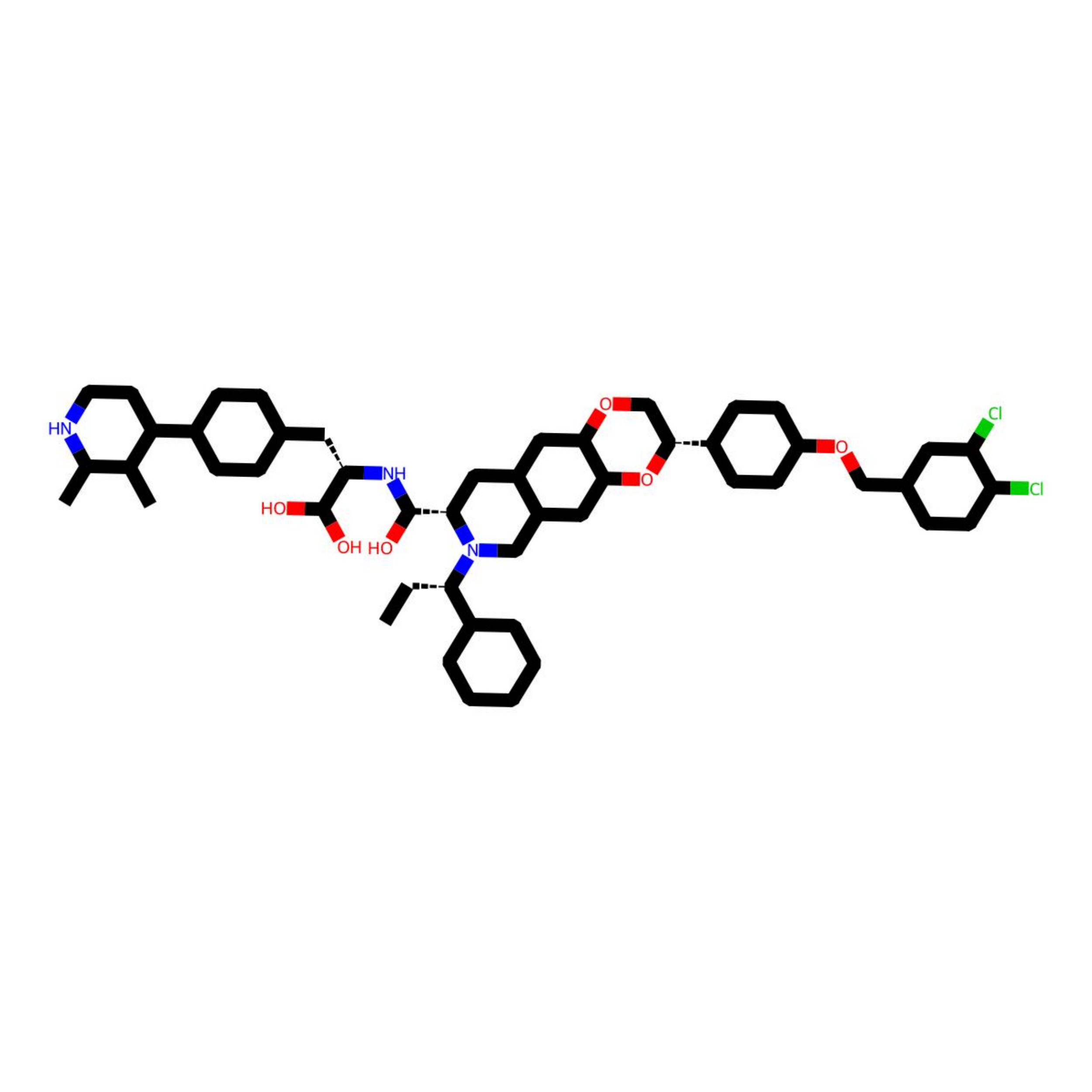}                                               & D                                            \\
HRAS                                  & 1RVD            &       \includegraphics[width=4cm]{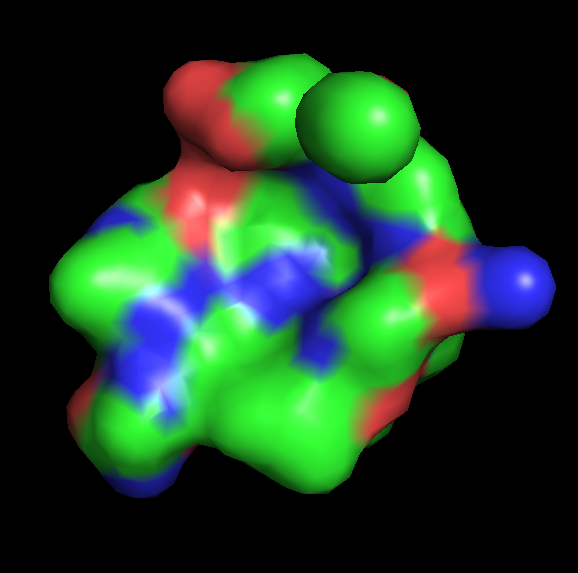}                    & DBG                    &\includegraphics[width=4cm]{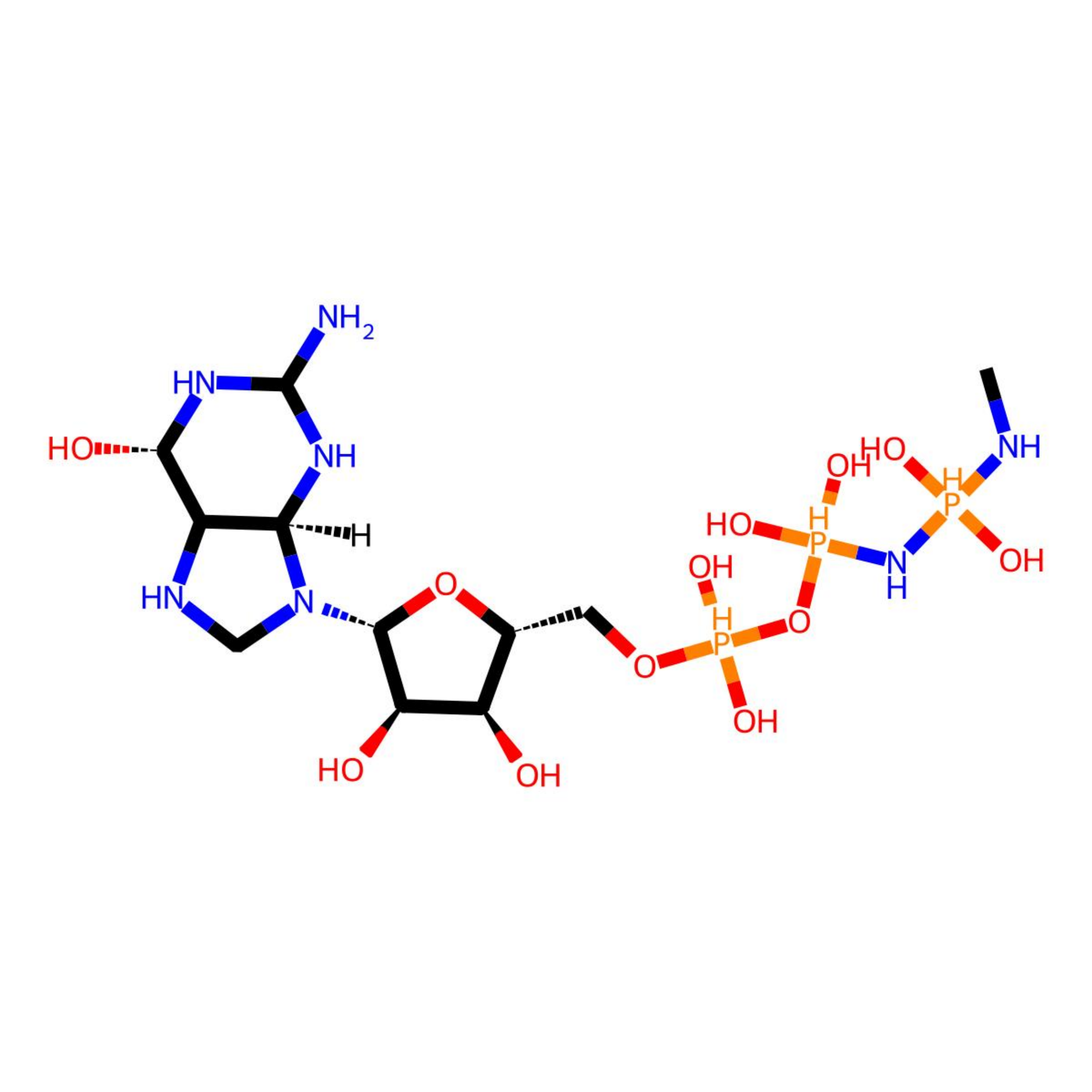}                                              & CA                                              \\
HTR2A                                 & 7WC8            &      \includegraphics[width=4cm]{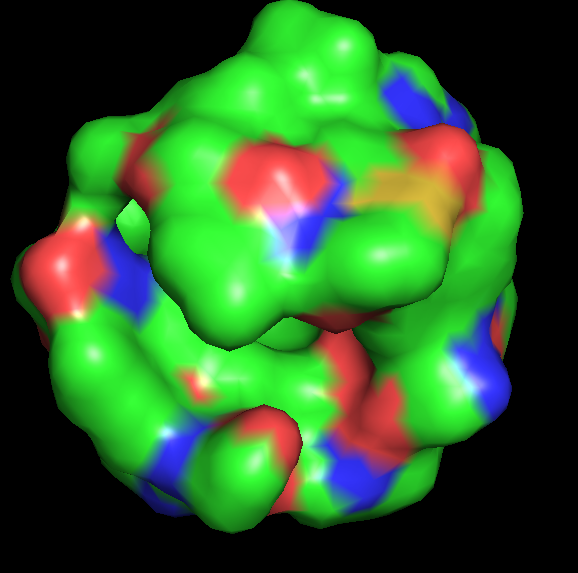}                     & 92S                    &\includegraphics[width=4cm]{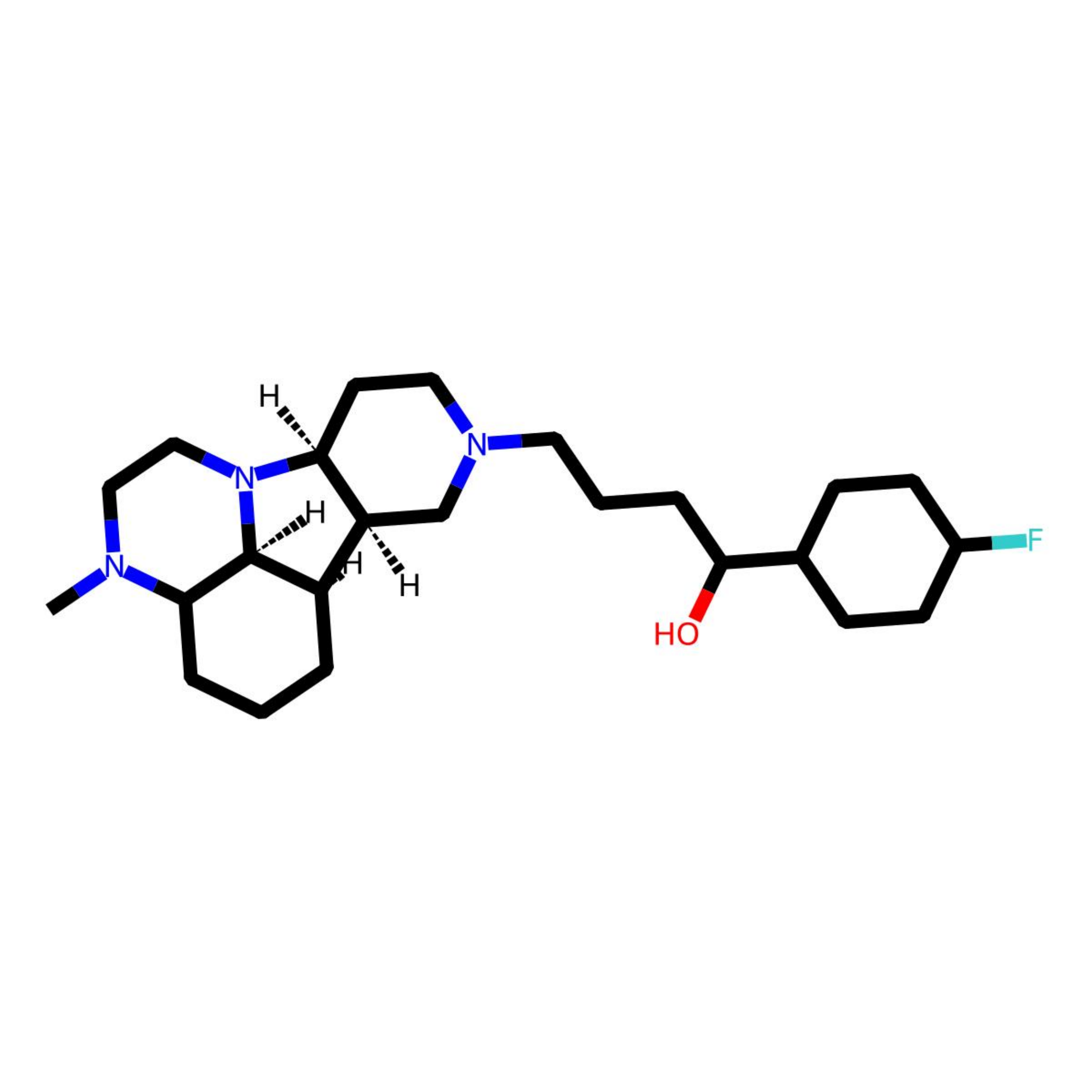}                                               & N                             \\
KEAP1                                 & 7X4X            &      \includegraphics[width=4cm]{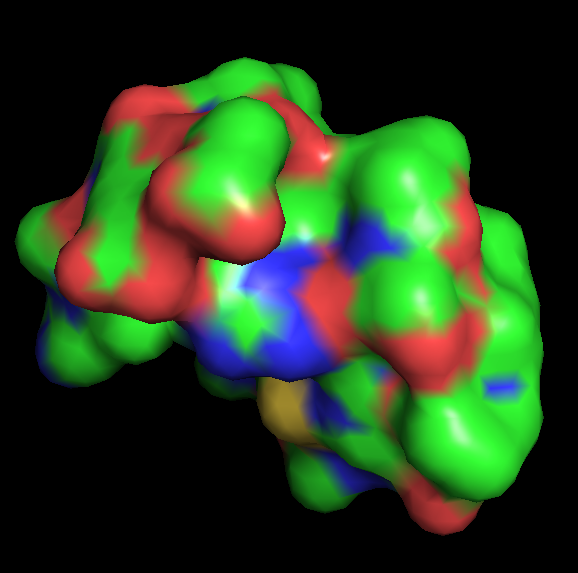}                     & 9J3                    &\includegraphics[width=4cm]{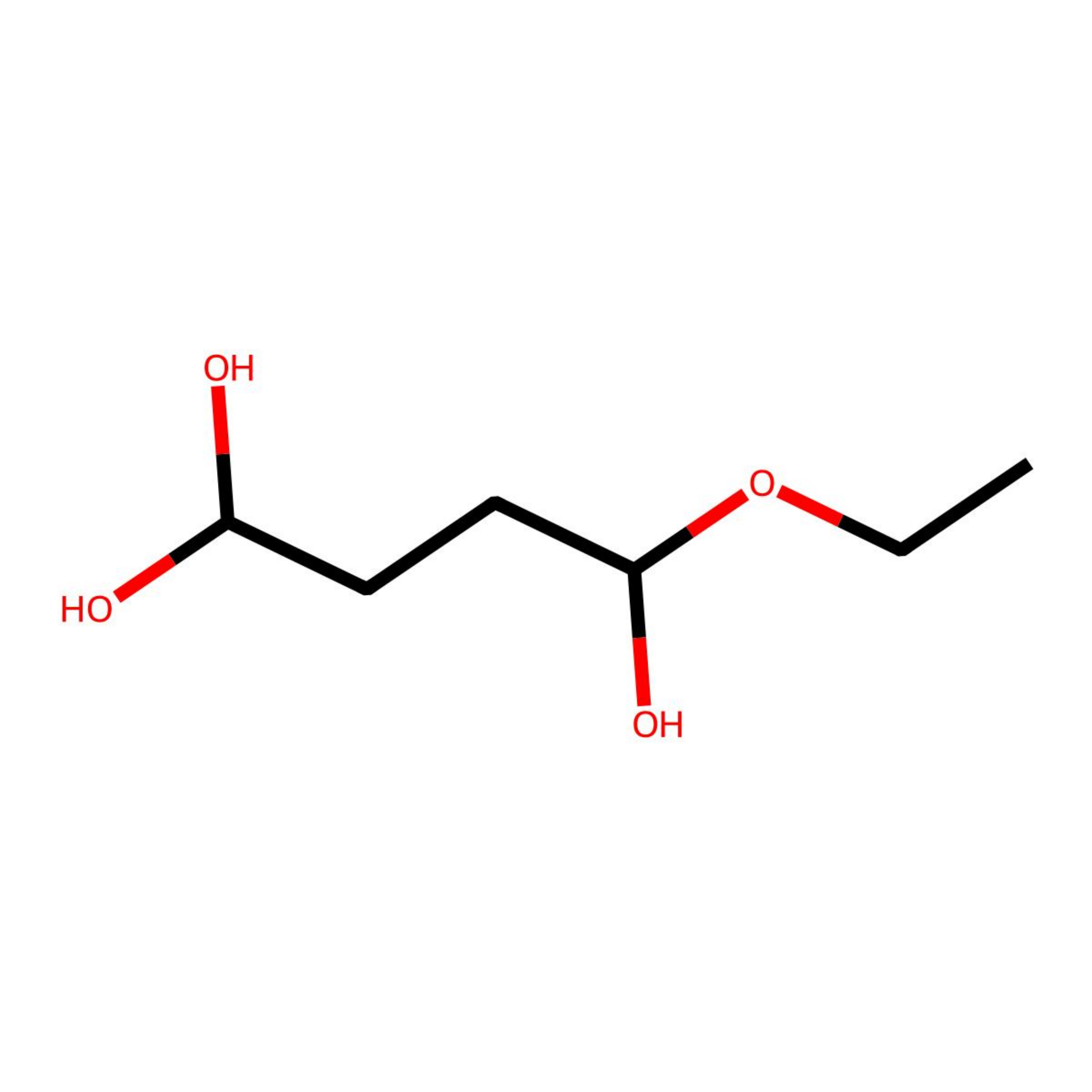}                                               & CA, N, A \\
KIT                                   & 4U0I            &      \includegraphics[width=4cm]{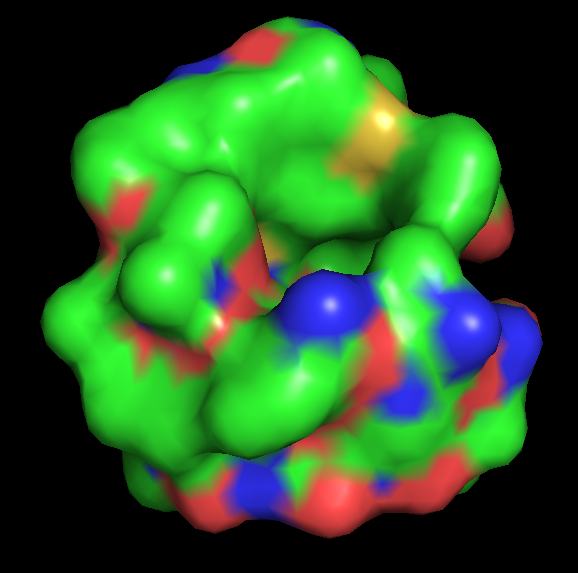}                     & 0LI                    &\includegraphics[width=4cm]{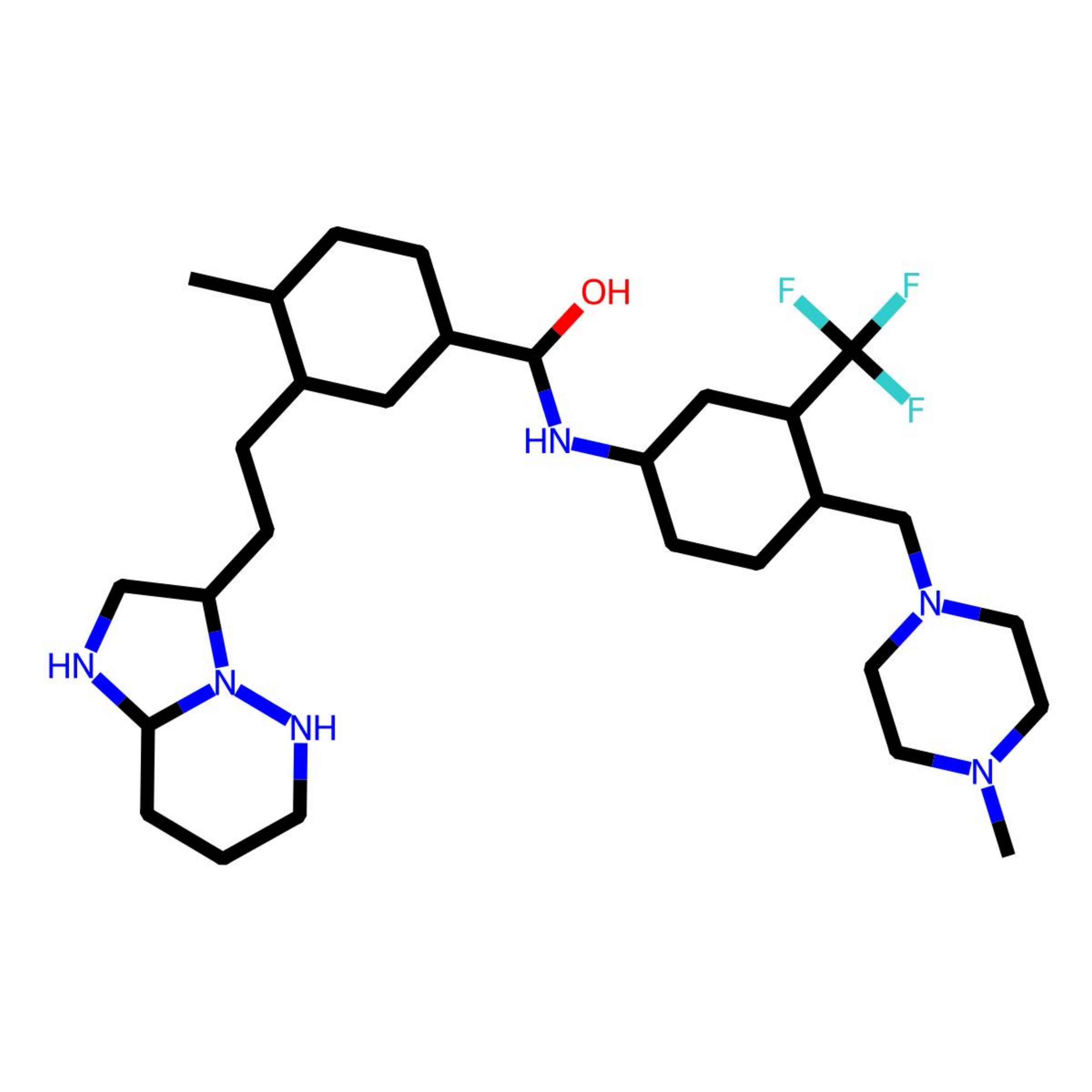}                                               & CA                                              \\
KRAS                                  & 4DSN            &      \includegraphics[width=4cm]{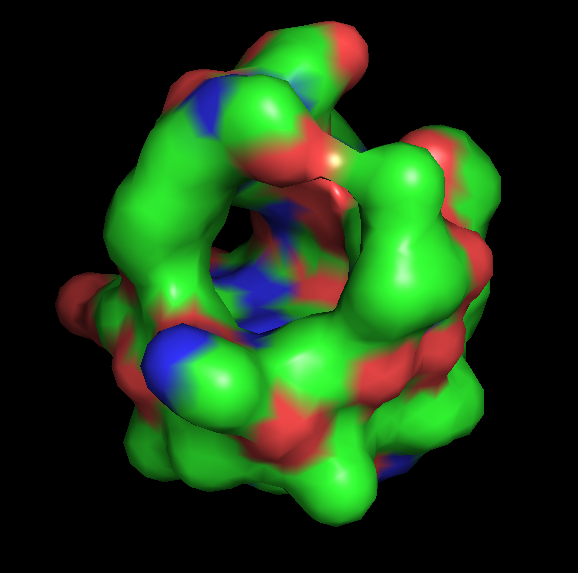}                     & GCP                    &\includegraphics[width=4cm]{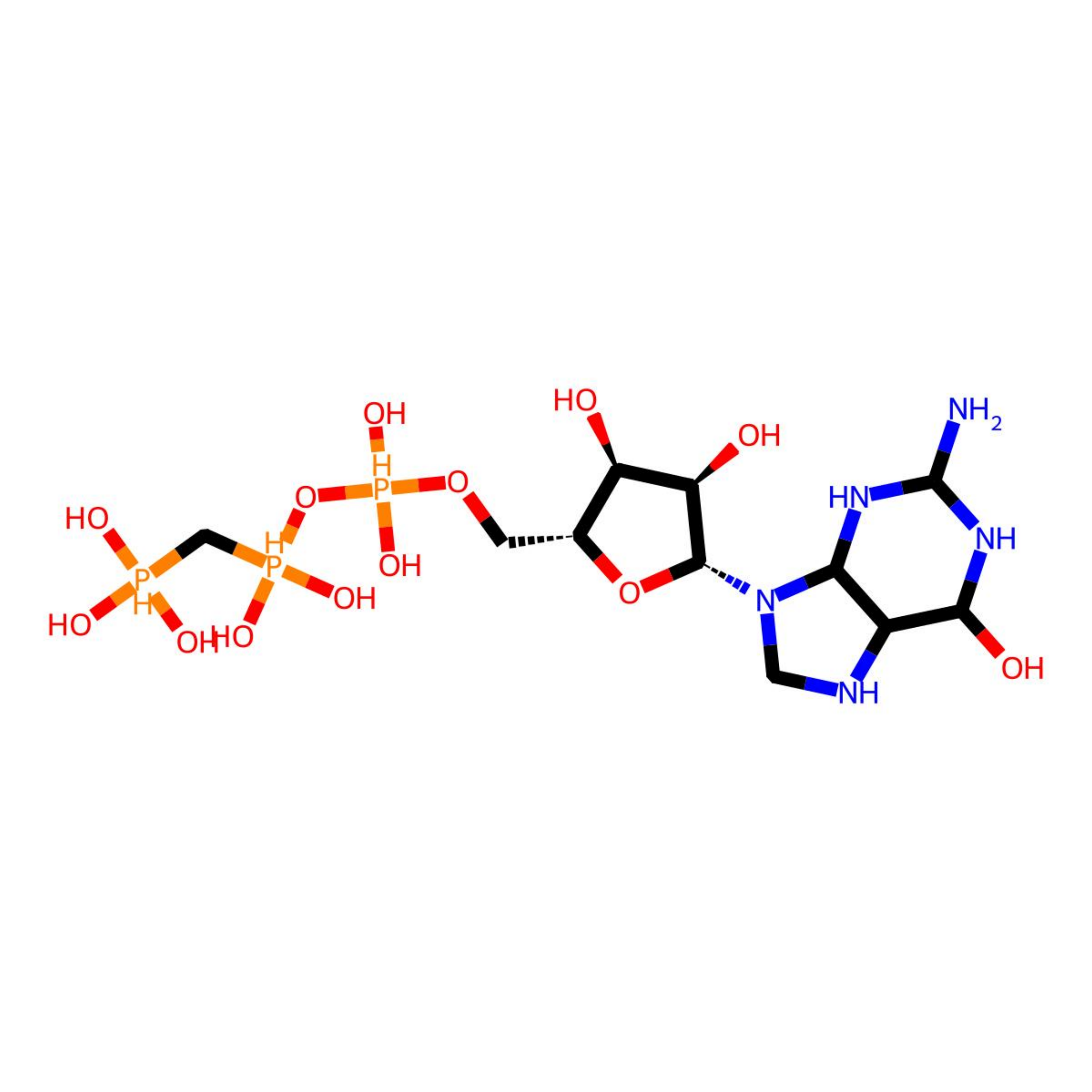}                                               & CA                                              \\
MET (HGFR)                            & 2RFN            &      \includegraphics[width=4cm]{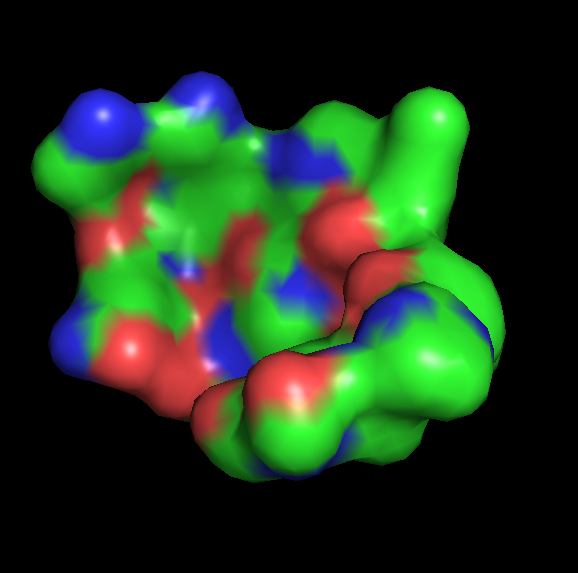}                     & AM7                    &\includegraphics[width=4cm]{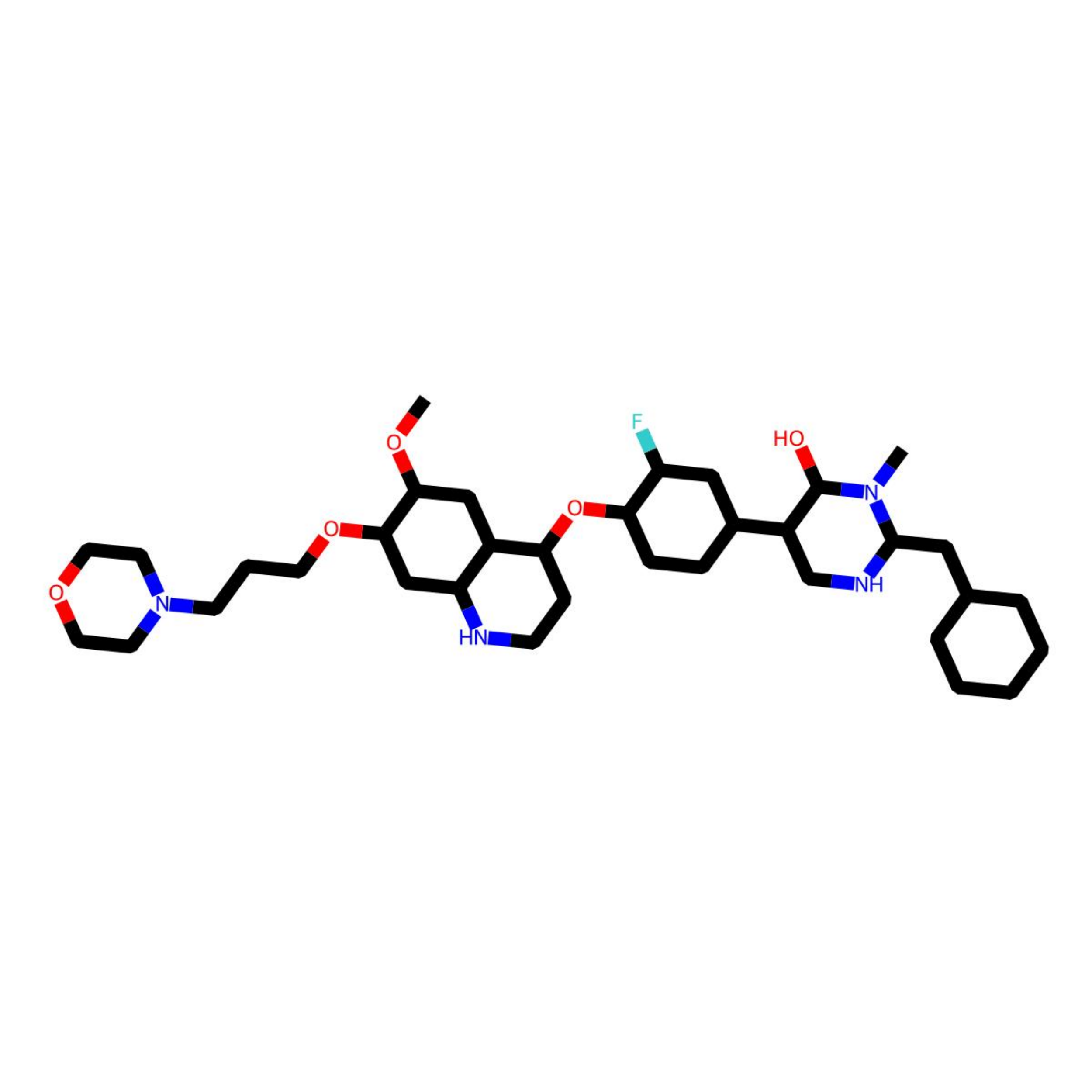}                                               & CA                                              \\
NR3C1 (GR)                            & 3K23            &        \includegraphics[width=4cm]{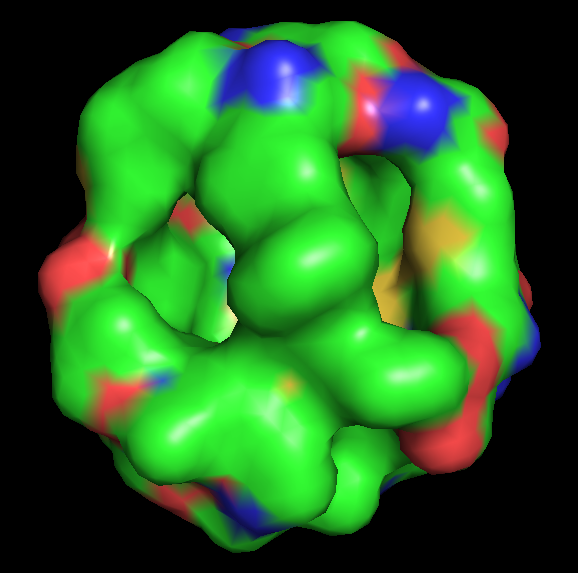}                   & JZN                    &\includegraphics[width=4cm]{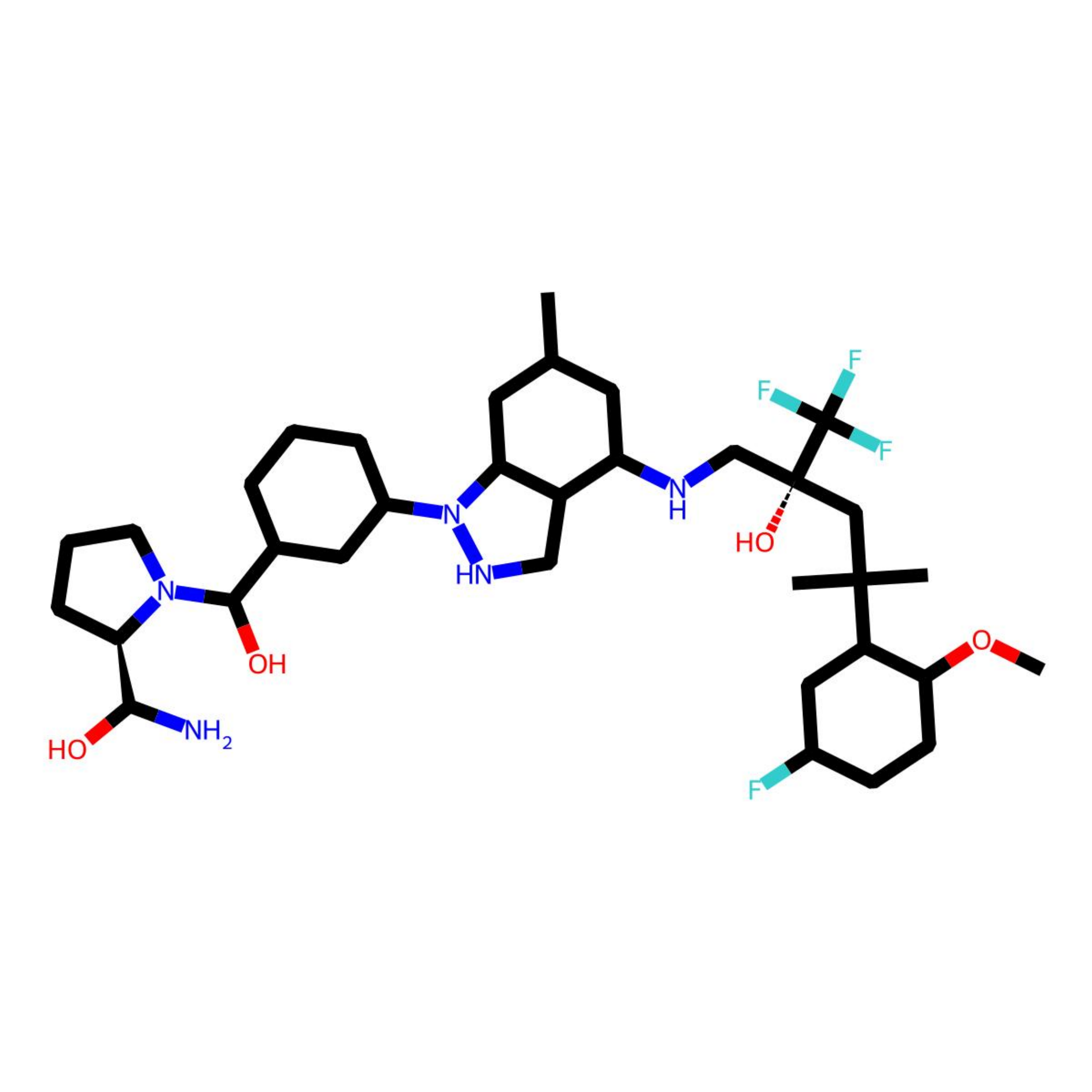}                                             & I, A              \\
P2RY12 (P2Y12)                        & 4NTJ            &        \includegraphics[width=4cm]{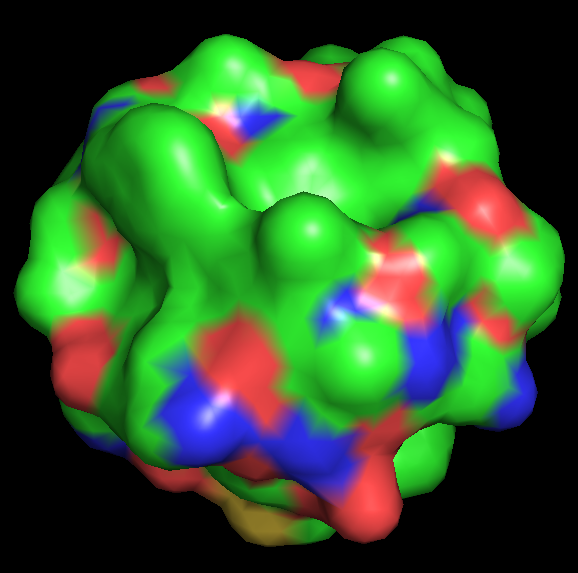}                   & AZJ                    &\includegraphics[width=4cm]{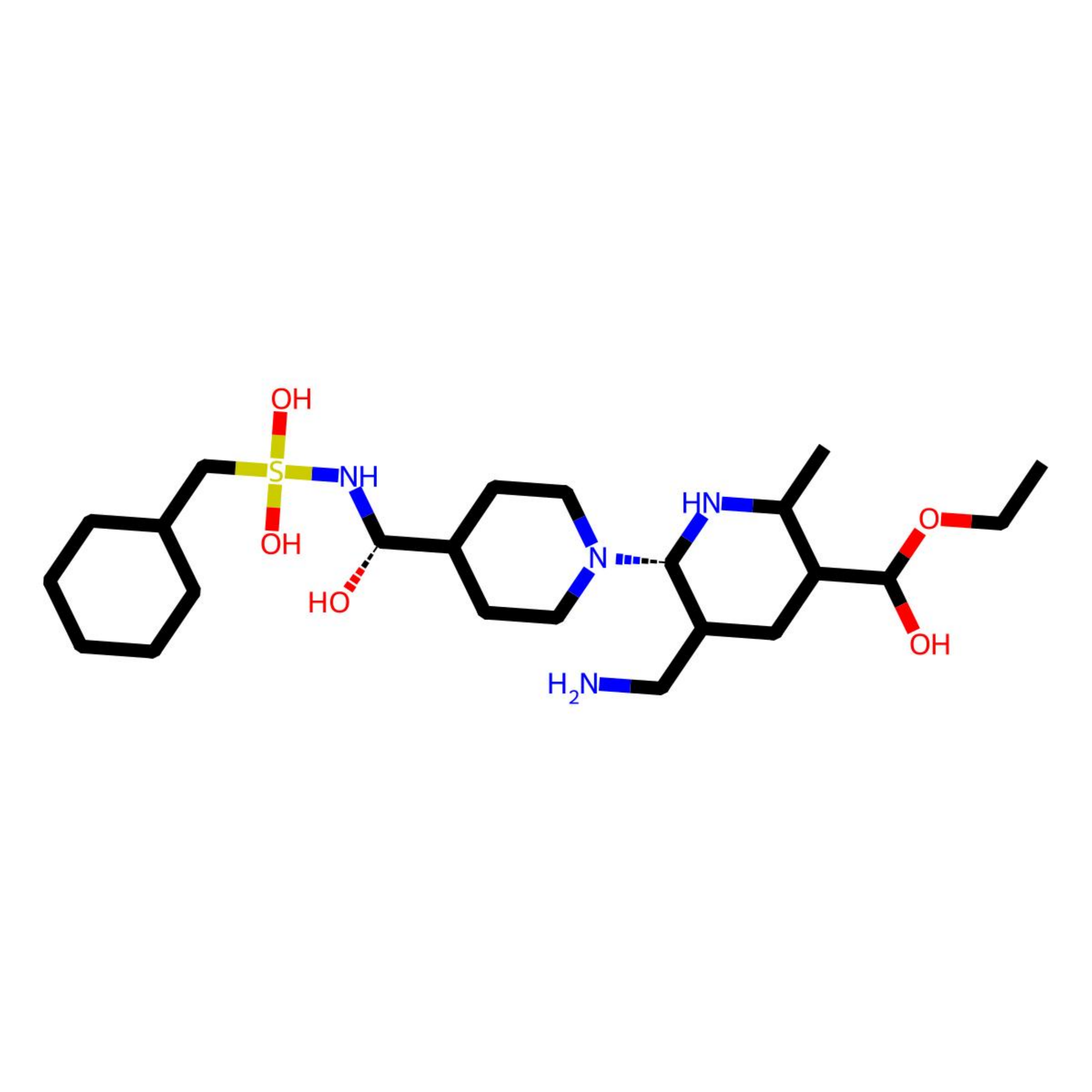}                                             & CV                              \\
PIK3CA                                & 4JPS            &         \includegraphics[width=4cm]{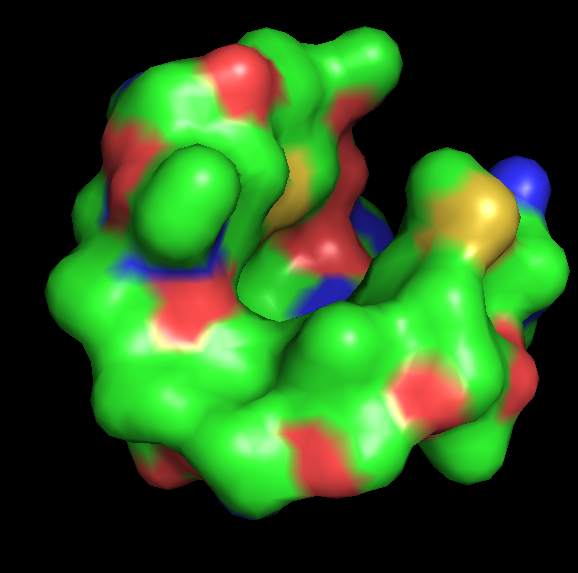}                  & 1LT                    &\includegraphics[width=4cm]{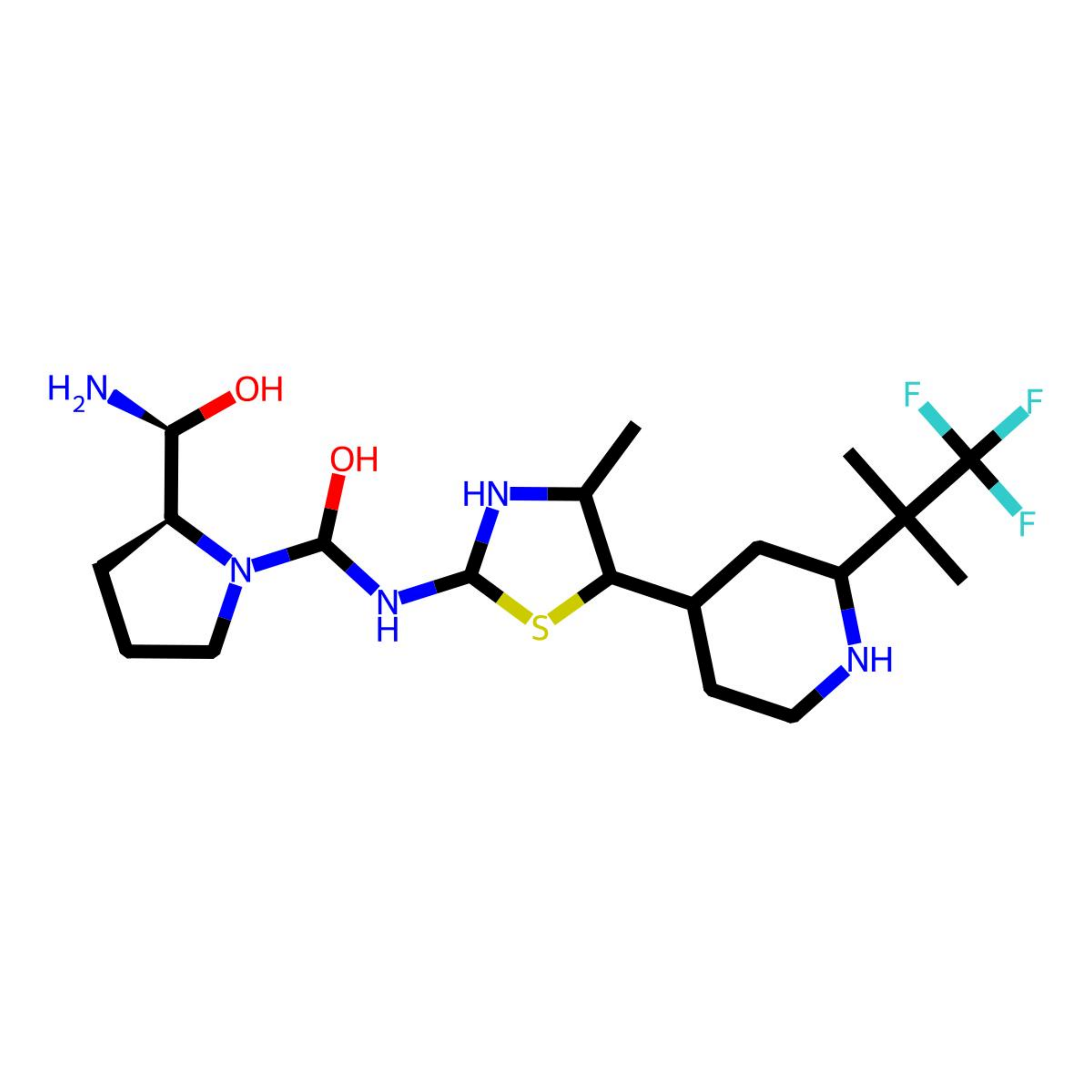}                                            & CA                                              \\
PSEN1                                 & 7C9I            &       \includegraphics[width=4cm]{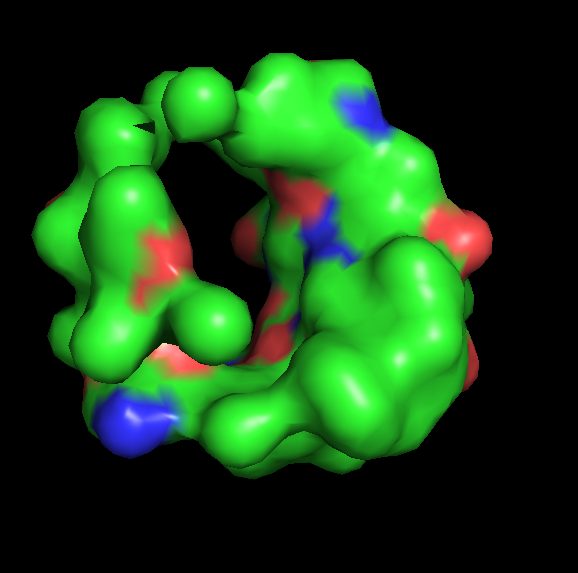}                    & FTO                    &\includegraphics[width=4cm]{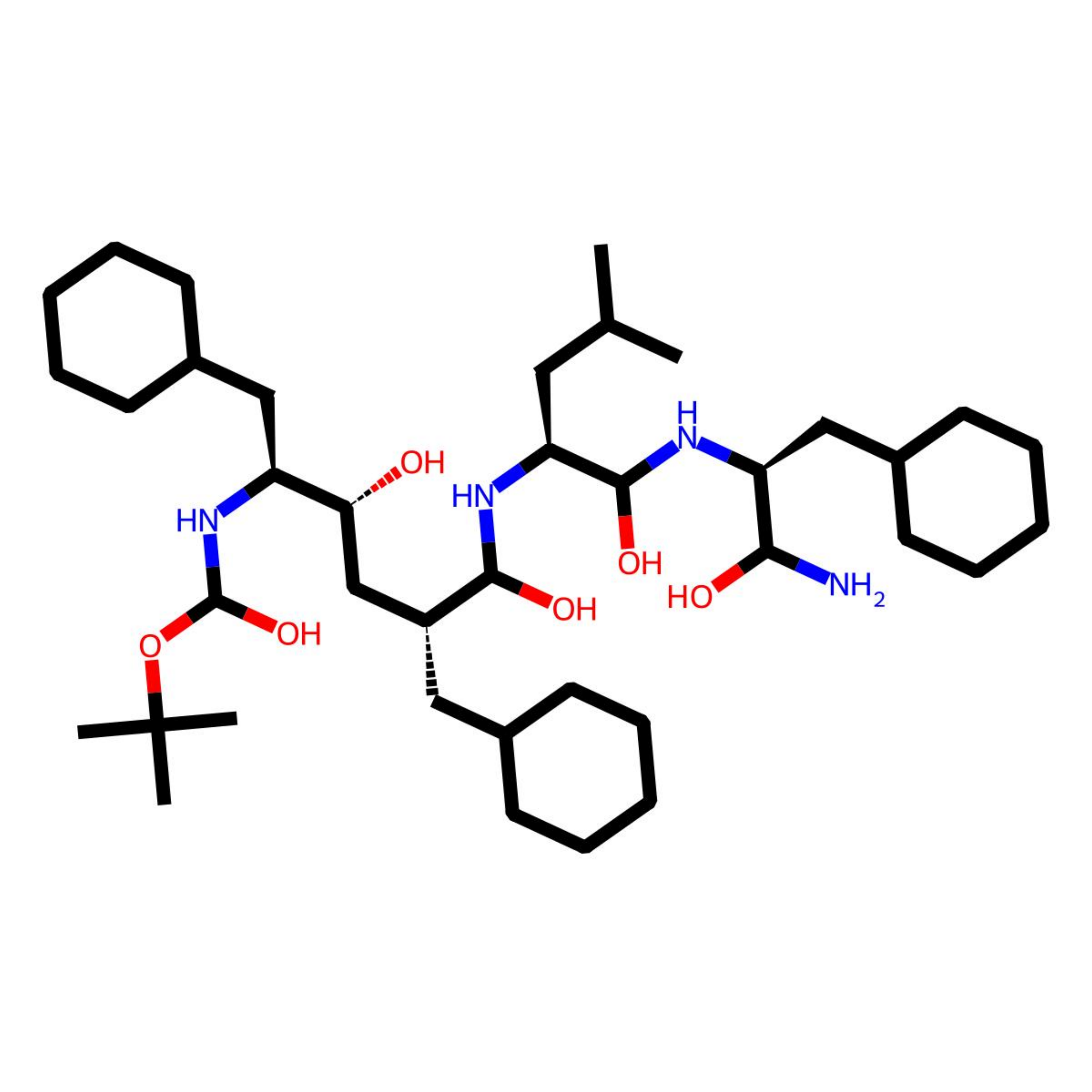}                                              & N, CV     \\
PTGS1 (COX1)                          & 2OYE            &      \includegraphics[width=4cm]{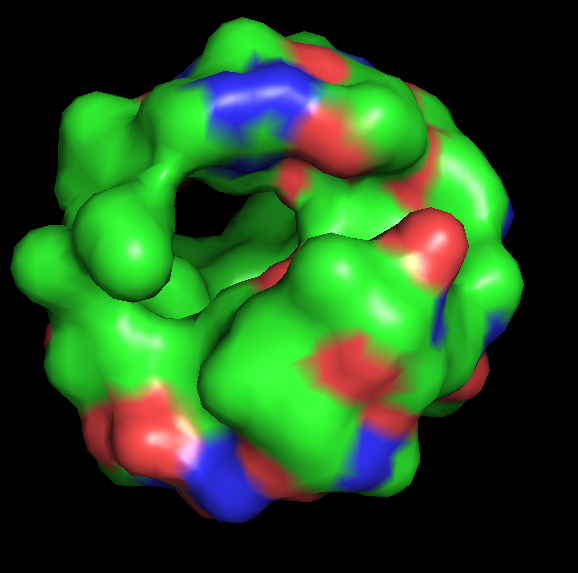}                     & IM8                    &\includegraphics[width=4cm]{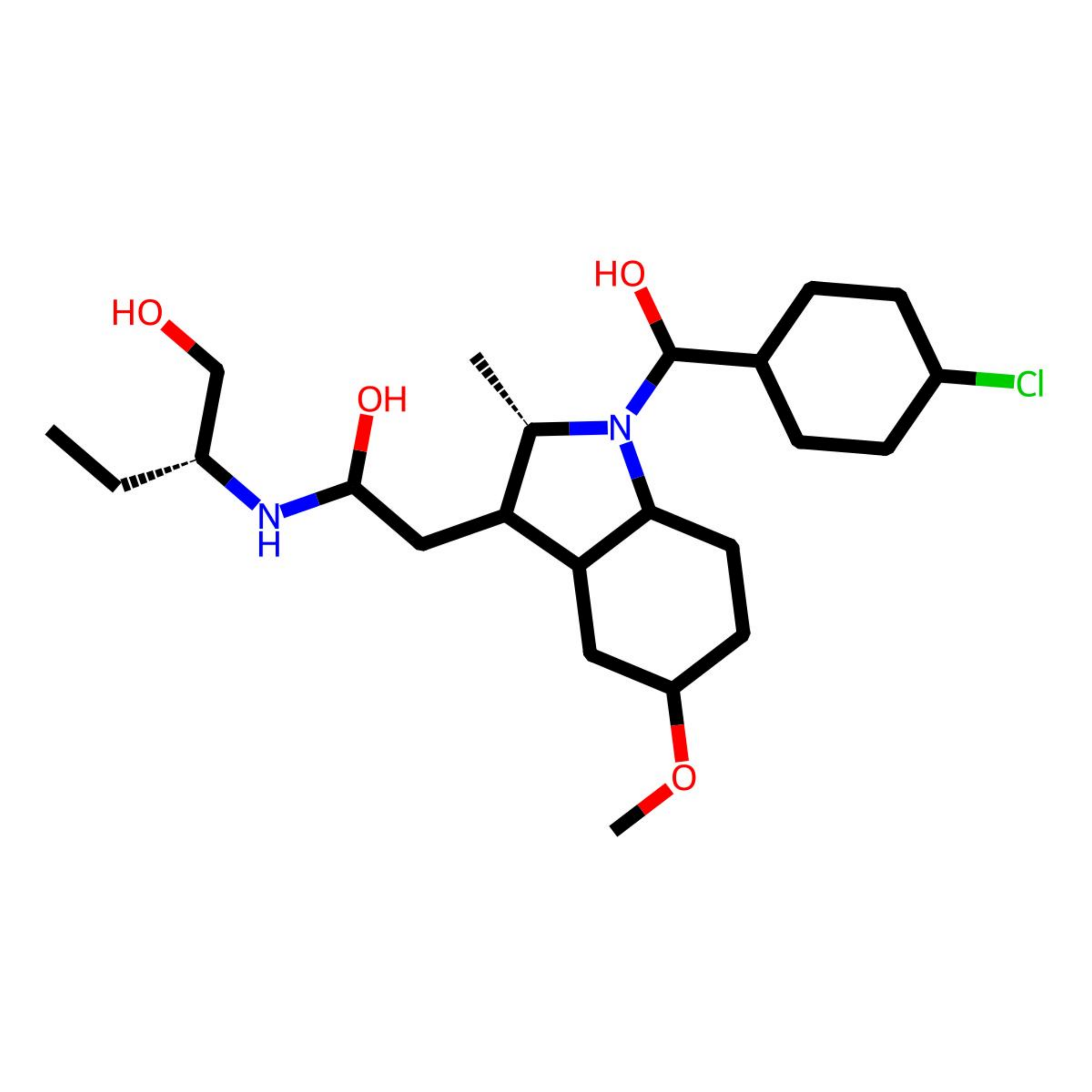}                                               & CV                              \\
PTGS2 (COX2)                          & 3LNO            &      \includegraphics[width=4cm]{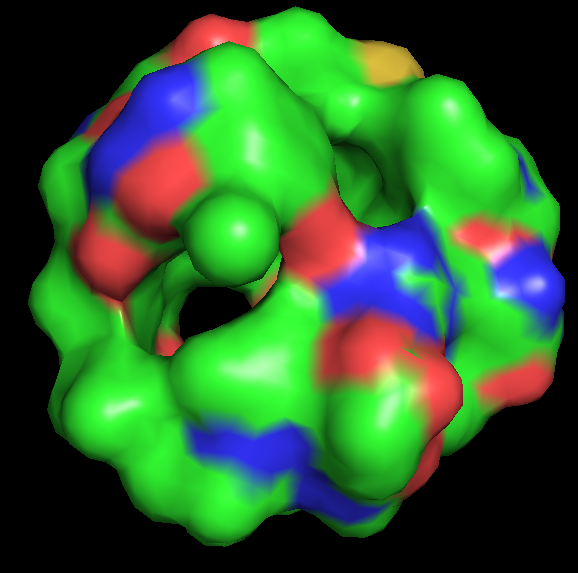}                     & 52B                    &\includegraphics[width=4cm]{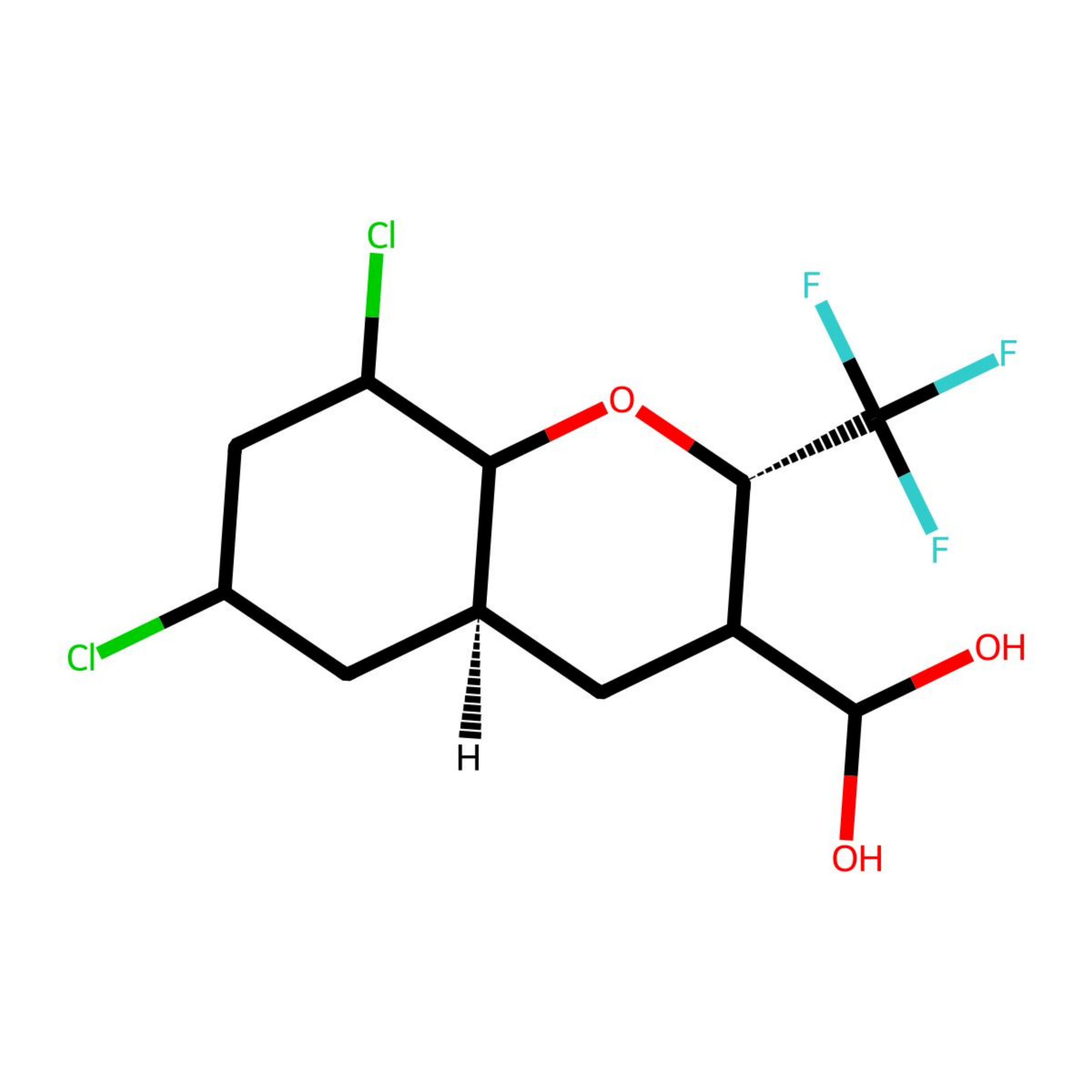}                                               & CV                              \\
SHP2 (PTPN11)                         & 7GS9            &      \includegraphics[width=4cm]{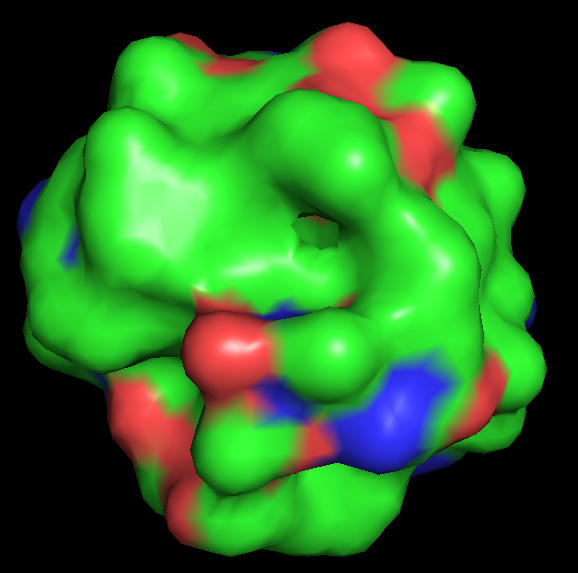}                     & LV7                    &\includegraphics[width=4cm]{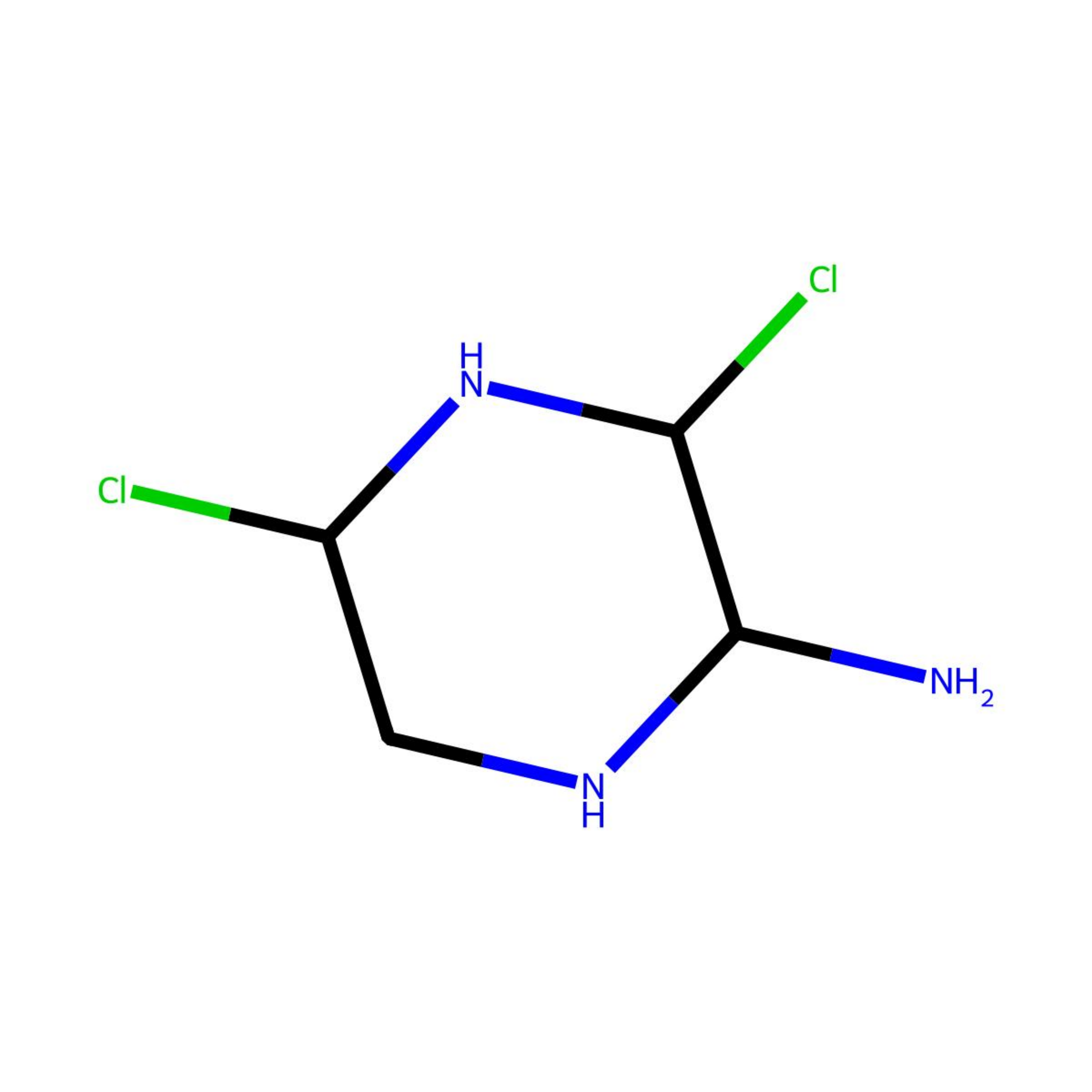}                                               & CA, D                                    \\
SLC6A2                                & 8HFL            &      \includegraphics[width=4cm]{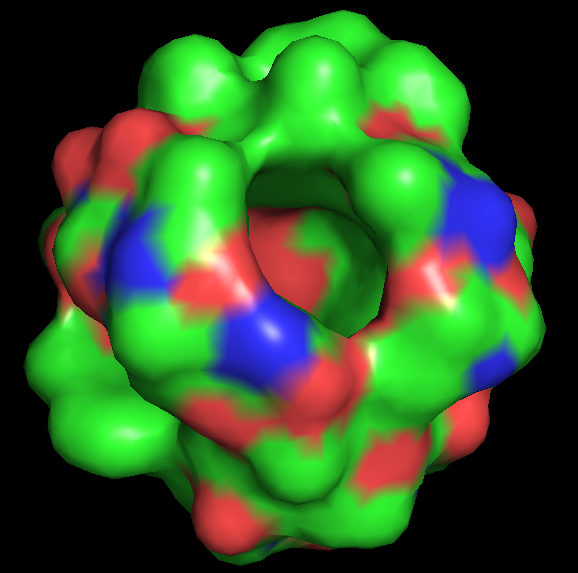}                     & 1XR                    &\includegraphics[width=4cm]{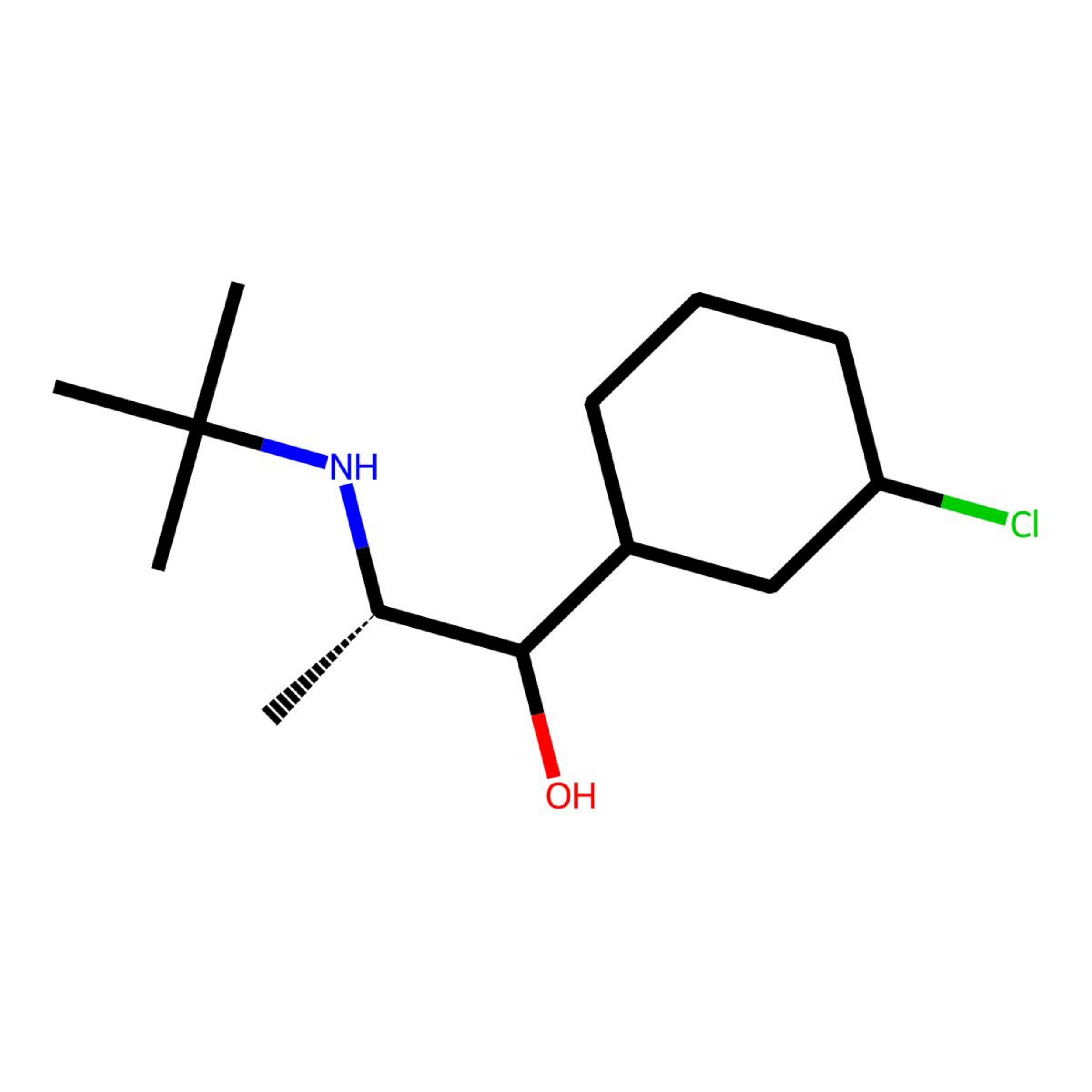}                                               & N                             \\
TNF                                   & 2AZ5            &       \includegraphics[width=4cm]{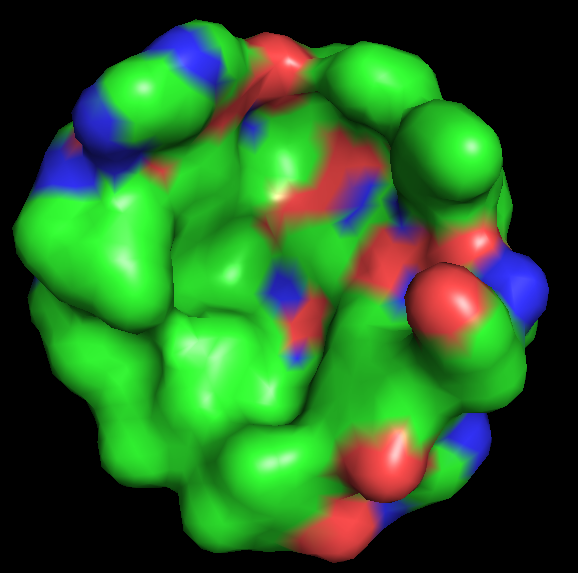}                    & 307                    &\includegraphics[width=4cm]{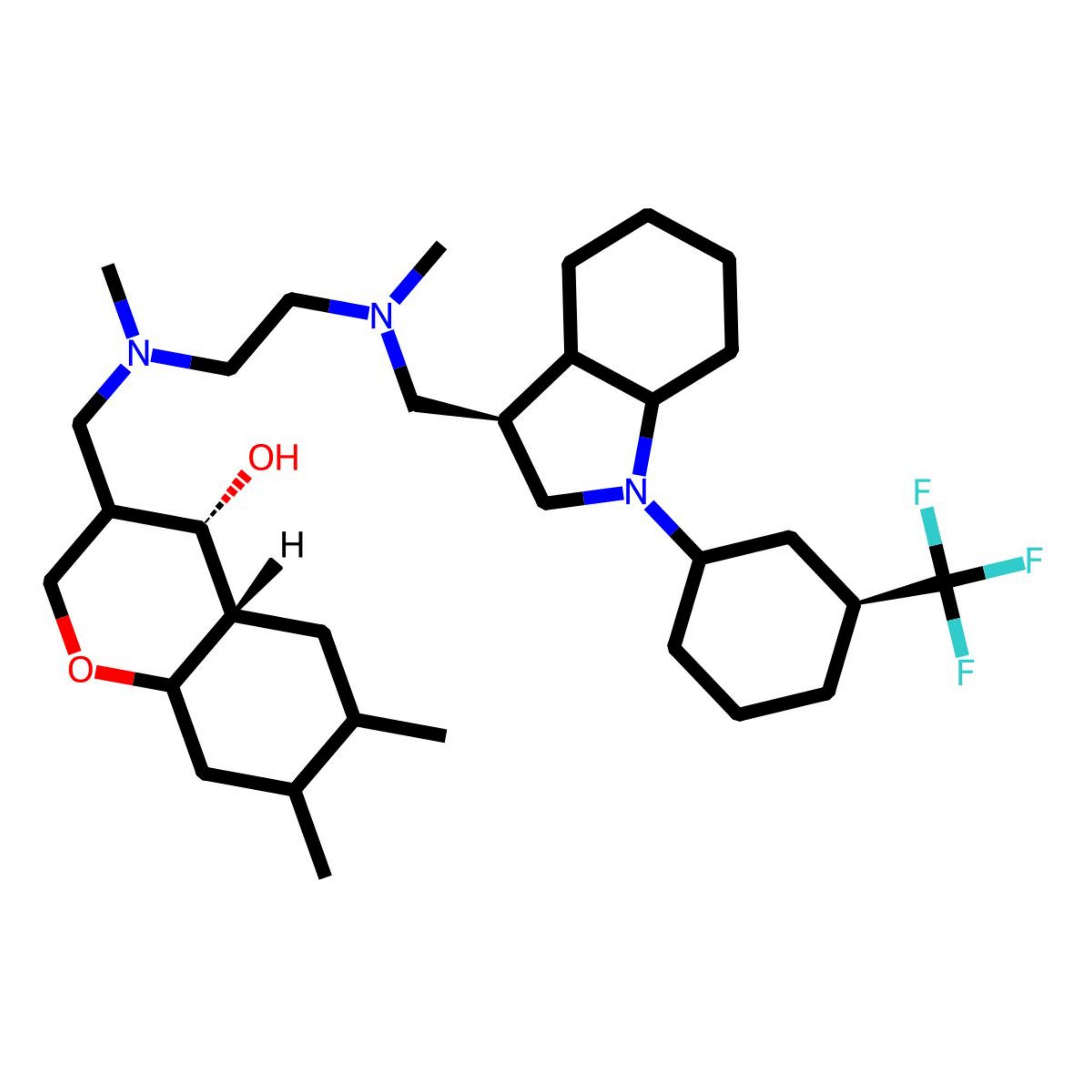}                                              & CA, I, A  

\end{longtable}   
\end{ThreePartTable}

\subsection{On the Necessity of Dataset Curation}
\label{sec:appendix:dataset:extra}

Existing benchmark datasets for drug discovery~\cite{guan_decompdiff_2023, guan_targetdiff_2023} contain non-human proteins. In fact, only around 40\% of the test set are human proteins. Furthermore, the targets in this test set do not necessarily have existing drugs associated with them, giving us no reference to the multi-property requirements needed to identify \HQ molecules. The lack of drugs associated with the targets is unsurprising, since the benchmark was originally designed for binding pose and affinity prediction~\cite{francoeur_three-dimensional_2020}. Instead, we have decided to manually curate a set of 30 proteins, all of which (1) are human proteins, (2) are associated with major diseases, and (3) have known drugs targeting them.

\section{Prompts}
\label{sec:appendix:prompts}

\subsection{\planner Prompt}

\begin{flushleft}
\texttt{You have access to the following molecules and pockets:}

\textbf{\texttt{\{pocket\_str\}\{mol\_str\}}}

\texttt{You also have access to a set of actions:}

\textbf{\texttt{\{action\_str\}}}

\texttt{Your job is to find molecules that satisfy these requirements:}

\textbf{\texttt{\{req\_str\}}}

\texttt{Here is a history of actions you have taken and the results:}

\textbf{\texttt{\{history\_str\}}}

\texttt{Here is the evaluation result from previous iteration:}

\textbf{\texttt{\{eval\_str\}}}

\vspace{1em}

\texttt{Let's think step by step and take your time before you answer the question. What is the best action to take and what is the input of the action?}

\vspace{1em}

\texttt{Remember that you currently have \textbf{\{resource\_str\}} left to solve the task.}

\texttt{Remember that you can only use one action.}

\vspace{1em}

\texttt{Your answer must follow this format:}

\vspace{1em}

\texttt{Action: [name of action]}

\texttt{Input: [input of the action, should be the identifier like ['MOL001'] or ['POCKET001']]}

\vspace{1em}

\texttt{If you plan to use "CODE" action, you need to include this additional format:}

\vspace{1em}

\texttt{Desc: [explain what you want to do with the input of the action. Be as verbose and descriptive as possible but at most three sentences. Always refer to the identifier of the action input.]}

\end{flushleft}

\subsection{\evaluator Prompt}

\begin{flushleft}

\texttt{You have access to the following pool of molecules:}

\textbf{\texttt{\{mol\_str\}}}

\texttt{Your job is to find molecules that satisfy these requirements:}

\textbf{\texttt{\{req\_str\}}}

\texttt{Does this pool of molecules satisfy the requirements?}

\texttt{Remember that all molecules in the pool must satisfy the requirements.}

\vspace{1em}

\texttt{Let's think step by step and answer with the following format:}

\texttt{Reason: (a compact and brief one-sentence reasoning)}

\texttt{Answer: (YES or NO)}

\end{flushleft}

\subsection{\code Prompt}

\begin{flushleft}

\texttt{Your job is to make a Python function called \_function.}

\texttt{The input is a Dict[str, pd.DataFrame] with the following columns:}

\texttt{["SMILES", "QED", "SAScore", "Lipinski", "Novelty", "Vina Score"].}

\texttt{The output must be a pandas DataFrame with the same columns as the input.}

\texttt{The function should be able to do the following task: \textbf{\{input\_desc\}}}

\vspace{1em}

\texttt{Your output must follow the following format:}

\vspace{1em}

\begin{verbatim}
import pandas as pd

def _function(Dict[str, pd.DataFrame]) -> pd.DataFrame:
    #---IMPORT LIBRARIES HERE---#
    #---IMPORT LIBRARIES HERE---#
    
    #---CODE HERE---#
    #---CODE HERE---#
    
    output_df = ...
    return output_df
\end{verbatim}

\vspace{1em}

\texttt{Make sure you import the necessary libraries.}

\end{flushleft}

\section{Additional Results}
\label{sec:appendix:results}

\subsection{Additional Baselines}
\label{sec:appendix:results:baselines}

We compare \agent with two more recent task-specific molecule generation methods, TargetDiff~\cite{guan_targetdiff_2023} and DecompDiff~\cite{guan_decompdiff_2023}. We run similar experiments as in~\autoref{tab:main} and present the results in~\autoref{tab:complete}. 
Overall, we observe similar results to~\autoref{sec:results:main}.
First, both \agent with DeepSeek-R1 and Claude significantly outperform both methods, highlighting the effectiveness of \agent. Second, both methods also struggle to generate new binding molecules better than existing drugs, while \agent does not. 

\subsection{Toxicity Predictions}
\label{sec:appendix:results:toxicity}

\begin{table}[t!]
    \centering
    \scriptsize
    \caption{Toxicity profiles of \agent's generated molecules and known drugs. Lower indicates better profile.}
    \label{tab:toxic}
    \begin{footnotesize}
    \begin{threeparttable}
    \begin{tabular}{
        @{\hspace{2pt}}l@{\hspace{5pt}}
        @{\hspace{5pt}}r@{\hspace{5pt}}
        @{\hspace{5pt}}r@{\hspace{2pt}}
    }
    \toprule
    Toxicity profile & \agent & Known Drugs \\
    \midrule
    Mutagenicity & 0.32 & \textbf{0.27} \\
    Carcinogenicity & \textbf{0.23} & 0.25 \\
    Clinical Toxicity & \textbf{0.08} & 0.33 \\
    DILI & 0.68 & \textbf{0.54} \\
    hERG Blocking & \textbf{0.31} & 0.59 \\
    Acute Toxicity & \textbf{2.45} & 2.78 \\
    \bottomrule
    \end{tabular}
    %
    \end{threeparttable}
    \end{footnotesize}
\end{table}

We use ADMET-AI \cite{swanson_admet-ai_2024} to predict the toxicity properties of \agent's generated molecules. 
We use toxicity properties from Chen et.al \cite{chen_generating_2025} as reference and select some that are available in ADMET-AI. We then compare them to drugs in our dataset and present the results in \autoref{tab:toxic}.

Overall, we observe that our generated molecules are better than or comparable to known drugs in terms of their toxicity properties.
Note that our agent is specifically designed to generate high-quality molecules, not ``safe'' molecules. Yet,~\autoref{tab:toxic} shows that their safety profiles are also promising. This presents interesting findings that (1) the current design of \agent can generate both high-quality and ``safe'' molecules and (2) a significant opportunity for improvements to \agent.

\subsection{Failure Analysis}
\label{sec:appendix:results:failure}

We observe that \agent yields suboptimal results on some targets (\autoref{sec:results:case_study_action}). First, we remark that on all targets, \agent can generate at least one high-quality molecule, including on PIK3CA, MET, and ADRB2.
However, they are suboptimal since: (1) the number of high-quality molecules is not sufficient (i.e., less than 5), or (2) the molecules are not diverse enough.
We hypothesize that existing tools are struggling because of the structure of the pockets.
For instance, the pocket may only allow a few specific scaffolds to bind,
making it extremely difficult for existing tools to generate many and diverse high-quality molecules.

\subsection{Computational Costs}
\label{sec:appendix:results:cost}

Overall, we observe that \agent takes about 9K input tokens and 5K output tokens per target in our experiments. The estimated cost to generate high-quality molecules using \agent is around \$0.03 (USD)
calculated using Claude pricing based on the number of used tokens.
This highlights the potential of \agent for 
low-cost autonomous drug discovery.

\begin{figure*}[tb]
    \centering
    \begin{subfigure}{0.3\linewidth}
        \includegraphics[width=\linewidth]{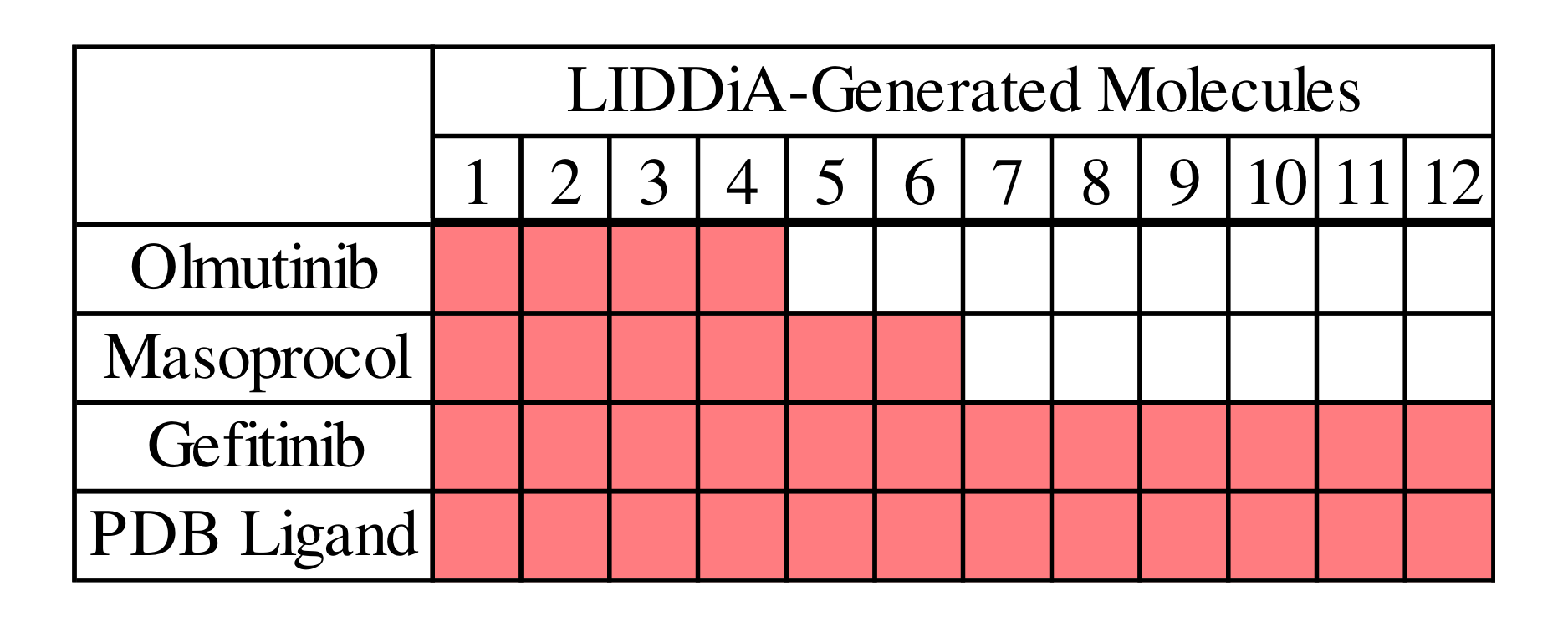} 
        \vskip -8pt
        \caption{\vinareq} 
        \label{fig:case_heatmap_vna} 
    \end{subfigure}
    \begin{subfigure}{0.3\linewidth}
       \includegraphics[width=\linewidth]{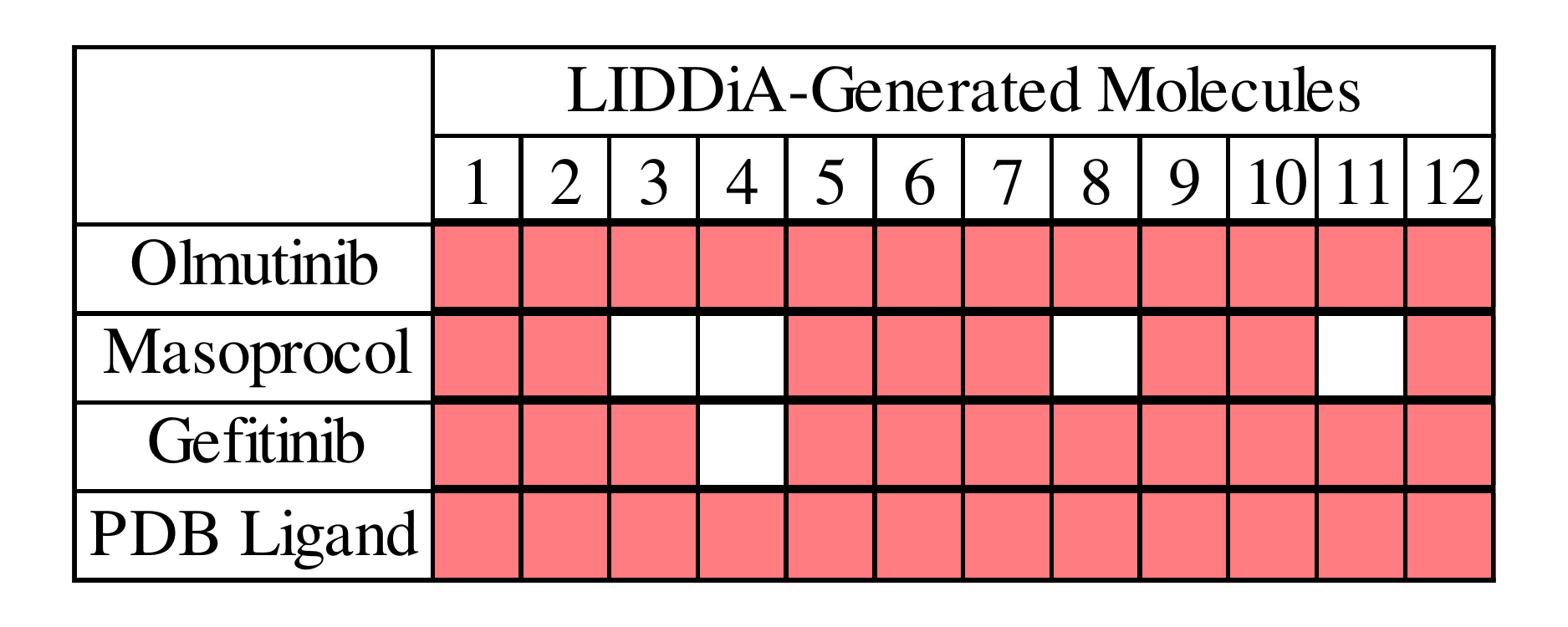}
       \vskip -8pt
        \caption{\qedreq} 
        \label{fig:case_heatmap_qeq} 
    \end{subfigure}
    \begin{subfigure}{0.3\linewidth}
       \includegraphics[width=\linewidth]{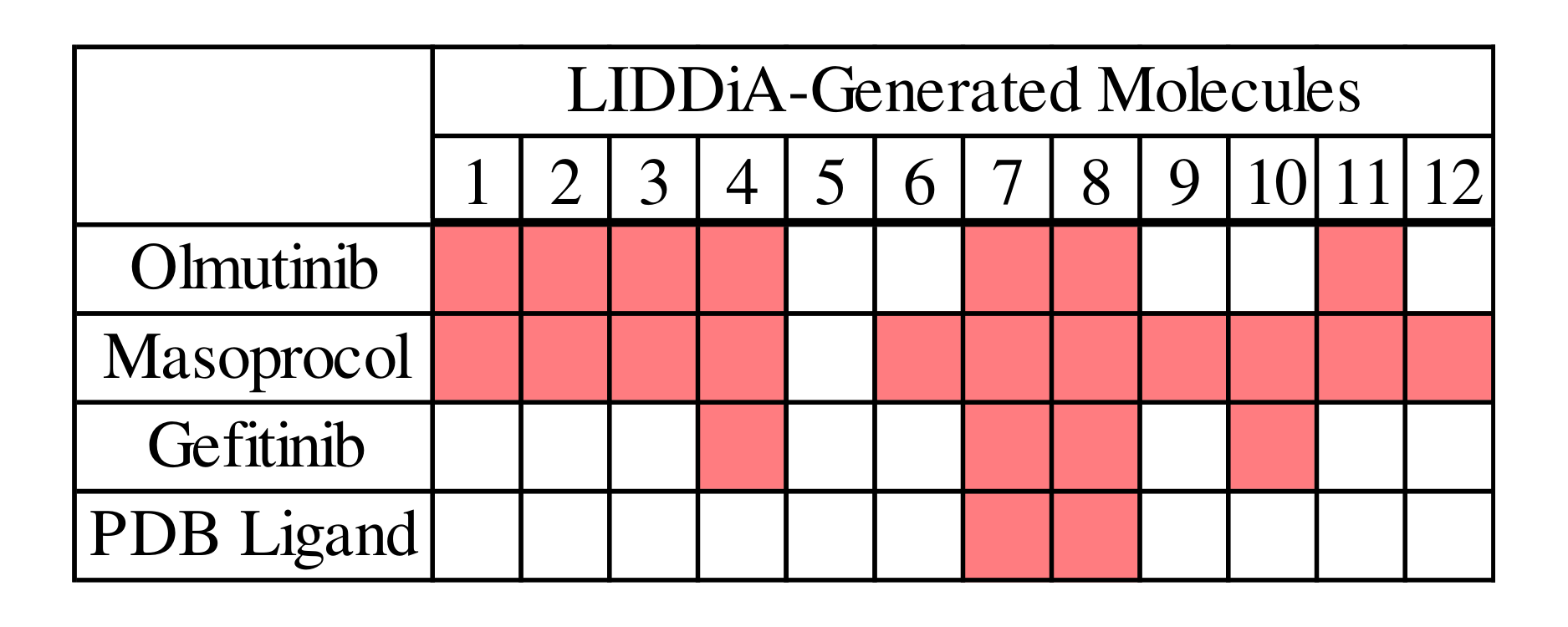}
       \vskip -8pt
        \caption{\sascorereq} 
        \label{fig:case_heatmap_sas} 
    \end{subfigure}
    \vspace{-10pt}
    \caption{Case study for EGFR. Each subfigure compares molecules generated by \agent to 
    three drugs and one binding ligand of EGFR on \vinareq, \sascorereq, and \qedreq, respectively. 
    Red squares indicate that the \agent molecule outperforms the reference molecule on respective metrics.}
    \label{fig:case_study_compounds}
\end{figure*}

\begin{figure}[t]
\centering
    \begin{subfigure}{0.275\linewidth}
        \includegraphics[width=\linewidth]{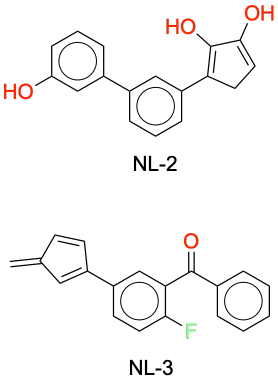}  
        \caption{}
        \label{fig:egfr:agent} 
    \end{subfigure}
    \hfill
    \begin{subfigure}{0.275\linewidth}
        \includegraphics[width=\linewidth]{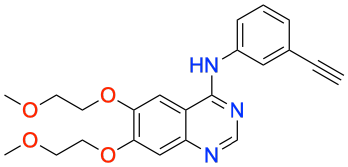}  
        \caption{}
        \label{fig:egfr:ligand} 
    \end{subfigure}
    \hfill
    \begin{subfigure}{0.400\linewidth}
        \includegraphics[width=\linewidth]{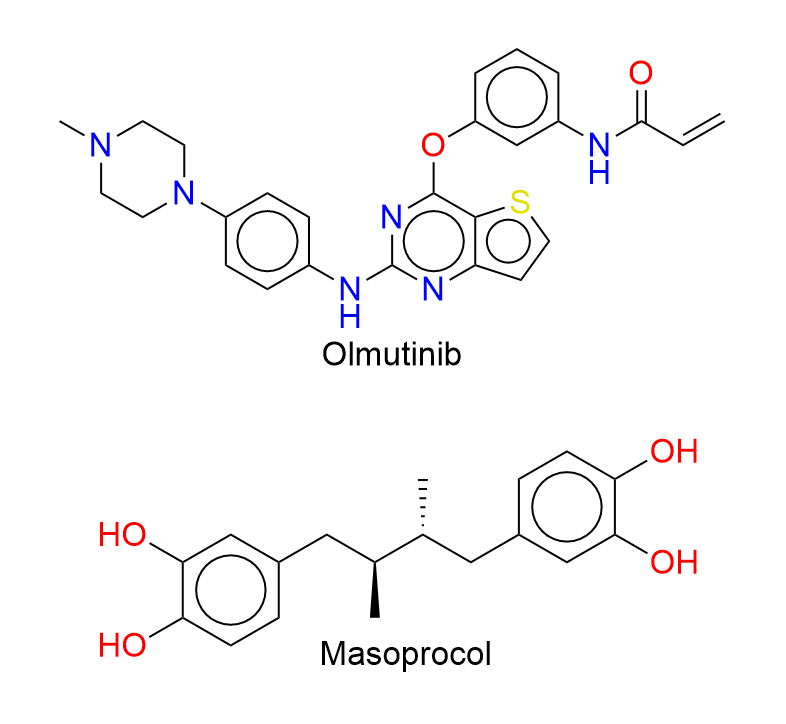}    
        \caption{}
        \label{fig:egfr:drugs} 
    \end{subfigure}
    \caption{Case study on EGFR. (a) \agent's generated molecules (NL-2 and NL-3). (b) Known ligand for EGFR. (c) Examples of known approved drugs for EGFR. NL-2 has two enol groups and NL-3 has a fulvene, both of which are problematic as drug candidates.}
    \label{fig:egfr}
\end{figure}

\begin{figure}[t]
\centering
    \begin{subfigure}{0.45\linewidth} 
        \includegraphics[width=\linewidth]{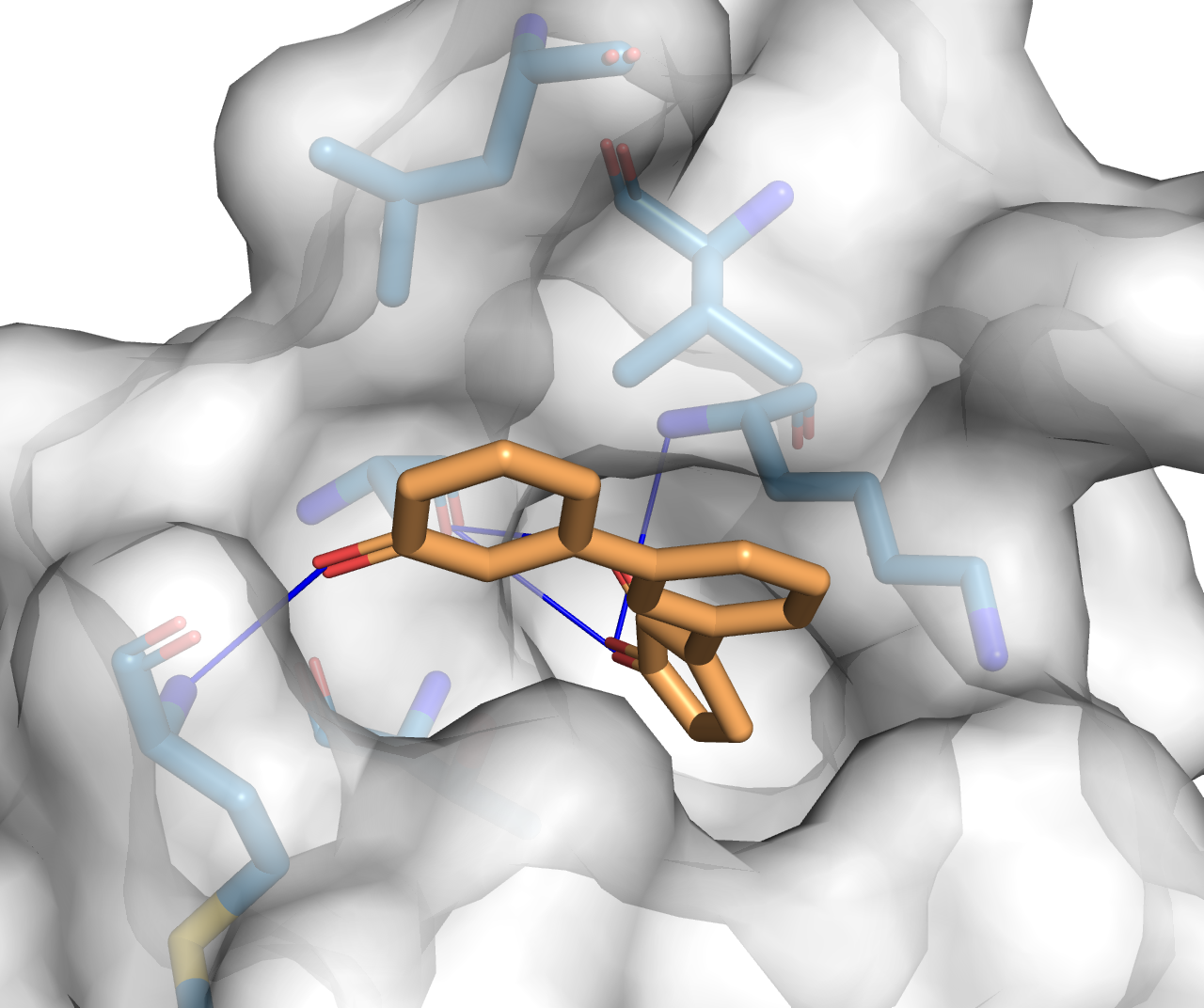} 
        \caption{}
        \label{fig:two_mol_docking:mol_1} 
    \end{subfigure}
    \hfill
    \begin{subfigure}{0.45\linewidth} 
        \includegraphics[width=\linewidth]{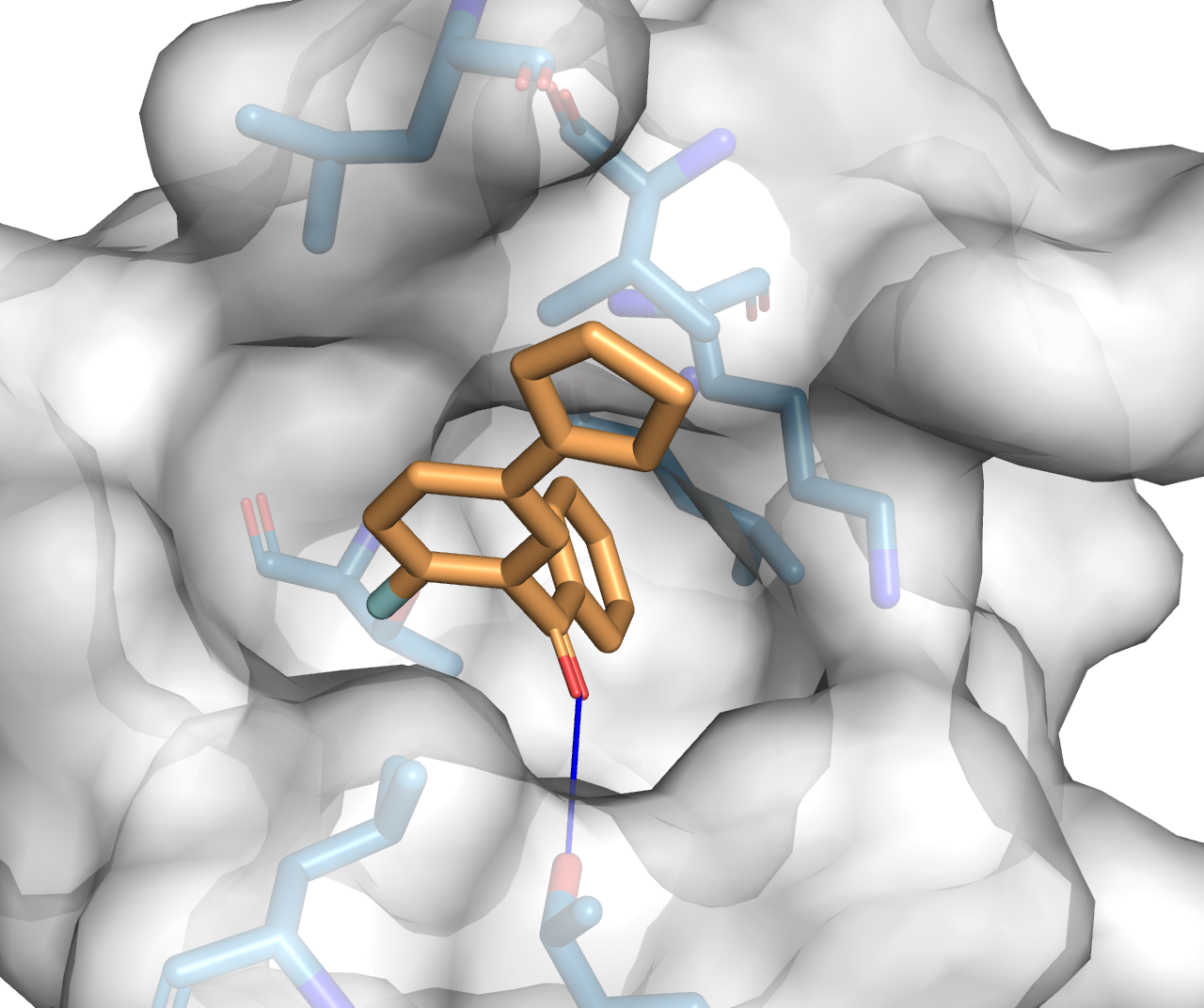} 
        \caption{}
        \label{fig:two_mol_docking:mol_2} 
    \end{subfigure}
    \caption{(a) Docking of NL-2 to EGFR pocket, with hydrogen bonds shown as solid blue lines. 
    (b) Docking of NL-3 to EGFR pocket, with hydrophobic van der Waals contacts shown using solid blue lines.} 
    \label{fig:two_mol_docking}
\end{figure}

\subsection{Case Study on EGFR}
\label{sec:appendix:results:case}

We present another case study and task \agent with discovering new potential drug therapies targeting the Epidermal growth factor receptor 1 (EGFR) protein.
EGFR is a transmembrane glycoprotein that plays a pivotal role in many cancers,
including breast cancer, esophageal cancer, and lung cancer~\cite{seshacharyulu_egfr_2012}.
Its role in cancer, as well as its accessibility on the cell membrane, 
has made it a prime therapeutic target~\cite{bento_chembl_2014, gaulton_chembl_2011}.
However, cancer cells mutate rapidly and can become resistant to drugs over time, 
leading to a need for novel drug therapies~\cite{das_next-generation_2024}.
We compare the molecules generated for EGFR by \agent 
with three approved drugs of EGFR -- Olmutinib, Masoprocol, and  Gefitinib,
which exhibit the best \vinareq, \qedreq and \sascorereq among all EGFR's approved drugs, respectively.
We also compare them with a known ligand for EGFR's binding pocket.
~\autoref{fig:case_study_compounds} presents the overall comparison results.
Note that most existing methods cannot generate any novel high-quality molecules. 
This emphasizes the strength of \agent, which can tackle even a challenging target.

\emph{\textbf{\agent effectively generates promising novel drug candidates on EGFR.}}
Notably, these molecules surpass the native ligand in both \vinareq and \qedreq, while displaying comparable overall profiles to approved drugs.
Moreover, some molecules (Figure~\ref{fig:case_study_compounds} columns 1 and 2) are better than Olmutinib and Masoprocol on all metrics (i.e., \vinareq, \qedreq, and \sascorereq).
We illustrate the two molecules, NL-2 and NL-3, in Figure~\ref{fig:egfr}.
These molecules possess structural features that allow them to bind the pocket well.
Notably, NL-2 has hydroxyl groups on both the five- and six-membered rings from the first molecule, which form strong hydrogen bonds with the protein target on opposite sides of the pocket (Figure~\ref{fig:two_mol_docking:mol_1}).
NL-3 utilizes a different binding strategy, relying on hydrophobic packing and shape complementarity rather than polar interactions.
As shown in Figure~\ref{fig:two_mol_docking:mol_2}, the fluorine substituent is positioned near the pocket entrance 
flanked by hydrophobic residues, 
which serve as favorable van der Waals contacts.
Meanwhile, the diarylketone moiety is buried deep within the binding pocket, anchoring the ligand through planar stacking and hydrophobic interactions, despite the absence of direct hydrogen bonding.
This finding aligns with previous literature~\cite{zhang2023palladium}, which highlights the potency of diarylketone for antitumor drugs.

\emph{\textbf{Medicinal chemists are essential for a successful real-world deployment of \agent}}. 
Despite the favorable binding and \textit{in silico} properties, closer examination reveals some concerning structural features in these molecules. 
NL-2 contains two enol groups (the -OH near the double bond)
---substructures with tautomeric instability and are highly unattractive for drugs~\cite{hart1979simple}.
NL-3 contains fulvene, known to be chemically reactive, thermally unstable, sensitive to oxygen, and photosensitive~\cite{swan2019overview}. 
The diarylketone moiety, despite its favorable binding and potency in antitumor drugs~\cite{zhang2023palladium}, is known to be phototoxic~\cite{dubois2021investigating}.
Such conflicts (e.g., favorable binding but phototoxic) are typical in drug discovery, highlighting its significant challenges and the necessity for more comprehensive evaluation for a successful practical deployment of \agent.

Furthermore, no standalone \emph{in silico} evaluation tools (e.g., computational filters from RDKit~\cite{rdkit} and Medchem~\cite{schuffenhauer2020evolution})
can detect all the issues presented in these molecules. 
Several filters (e.g., PAINS~\citep{baell2010new}, BRENK~\citep{brenk2008lessons}, NIH~\citep{jadhav2010quantitative, doveston2015unified}) cannot capture the problematic features in NL-2, highlighting the limitation of existing tools.
Lilly rules~\citep{bruns2012rules} are able to identify the enol groups, but do not raise any alerts for NL-3.
These findings underscore that human expertise remains irreplaceable in drug discovery---a domain where nuanced understanding and reliable assessments are critical for mitigating risks.
They also highlight three priorities for the future work of \agent: (1) human-in-the-loop validation, (2) development and integration of more sophisticated \textit{in silico} tools, and (3) wet-lab validation of generated molecules. 

As discussed, the lack of nuanced understanding by existing \textit{in silico} tools contribute to the problematic features existed in \agent's pool of generated molecules. 
A more reliable option is the inclusion of human experts in the loop for validation for a more nuanced and comprehensive evaluation of the molecules. 
Meanwhile, existing \textit{in silico} tools, particularly in evaluation, have rooms for improvement. 
The integration of more and better state-of-the-art tools can certainly benefit \agent in generating more and better high-quality molecules. 
Ultimately, \textit{in vitro} and \textit{in vivo} in a laboratary will be necessary to analyze how \agent's performance translates to real-world impacts.
Thus, though encouraging, \agent calls for more comprehensive and systematic investigation for a successful practical deployment.

\begin{table*}[!t]
\centering
\rotatebox{90}{
\begin{minipage}{\textheight}
\centering
\caption{Performance comparison between task-specific molecule generation methods (Pocket2Mol, DiffSMol, TargetDiff, DecompDiff), general-purpose LLMs (Claude, GPT-4o, o1-mini and o1), and \agent variants.} 
\vspace{-8pt}
\label{tab:complete}
   \begin{footnotesize}	
  \begin{threeparttable}
      \begin{tabular}{
        @{\hspace{0pt}}l@{\hspace{0pt}}
        @{\hspace{2pt}}l@{\hspace{1pt}}
        @{\hspace{0pt}}r@{\hspace{1pt}} 
        @{\hspace{1pt}}r@{\hspace{0pt}}
        @{\hspace{3pt}}c@{\hspace{3pt}}	
        @{\hspace{0pt}}r@{\hspace{1pt}} 
        @{\hspace{1pt}}r@{\hspace{0pt}}
        @{\hspace{3pt}}c@{\hspace{3pt}}	
        @{\hspace{0pt}}r@{\hspace{1pt}} 
        @{\hspace{1pt}}r@{\hspace{0pt}}
        @{\hspace{3pt}}c@{\hspace{3pt}}	
        @{\hspace{0pt}}r@{\hspace{1pt}} 
        @{\hspace{1pt}}r@{\hspace{0pt}}
        @{\hspace{3pt}}c@{\hspace{3pt}}	
        @{\hspace{0pt}}r@{\hspace{1pt}} 
        @{\hspace{1pt}}r@{\hspace{0pt}}
        @{\hspace{3pt}}c@{\hspace{3pt}}	
        @{\hspace{0pt}}r@{\hspace{1pt}} 
        @{\hspace{1pt}}r@{\hspace{0pt}}
        @{\hspace{3pt}}c@{\hspace{3pt}}	
        @{\hspace{0pt}}r@{\hspace{1pt}} 
        @{\hspace{1pt}}r@{\hspace{0pt}}
        @{\hspace{3pt}}c@{\hspace{3pt}}	
        @{\hspace{0pt}}r@{\hspace{1pt}} 
        @{\hspace{1pt}}r@{\hspace{0pt}}
        @{\hspace{3pt}}c@{\hspace{3pt}}	
        @{\hspace{0pt}}r@{\hspace{1pt}} 
        @{\hspace{1pt}}r@{\hspace{0pt}}
        @{\hspace{3pt}}c@{\hspace{3pt}}	
        @{\hspace{0pt}}r@{\hspace{1pt}} 
        @{\hspace{1pt}}r@{\hspace{0pt}}
        @{\hspace{3pt}}c@{\hspace{3pt}}	
        @{\hspace{0pt}}r@{\hspace{1pt}} 
        @{\hspace{1pt}}r@{\hspace{0pt}}
      }
      \toprule
      & 
      & \multicolumn{2}{c}{Pocket2Mol} & 
      & \multicolumn{2}{c}{DiffSMOL}    & 
      & \multicolumn{2}{c}{TargetDiff*} & 
      & \multicolumn{2}{c}{DecompDiff} & 
      & \multicolumn{2}{c}{Claude}        & 
      & \multicolumn{2}{c}{GPT-4o}        & 
      & \multicolumn{2}{c}{o1-mini}        & 
      & \multicolumn{2}{c}{o1}                & 
      & \multicolumn{2}{c}{\makecell{\agent\\(None)$\dagger$}} &
      & \multicolumn{2}{c}{\makecell{\agent\\(DeepSeek)}} & 
      & \multicolumn{2}{c}{\makecell{\agent\\(Claude)}} 
      \\
      \midrule
      &
      & \%\molecule/\target & \#\molecule/\target & 
      & \%\molecule/\target & \#\molecule/\target & 
      & \%\molecule/\target & \#\molecule/\target & 
      & \%\molecule/\target & \#\molecule/\target & 
      & \%\molecule/\target & \#\molecule/\target & 
      & \%\molecule/\target & \#\molecule/\target & 
      & \%\molecule/\target & \#\molecule/\target & 
      & \%\molecule/\target & \#\molecule/\target & 
      & \%\molecule/\target & \#\molecule/\target & 
      & \%\molecule/\target & \#\molecule/\target & 
      & \%\molecule/\target & \#\molecule/\target \\
      \cmidrule(){3-4} \cmidrule(){6-7} \cmidrule(){9-10} \cmidrule(){12-13} \cmidrule(){15-16} \cmidrule(){18-19} \cmidrule(){21-22} \cmidrule(){24-25} \cmidrule(){27-28} \cmidrule(){30-31} \cmidrule(){33-34}
      \multirow{2}{*}{\rotatebox{90}{\centering{initial}}}
      & \ANGMPT 
      & - & 100.0  & 
      & - & 100.0  & 
      & - & 100.0  & 
      & - & 100.0  & 
      & - & \phantom{0}\phantom{0}5.0 & 
      & - & \phantom{0}\phantom{0}5.0 & 
      & - & \phantom{0}\phantom{0}5.0 & 
      & - & \phantom{0}\phantom{0}5.0 & 
      & - & 100.0  & 
      & - & \phantom{0}25.1  & 
      & - & \phantom{0}24.5 \\
      %
      & \ANGVMPT 
      & 100.0 & 100.0 &
      & 99.9 & 99.9     & 
      & 87.5 & 87.5 &
      & 78.5 & 78.5     & 
      & 98.7 & 4.9       &
      & 97.3 & 4.9       & 
      & 91.3 & 4.6       & 
      & 95.3 & 4.8       & 
      & 100.0 & 100.0     & 
      & 100.0 & 25.1     & 
      & 100.0 & 24.5 \\
      %
      \cmidrule(){2-34}
      %
      \multirow{6}{*}{\rotatebox{90}{\parbox{60pt}{\centering{generated molecules}}}}
      & \quad$\qedreq\!\ge\!\overline{\qedreq}_{t}$ 
      & 53.4 & 53.4 &
      & 60.0 & 60.0 & 
      & 47.3 & 47.3 &
      & 39.1 & 39.1 & 
      & \underline{96.7} & 4.8 & 
      & 88.2 & 4.4 & 
      & 90.1 & 4.5 & 
      & 88.3 & 4.4 & 
      & 81.6 & 81.6 & 
      & \textbf{100.0} & 25.1 & 
      & 97.2 & 21.8 \\
      %
      &\quad$\lipinskireq\!\ge\!\overline{\lipinskireq}_t$
      & \textbf{99.7} & 99.7 & 
      & 72.1 & 72.1 &
      & 80.2 & 80.2 & 
      & 56.3 & 56.3 &
      & \underline{98.7} & 4.9 & 
      & 95.9 & 4.8 & 
      & 90.7 & 4.5 & 
      & 95.3 & 4.8 & 
      & 87.6 & 87.6 &
      & \textbf{100.0} & 25.1 &
      & 96.7 & 21.8 \\
      %
      &\quad$\sascorereq\!\le\!\overline{\sascorereq}_t$
      & 77.4 & 77.4 &
      & 7.5 & 7.5 & 
      & 22.3 & 22.3 &
      & 13.4 & 13.4 & 
      & \textbf{92.7} & 4.6 & 
      & 90.7 & 4.5 &
      & 81.4 & 4.1 & 
      & \underline{92.6} & 4.6 & 
      & 69.6 & 69.6 & 
      & \textbf{100.0} & 25.1 & 
      & 88.3 & 17.4 \\
      %
      & \quad$\vinareq\!\le\!\overline{\vinareq}_t$
      & 15.3 & 15.3 &
      & 24.7 & 24.7 &
      & 19.5 & 19.5 &
      & 26.4 & 26.4 &
      & \underline{63.3} & 3.2 &
      & 59.2 & 3.0 &
      & 47.9 & 2.3 &
      & 34.6 & 1.8 &
      & 80.7 & 80.7 &
      & \textbf{99.7} & 24.8 &
      & 95.8 & 21.2 \\
      %
      & \quad$\noveltyreq\!\ge\!0.8$
      & 87.6 & 87.6 &
      & \textbf{98.2} & 98.2 & 
      & 82.2 & 82.2 &
      & 70.1 & 70.1 & 
      & 46.9 & 2.4 &
      & 68.3 & 3.4 & 
      & 64.1 & 3.2 & 
      & 55.9 & 2.8 & 
      & 83.9 & 83.9 & 
      & \textbf{100.0} & 25.1 & 
      & \underline{97.8} & 22.4 \\
      \cmidrule(l{10pt}){2-34}
      & \HQ 
      & 6.4 & 6.4 &
      & 0.7 & 0.7 &
      & 1.9 & 1.9 &
      & 1.2 & 1.2 &
      & 30.3 & 1.5 &
      & \underline{35.0} & 1.7 &
      & 28.2 & 1.4 &
      & 20.7 & 1.0 &
      & 35.0 & 35.0 &
      & \textbf{99.7} & 24.8 &
      & 84.0 & 14.5 \\
      \cmidrule(){1-34}

      \multirow{5}{*}{\rotatebox{90}{\parbox{70pt}{among all targets}}}
      &
      & \%\target & \#\target & 
      & \%\target & \#\target & 
      & \%\target & \#\target & 
      & \%\target & \#\target & 
      & \%\target & \#\target & 
      & \%\target & \#\target & 
      & \%\target & \#\target & 
      & \%\target & \#\target & 
      & \%\target & \#\target & 
      & \%\target & \#\target &
      & \%\target & \#\target \\
      \cmidrule(){3-4} \cmidrule(){6-7} \cmidrule(){9-10} \cmidrule(){12-13} \cmidrule(){15-16} \cmidrule(){18-19} \cmidrule(){21-22} \cmidrule(){24-25} \cmidrule(){27-28} \cmidrule(){30-31}\cmidrule(){33-34}

      & \quad$\diversityreq\!\ge\!0.8$
      & \textbf{100.0} & 30 &
      & \textbf{100.0} & 30 &
      & \textbf{100.0} & 29 &
      & 97.7 & 29 &
      & 30.0 & 9 &
      & 90.0 & 27 &
      & 67.7 & 20 &
      & 70.0 & 21 &
      & 70.0 & 21 &
      & 76.7 & 23 &
      & \underline{97.7} & 29 \\
      %
      & \quad$N$$\ge$5\! \&\! \diversityreq
      & \textbf{100.0} & 30 &
      & \textbf{100.0} & 30 &
      & \textbf{100.0} & 29 &
      & 97.7 & 29 &
      & 27.7 & 8 &
      & 77.7 & 23 &
      & 43.3 & 13 &
      & 57.7 & 17 &
      & 70.0 & 21 &
      & 73.3 & 22 &
      & \underline{90.0} & 27 \\
      %
      & \quad$N$$\ge$5\! \&\! \HQ 
      & \underline{27.7} & 8 &
      & 3.3 & 1 &
      & 13.8 & 4 &
      & 13.3 & 4 &
      & 23.3 & 7 &
      & 10.0 & 3 &
      & 0.0 & 0 &
      & 3.3 & 1 &
      & 93.3 & 28 &
      & \textbf{97.7} & 29 &
      & 73.3 & 22  \\
      %
      & \quad\diversityreq \& \HQ
      & 23.3 & 7 &
      & 10.0 & 3 &
      & 24.1 & 7 &
      & 17.7 & 5 &
      & 10.0 & 3 &
      & \underline{33.3} & 10 &
      & \underline{33.3} & 10 &
      & 20.0 & 6 &
      & 33.3 & 10 &
      & 76.7 & 23 &
      & \textbf{90} & 27 \\
      %
      \cmidrule(l{10pt}){2-34}
      & \successrate 
      & \underline{23.3} & 7 &
      & 0.0 & 0 &
      & 13.8 & 4 &
      & 13.3 & 4 &
      & 6.7 & 2 &
      & 6.7 & 2 &
      & 0.0 & 0 &
      & 0.0 & 0 &
      & 30.0 & 9 &
      & \textbf{73.3} & 22 &
      & \textbf{73.3} & 22 \\
      \midrule
 	%
      %
      \multicolumn{34}{c}{Quality of Generated Molecules}
      \\
      \cmidrule(){2-34}	
      & \noveltyreq$\uparrow$ 
      & \multicolumn{2}{c}{\underline{0.87}} &
      & \multicolumn{2}{c}{\textbf{0.89}} &
      & \multicolumn{2}{c}{0.88} &
      & \multicolumn{2}{c}{0.86} &
      & \multicolumn{2}{c}{0.77} &
      & \multicolumn{2}{c}{0.82} & 
      & \multicolumn{2}{c}{0.79} &
      & \multicolumn{2}{c}{0.80} &
      & \multicolumn{2}{c}{0.85} &
      & \multicolumn{2}{c}{0.86} &
      & \multicolumn{2}{c}{0.86} \\
     &  \qedreq$\uparrow$ 
     & \multicolumn{2}{c}{0.51} &
     & \multicolumn{2}{c}{0.55} &
     & \multicolumn{2}{c}{0.51} &
     & \multicolumn{2}{c}{0.47} &
     & \multicolumn{2}{c}{\textbf{0.78}} &
     & \multicolumn{2}{c}{0.74} &
     & \multicolumn{2}{c}{0.75} &
     & \multicolumn{2}{c}{\underline{0.77}} &
     & \multicolumn{2}{c}{0.69} &
     & \multicolumn{2}{c}{0.70} &
     & \multicolumn{2}{c}{0.69} \\
     & \lipinskireq$\uparrow$ 
     & \multicolumn{2}{c}{\textbf{4.00}} &
     & \multicolumn{2}{c}{3.43} &
     & \multicolumn{2}{c}{3.76} &
     & \multicolumn{2}{c}{3.20} &
     & \multicolumn{2}{c}{\textbf{4.00}} &
     & \multicolumn{2}{c}{\underline{3.99}} &
     & \multicolumn{2}{c}{3.85} &
     & \multicolumn{2}{c}{\textbf{4.00}} &
     & \multicolumn{2}{c}{3.82} &
     & \multicolumn{2}{c}{3.97} &
     & \multicolumn{2}{c}{3.93} \\
     & \sascorereq$\downarrow$ 
     & \multicolumn{2}{c}{2.46} &
     & \multicolumn{2}{c}{6.15} &
     & \multicolumn{2}{c}{4.23} &
     & \multicolumn{2}{c}{4.73} &
     & \multicolumn{2}{c}{2.30} &
     & \multicolumn{2}{c}{2.16} &
     & \multicolumn{2}{c}{\textbf{2.02}} &
     & \multicolumn{2}{c}{\underline{2.03}} &
     & \multicolumn{2}{c}{2.79} &
     & \multicolumn{2}{c}{2.49} &
     & \multicolumn{2}{c}{2.62} \\
     & \vinareq$\downarrow$ 
     & \multicolumn{2}{c}{-4.74} &
     & \multicolumn{2}{c}{-4.23} &
     & \multicolumn{2}{c}{-4.94} &
     & \multicolumn{2}{c}{-4.88} &
     & \multicolumn{2}{c}{\underline{-6.69}} &
     & \multicolumn{2}{c}{-6.56} &
     & \multicolumn{2}{c}{-6.31} &
     & \multicolumn{2}{c}{-5.97} &
     & \multicolumn{2}{c}{-7.43} &
     & \multicolumn{2}{c}{-7.24} &
     & \multicolumn{2}{c}{\textbf{-7.17}} \\
      & \diversityreq$\uparrow$ 
      & \multicolumn{2}{c}{\underline{0.88}} &
      & \multicolumn{2}{c}{\textbf{0.89}} &
      & \multicolumn{2}{c}{\textbf{0.91}} &
      & \multicolumn{2}{c}{\underline{0.88}} &
      & \multicolumn{2}{c}{0.76} &
      & \multicolumn{2}{c}{0.84} &
      & \multicolumn{2}{c}{0.79} &
      & \multicolumn{2}{c}{0.80} &
      & \multicolumn{2}{c}{0.81} &
      & \multicolumn{2}{c}{0.81} &
      & \multicolumn{2}{c}{0.82} \\

      \bottomrule
      \end{tabular}
      \begin{tablenotes}[normal,flushleft]
	\begin{scriptsize}
	\setlength\labelsep{0pt}
    	\item 
	\%\molecule/\target: average percentage of molecules per target; 
	\#\molecule/\target: average number of molecules per target; 
	\ANGMPT: initially generated molecules; 
	\ANGVMPT: generated molecules that are also valid; 
	$\overline{\text{overline}}_{\scriptsize{\target}}$: the average value of corresponding property in the known drugs for 
	the target \target. 
	\%\target: average percentage of targets among all targets; 
	\#\target: average number of targets; 
	$N$$\!\ge\!$5\&\diversityreq: at least 5 molecules are generated and the set is diverse;  
	$N$$\!\ge\!$5\&\HQ: at least 5 molecules are generated and they are of high quality; 
	$\uparrow$/$\downarrow$ indicates higher/lower values are better. 
    \textbf{Bold} and \underline{underline} indicates the best and second-best results, respectively.
    *We run TargetDiff only on 29 pockets, as we cannot generate any molecules in one of the pockets (EZH2) due to some bugs in the code. 
    $\dagger$ This method represents the ``simple deterministic loop'' described in the ablation study or \agent without reasoning.
    \par
	\end{scriptsize}
  \end{tablenotes}
  \end{threeparttable}
  \end{footnotesize}      
  \vskip -5pt
\end{minipage}
}
\end{table*}

\end{document}